\PassOptionsToPackage{a5paper,total={118mm,155mm},headheight=15pt,top=17mm}{geometry}

\documentclass[plain,biber]{nowfnt}
\usepackage[noalg]{stroop}
\usepackage{tabularx}
\usepackage{comment}
\usepackage{etoolbox}
\usepackage{algorithm}
\usepackage[noend]{algpseudocode}
\usepackage{tikz-dependency}
\usepackage{forest}
\usepackage[figuresleft]{rotating}
\usepackage{euflag}
\usepackage{ltablex}
\usepackage[normalem]{ulem}
\usepackage{marginnote}

\algnewcommand{\LeftComment}[1]{\State \(\triangleright\) #1}

\colorlet{insertcolor}{orange!50!darkgray}
\colorlet{delcolor}{darkgray!80}
\colorlet{hl}{red!50!darkgray}
\colorlet{mylightpurple}{purple!30!white}
\colorlet{mydarkpurple}{purple!80!black}
\colorlet{examplecolor}{yellow!10!white}
\colorlet{theoremcolor}{black!10!white}
\mdfsetup{
    backgroundcolor=theoremcolor,
    linewidth=1pt,
    innertopmargin=0,
    leftmargin=10pt,
    rightmargin=10pt,
    innerleftmargin=5pt,
    innerrightmargin=5pt,
    leftline=false,
    rightline=false,
}

\newcommand{\structspace}{\addlinespace[1em]}

\cleartheorem{definition}
\cleartheorem{assumption}
\cleartheorem{proposition}
\newtheorem*{myremark}{Remark}
\newmdtheoremenv{definition}{Definition}
\newmdtheoremenv{assumption}{Assumption}
\newmdtheoremenv{proposition}{Proposition}
\newmdtheoremenv[
    backgroundcolor=examplecolor
]{myexample}{Example}

\newcommand\pd[2]{\partial_{#2} #1}

\newcommand\var[1]{\textsf{#1}}

\def\simplex{\triangle}

\newcommand\defeq{\coloneqq}

\creflabelformat{equation}{#2\textup{#1}#3}

\DeclareMathOperator*{\sigmoid}{sigmoid}
\DeclareMathOperator{\softmax}{softmax}
\DeclareMathOperator{\sparsemax}{sparsemax}

\DeclareMathOperator{\sparsemap}{SparseMAP}
\DeclareMathOperator{\map}{MAP}

\def\RR{{\mathbb{R}}}
\def\EE{{\mathbb{E}}}

\def\param{{\bm{\theta}}}
\def\yparam{\param_g}
\def\zparam{\param_f}

\def\e{\bm{e}}

\def\w{{\bm{w}}}
\def\x{x}
\def\y{y}

\def\szP{{|\cP|}}

\def\bmu{\bm{\mu}}

\renewcommand{\ss}{s}
\newcommand{\s}{\bm{\ss}}

\newcommand{\rv}{random variable\xspace}

\def\zz{z}
\def\z{\bm{\zz}}
\def\zhat{\widehat{\z}}

\newcommand\zsurr{\widehat{\z}}
\newcommand\zzsurr{\widehat{\zz}}
\newcommand{\zrel}{\z}
\newcommand{\zzrel}{\zz}
\let\Pr\relax
\newcommand\Pr{p}
\newcommand\dolo{g} %

\title{Discrete Latent Structure in~Neural~Networks}

\subtitle{}

\maintitleauthorlist{
Vlad Niculae
\and
Caio Corro
\and
Nikita Nangia
\and
Tsvetomila Mihaylova
\and
Andr\'{e}~F.~T.\ Martins
}

\issuesetup
{%
 copyrightowner={V. Niculae and C. Corro and N. Nangia and T. Mihaylova and A.F.T. Martins},
 volume        = xx,
 issue         = xx,
 pubyear       = 2025,
 isbn          = xxx-x-xxxxx-xxx-x,
 eisbn         = xxx-x-xxxxx-xxx-x,
 doi           = 10.1561/XXXXXXXXX,
 firstpage     = 1, %
 lastpage      = 18
 }

\author[1]{Niculae,Vlad}
\author[2]{Corro,Caio}
\author[3]{Nangia,Nikita}
\author[4]{Mihaylova,Tsvetomila}
\author[5,6,7]{Martins,Andr\'e~F.~T.}

\affil[1]{Language Technology Lab,
Informatics Institute, Faculty of Science, University of Amsterdam, Netherlands}
\affil[2]{INSA Rennes, IRISA, Inria, CNRS, Université de Rennes, France}
\affil[3]{Amazon}
\affil[4]{Department of Electrical Engineering and Automation, Aalto University, Finland}
\affil[5]{Instituto Superior T\'{e}cnico, Universidade de Lisboa, Lisbon, Portugal}
\affil[6]{Instituto de Telecomunica\c{c}\~{o}es, Lisbon, Portugal}
\affil[7]{Unbabel, Lisbon, Portugal}
\articledatabox{\nowfntstandardcitation}

\usepackage{titlesec}
\date{}

\makeatletter

\titleformat*{\section}{\nowfnt@font@section}
\titleformat*{\subsection}{\nowfnt@font@subsection}
\titleformat*{\subsubsection}{\nowfnt@font@paragraph}
\titleformat*{\paragraph}{\nowfnt@font@paragraph}
\titleformat*{\subparagraph}{\nowfnt@font@paragraph}

\def\@makechapterhead#1{\vbox to 10pc{%
  \vspace*{15\p@}%
  {\parindent \z@ \raggedright \normalfont
    \ifnum \c@secnumdepth >\m@ne
      \if@mainmatter
        \nowfnt@font@chaptertext\thechapter\par
      \fi
    \fi
    \interlinepenalty\@M
    {\nowfnt@font@chaptertext#1\strut\par}\addvspace{10pt}\nobreak\vfill
  }}}
\def\@makeschapterhead#1{%
\vbox to 10pc{%
  \vspace*{15\p@}%
  {\parindent \z@ \raggedright
    \interlinepenalty\@M
    {\nowfnt@font@chaptertext#1\strut\par}\addvspace{8pt}\nobreak
  }\vfill}} %

\renewcommand\chapter{\if@openright \cleardoublepage \else \clearpage \fi
\global \@topnum \z@ \@afterindentfalse \secdef \@chapter\@schapter}

\renewcommand\tableofcontents{%
    \chapter*{\nowfnt@contentsname
        \@mkboth{%
           \nowfnt@font@header\nowfnt@contentsname}{\nowfnt@font@header\nowfnt@contentsname}}%
    {\@starttoc{toc}}%
    \if@restonecol\twocolumn\fi
    }

\newcommand\myheaderfont{\footnotesize\normalfont\sffamily}
  \fancyheadoffset{0pt}%
\fancypagestyle{plain}
 {%
   \fancyhead{}%
   \fancyfoot[C]{\nowfnt@font@pagenumber\thepage}%
 }

\fancypagestyle{abstractpage}{%
 \fancyhf{}%
 \lhead{%
        }
 \lfoot{\raisebox{\footskip}[0pt][0pt]{\usebox\nowfnt@article@databox}}}%

\fancypagestyle{titlepage}{%
 \fancyhf{}%
 \lhead{%
       }
 }%

\renewcommand\maketitle{%
\begin{titlepage}
\let\footnotesize\small
\let\footnoterule\relax
\let\footnote\thanks\null \vfil \vskip 60\p@
\begin {center}%
{\LARGE\sffamily\bfseries \@title \par }\vskip 6em{\large \lineskip .75em
\nowfnt@printauthors\par\medskip
}
\end{center}%
\nowfnt@printaffils
\begin{center}
\vskip 2.5em{\large \@date \par }
\end{center}
\par \@thanks \vfil \null
\end{titlepage}
\setcounter {footnote}{0}\global
\let \thanks \relax
\global \let \maketitle \relax
\global \let \@thanks \@empty
\global \let \@author \@empty
\global \let \@date \@empty
\global \let \@title \@empty
\global \let \title \relax
\global \let \author \relax
\global \let \date \relax
\global \let \and \relax
}

\renewenvironment{abstract}{%
      \if@twocolumn
        \section*{\abstractname}%
      \else
        \chapter*{\abstractname}%
      \fi}

\makeatother

\begin{document}

\begin{abstract}
Many types of data from fields including natural language processing, computer vision,
and bioinformatics, are well represented by discrete, compositional structures
such as trees, sequences, or matchings.
Latent structure models are a powerful tool for learning to extract such
representations, offering a way to incorporate structural bias, discover insight
about the data, and interpret decisions.
However, effective training is challenging,
as neural networks are typically designed for continuous computation.

This text explores three broad strategies for learning with discrete latent
structure:
continuous relaxation, surrogate gradients, and probabilistic estimation.
Our presentation relies on consistent notations for a wide range of models.
As such, we reveal many new connections between latent structure
learning strategies, showing how most consist of the same small set of fundamental
building blocks, but use them differently, leading to substantially different applicability and properties.

\end{abstract}

\tableofcontents
\clearpage

\chapter*{Notation}

{%
\newcommand\smalltabspace{\addlinespace[0.33em]}
    \begin{tabularx}{\textwidth}{r X}

    \multicolumn{2}{l}{\textbf{Vectors, matrices, and indexing.}} \\
        $u, \mbv, \mbW, \cX$ &
        a scalar, a vector, a matrix, and a set. \\

        $v_i$ &
            the $i$th element of vector $\mbv$. \\

        $\mbw_j$ &
            the $j$th column of matrix $\mbW$. \\ %

        $\norm{\mbv}_p$ &
$\defeq \left(\sum_{i=1}^d |v_i|^p \right)^{1/p}$,
            the $p$-norm of $\mbv\in\bbR^d$. \\

\smalltabspace
\multicolumn{2}{l}{\textbf{Probabilities and distributions.}}\\
$\var{Y}$ & a random variable with values $y\in\mathcal{Y}$. \\
$\Pr(\var{Y}=y)$ & probability that $\var{Y}$ take the specific value $y$. \\
$\Pr(y \mid x)$ & short for $\Pr(\var{Y}=y \mid \var{X}=x)$ when unambiguous.\\
$\bbE_{\Pr(\var{Y})} [\var{Y}]$ & \(\defeq\sum_{y \in \mathcal{Y}} y\Pr(y)\),
the expected value of \(\var{Y}\).
\\

\smalltabspace
\multicolumn{2}{l}{\textbf{Differentiation.}}\\

\(\pd{f}{i}\) & the partial derivative of
\(f : \bbR^{d_1} \times \ldots \times \bbR^{d_n} \to \bbR^d\)
\wrt the \(i\)th argument.
\((\pd{f}{i})(\mbx_1, \ldots, \mbx_n)\)
is a linear \(\bbR^{d} \to \bbR^{d_i}\) map
(the pullback of \(f\)),
identified with
a \(d_i \times d\) matrix: the Jacobian transpose.
For single-argument \(f: \bbR^{d_1} \to \bbR^d\)
we omit the subscript, and if
\(\mbJ_{\mbx}\) is the Jacobian of \(f\) at \(\mbx\)
then \(\pd{f}{}(\mbx)(\mbv) = \mbJ_{\mbx}^\top \mbv\).
This transposed convention is more convenient for backpropagation.
\\

        $\pd{}{\theta}(\mathit{expr.})$ &
interprets the (possibly complicated) expression as a
single-argument function of \(\theta\) and applies \(\pd{}{}\).
\\

\pagebreak & \\
\multicolumn{2}{l}{\textbf{Convex sets.}}\\
$\bbR^d_+$ & $\defeq \{\bm{\alpha} \in \bbR^d; \alpha_i \geq 0 \text{ for all } 1 \leq i
\leq d \}$, the non-negative orthant;\\
$\simplex_d$ & $\defeq \{\bm{\alpha} \in \bbR_+^d; \sum_i \alpha_i = 1\}$,
the simplex with $d$ bins, containing all probability distributions over $d$
choices; \\
$\conv(\cZ)$ & the convex hull of $\cZ$, \ie, the smallest convex set containing
$\cZ$.
\end{tabularx}
}

\chapter{Introduction}\label{chapter:intro}

\section{Motivation}

Machine learning (ML) is often employed to
build predictive models for analyzing rich data, such as images,
text, or sound.
Most such data is governed by underlying \emph{structured
representations}, such as segmentations, hierarchy, or graph structure. For example,
natural language sentences can be analyzed in terms of their \emph{dependency
structure}, yielding an arborescence of directed grammatical relationships
between words~(\cref{fig:structs}).

It is common for practical ML systems to be structured as \textbf{pipelines},
including off-the-shelf components (analyzers) that produce structured representations of
the input, used as features in subsequent steps of the pipeline.
On the one hand, such architectures require availability of these analyzers (or of the data to train
them). Since the analyzer may not be built with the downstream goal in
mind, a disadvantage of pipelines is that they are prone to error propagation.
On the other hand, they are
transparent: the predicted structures can be directly inspected and used to interpret downstream predictions.
In contrast, \emph{deep neural networks} rival and even outperform pipelines by
learning dense, continuous representations of the data, solely driven by the
downstream objective.

However, the popular success of end-to-end deep learning hides some fundamental challenges.
For example, large language models are still based on a pipeline system in which tokenization is an independent pre-processing step.
Another known limitation is the
structural generalization problem \citep{yao2022structuralgen}:
sequential architectures (both recurrent neural networks and self-attentive networks) have difficulties to generalize to unseen (recursive) combinations of known parts.
It is possible to tackle this problem by inducing latent structured representations \citep{bogin2021latemtcomp,liu2021compgen}.
Similar limitations are known for length generalization \citep{anil2022length,zhou2024length}.
Another important research direction in the natural language processing community is intermediate plan-based representations for text generation \citep{narayan2023conditionalgen,liu2023visualstorytelling},
where latent structures may play an important role, for example when learning with limited information \citep{xu2022latentqueries}.
Beside natural language processing, latent structure inference is also a key topic in computer vision for unsupervised segmentation and learning object-centric representations \citep{van2018relational,greff2019multiobject,locatello2020slotatt,elsayed2022savi}.

This text is about neural network models that induce \textbf{discrete latent
structure}, combining the strengths of both end-to-end and pipeline
systems.
In the following, we do not assume a specific downstream application in natural language processing nor computer vision.
Our presentation follows an abstract framework that allows to focus on technical
aspects related to end-to-end learning with deep neural networks.

\section{Supervised Learning}
We begin by establishing the common setup of predictive machine learning.
A prediction function
is a map associating to input $x \in \cX$ an output $y \in \cY$.
Prediction functions usually rely on a \emph{scoring} function
\begin{equation}
	M(x, y;~\param),
\end{equation}
which returns the score, or preference, for some candidate $y \in \cY$, given an
input $x$.
In our setting, $M$ is a parametric function with learnable parameters \(\param\).
For simple classification problems, 
$M$
could be a feed-forward network with $x$ as input,
and a $|\cY|$-dimensional output, such that the $y$th position of the output is
$M(x, y;~\param)$.
Our notation allows for more involved setting like predicting structured
	objects (for examples graphs, see \cref{chapter:struct}).
To make predictions,
we search for the output of maximum weight
\begin{equation}
	\hat{y}(x;~\param) \defeq \argmax_{y' \in \cY} M(x, y';~\param)\,. \label{eq:ch1_argmax}
\end{equation}
In many cases, we are also interested in a distribution over outputs.
Assuming $\cY$ is a finite set, a common choice is to rely on a Boltzmann-Gibbs distribution, also called \emph{softmax} \citep{bridle1989softmax},
defined as follows:
\begin{align*}
	\Pr(y \mid x)
	&=\frac{\exp M(x,y;~\param)}{\sum_{y' \in \cY} M(x,y';~\param) } \quad \text{for}~y \in \cY\,, \\
	&\propto \exp M(x,y;~\param)\,.
\end{align*}
Note that the most probable output under distribution $p(\cdot|x)$ is equal to $\hat{y}(x;~\param)$.

In the supervised learning scenario, we assume access to a dataset $\mathcal D$ containing samples of input/output pairs $(x, y) \in \mathcal D$.
Parameters $\param$ are fixed to minimize the empirical risk
\begin{equation}\label{eqn:avg-loss}
	{L}_\text{avg}(\param) \defeq \frac{1}{|\mathcal{D}|}\sum_{(x,y) \in \mathcal{D}}
	L(y, x;~\param)\,,
\end{equation}
where $L$ is a loss function \citep{vapnik1991erm}.
For practical reasons, the loss function used for classification problems is usually not the targeted evaluation function (for example the 0-1 loss which is equal to 1 if and only if the model predicts the expected output) but a surrogate loss that is amenable for gradient-based optimization.
Statistical consistency of such surrogates has been widely studied \citep{bartlett2006convexity,reid2010binary,vernet_2016}.
A common choice is the cross-entropy loss,
\begin{equation}
	\label{eqn:cross-entropy}
	L(x, y;~\param) = -M(x, y;~\param) + \log \sum_{y'\in\cY} \exp M(x, y'; ~\param)\,,
\end{equation}
which is simply the model negative log-probability of gold output under a Boltzmann-Gibbs distribution.
Then, \cref{eqn:avg-loss} can be interpreted as
\emph{maximum likelihood} estimation of $\param$. Non-probabilistic losses
like the hinge loss or the perceptron loss fit the framework as well.

From a computational point of view, both training and prediction under such a
model eventually requires evaluating or optimizing a function of the form
\[ g(x, y;~\param), \]
which may refer to either the scoring model $M$ or the loss $L$.
Therefore, we shall use the generic functional notation $g(x,y;~\param)$ in
the following. In this text, we are interested in on computing (or approximating) partial derivatives with respect to all values in $\theta$ via the backpropagation algorithm for automatic differentiation \citep{linnainmaa1970representation}.

\paragraph{Gradient-based learning.}
The gradient method for minimizing a differentiable function \(F: \bbR^d \to
\bbR\) iterates
\begin{equation}\label{eqn:g-update}
\param^{(t+1)} \gets \param^{(t)} + \eta^{(t)}
(\pd{F}{})(\param^{(t)})\,,
\end{equation}
where $\eta^{(t)}$ is a step size schedule, and \(\pd{F}{}(\cdot)\) is
identified with its column-vector Jacobian.
This method converges to a
stationary point of $F$ under some assumptions on the step size
\citep[\S 1.2.2]{bertsekas-nonlin}.
Often in machine learning evaluating \(F\) is slow and memory-intensive, as it depends on the entire training data;
this is the case in \cref{eqn:avg-loss}.
In such cases, the stochastic gradient \citep[SG,][]{robbinsmonro} method may be preferred.
The SG
method replaces the gradient with a stochastic
direction $\var{G}$ such that
\begin{equation}\label{eqn:grad-unbiased}
\EE[\var{G}] = \pd{F}{}(\param^{(t)})\,,
\end{equation}
followed by updating
\begin{equation}\label{eqn:sg-update}
\param^{(t+1)} \gets \param^{(t)} + \eta^{(t)} \var{G}\,.
\end{equation}
This method also converges to a stationary point under mild assumptions
\citep{bertsekas2000gradient}
: mainly, requiring smooth $F$, square-summable
decreasing step sizes, and a linear bound on the variance of
$\var{G}$ \wrt the norm of the gradient of $F$.
If $F$ takes the form of an average, \ie, $F(\param) = \frac{1}{N}
\sum_{i=1}^N F_i(\param)$ (for instance \cref{eqn:avg-loss}),
then $\var{G}$ may be chosen as a
single sample $F_i(\param)$ where $i$ is drawn uniformly from $\{1,\dots,N\}$,
or a mini-batch estimator.
The gradient and stochastic gradient methods can be extended to a broader family
using acceleration, momentum, and adaptivity
\citep{nesterov1983method,adam,loshchilov2018decoupled,acceleration}.
Algorithms in
this family
are the \emph{de facto} choice in deep learning at the
time of writing. For this reason,
our work focuses on compatibility with gradient-based
learning.

\paragraph{Backpropagation and the Chain Rule.}
Given a composition of functions
\(u: \bbR^m \to \bbR^n\),
\(v: \bbR^n \to \bbR^p\),
\(w: \bbR^p \to \bbR^q\),
and their composition
\((w \circ v \circ u)(\param) \defeq w(v(u(\param)))\)
we have:
\begin{equation}\label{eq:chainrule-def}
\pd{(w \circ v \circ u)}{}(\param)
=
(\pd{u}{})(\param)
\circ
(\pd{v}{})(u(\param))
\circ
(\pd{w}{})(v(u(\param)))\,.
\end{equation}
The derivatives are applied in the opposite order compared to the computation.
This is known as \emph{backpropagation} or \emph{reverse-mode automatic
differentiation}
\citep{evaluatingderivatives}
and is popular in deep learning, where models are built using
such compositions, with the final layer \(w\) having a scalar output
(loss).
The \emph{forward pass} computes and stores the intermediate values that appear
in \(w \circ v \circ u\), and the backward pass invokes the \(\pd{}{}\) operator
to propagate gradients from the output to the input.
In the most popular software frameworks today \citep[\eg,][]{pytorch},
elementary building blocks are provided as composable modules, with
implementations providing \emph{forward} calls \(f(\param)\) and
\emph{backward} calls (vector-Jacobian products) \(\pd{f}{}(\param)(\mbz)\),
and the automatic differentiation engine handles the composition.

\section{Latent Representations}
Our main motivation is to go beyond direct mappings $x \to y$, toward machine
learning models with latent representations.
In this text, we take a rather inclusive view of what constitutes a latent
representation
\citep{replearning}.
We call a latent representation $z \in \mathcal{Z}$
an object designed to capture some relevant property of a data point $x \in \cX$,
which can be inferred based on $x$, but is typically unobserved. In particular,
we cover but do not require probabilistic modeling of $z$
\citep{Bishop1998}.
On the other hand, we are explicitly interested in discrete and structured
latent representations.

Latent representations are often designed with downstream tasks in mind:
we may look for a model of $y \in \cY$ that has access not only to $x$ but also to the
representation $z$:
\begin{equation}\label{eqn:downstreammodel}
g(x, y, z;~\yparam)\,. \qquad \text{\emph{(downstream model)}}
\end{equation}

\begin{myremark}
During prediction from a pretrained model, we may think of $g$
as a classifier returning the score of
class \(y\).
For training the model, however,
we may want to think of $g$ as some loss function on top of the same
classifier. Mathematically, this distinction is irrelevant for the
purpose of our text, which is the modelling of $z$, so we henceforth
use
$g(x, y, z;~\yparam)$
to denote either. Practitioners
should exercise caution.
\end{myremark}

A downstream model
as in \cref{eqn:downstreammodel}
is not directly usable, since $z$ is unknown both at training
and at test time.
Therefore, the problem we are concerned with in this text is jointly learning
to predict $z$ from $x$ using an encoder $f : \cX \times \cZ \to \RR$:
\begin{equation}
f(x, z; ~\zparam),
\qquad\text{\emph{(encoder model)}}
\end{equation}
assigning higher values to better-fitting choices of $z$ to the given $x$.

The key challenge of learning latent variable models is that we cannot learn
$\zparam$ using standard supervised approaches, since \(z\) is not
observed.
This text is about how to learn a good encoder model \(f\)
jointly with the downstream model only from pairs \((x,y)\).
During training, the downstream model gets direct supervision,
but the encoder model only gets a form of \emph{distant supervision},
its only learning signal is coming in the form of gradients propagated through
the downstream model.
Joint learning with latent structure in this scenario is the main topic of our
text.
The next three paragraphs outline the main ways to train end-to-end models in
such encoders; the main part of our text
(\cref{chapter:relax,chapter:surrogate,chapter:expectation}) later goes into detail.

\paragraph{Pretraining and pipelines.}
A first strategy is to sidestep the issue altogether and obtain
supervision. This poses no challenge mathematically, and is not studied further
in this text, but serves as a motivating base case:
If in fact some training pairs $(x, z)$ are available, it is promising to first
train a model $f(x, z; ~\zparam)$ and then deploy a two-step pipeline:
\begin{enumerate}
\item predict $\hat{z} = \argmax_{z' \in \cZ} f(x, z'; ~\zparam)$,
\item use downstream model
$g(x, y, \hat{z};~\yparam)$.
\end{enumerate}
The parameters of the downstream model \(\yparam\) can now be trained in a
fully-supervised fashion, since \(\hat{z}\) is a known fixed input.
This corresponds to the
time-tested approach of using off-the-shelf analysis models (parsers, object
detection, entity recognizers, etc.) as a pre-processing step.
This approach is vulnerable to two main sources of error: \emph{domain shift},
due to the fact that $\zparam$ is likely trained on samples coming from a
different distribution than the one $\cD$ is drawn from, and \emph{error
propagation}, due to the lack of mechanism for improving $\zparam$ if the model
makes errors. The latent representation treatment we propose mitigates both
these concerns by allowing the fine-tuning of $\zparam$ with signal from
downstream, see \citep{spigot} for examples.

\paragraph{Deterministic latent representations.}
A straightforward idea for end-to-end learning would be to characterize the
mapping from $x$ to a promising candidate $\hat{z}$ as a function,
\[ \hat{z}(x; \zparam), \]
which implicitly defined by the encoder $f$. (For example,
$\hat{z}(x) = \argmax_{z \in \cZ} f(x,z)$.)
Then,
an end-to-end
model emerges as a composition of functions:
\begin{equation}
g\left(x, y, \hat{z}(x; \zparam);~\yparam\right).
\end{equation}
This resembles the pipeline approach, but now we aim to train $\zparam$ and
$\yparam$ jointly using gradient methods.
Depending on how $\hat{z}$ is constructed, we may have an end-to-end
differentiable relaxed model (\cref{chapter:relax}) or a discrete model
optimized with surrogate gradient heuristics (\cref{chapter:surrogate}).
Both cases will require further assumptions compared to the pipeline
approach with frozen $\zparam$,
but require no supervision on \(z\).

\paragraph{Probabilistic latent variables.}\label{sec:proba-rv}
Alternatively, we can gain expressiveness by modelling
latent representations as \textbf{random
variables} whose distribution is induced by the encoder $f$.
Notationally, we define a \rv $\var{Z}$
taking values $z \in \cZ$,
with distribution
$\Pr(\var{Z}=z \mid x;~\zparam)$
parametrized in some way using $f$
(\eg, $\Pr(z\mid x;~\zparam) \propto \exp f(x, z;~\zparam)$.)
Then, the end-to-end model will consider not a single value of $z$ but the expectation over
all possible values $z \in \cZ$:
\begin{equation}\label{eqn:marginalization}
\begin{aligned}
\bar g(x, y;~\zparam,\yparam)
  \defeq & \EE_\var{Z}\left[g(x, y, \var{Z}; \yparam)\right] \\
       = & \sum_{z \in \mathcal{Z}} g(x, y, z; \yparam) \Pr(z \mid x; \zparam). \\
\end{aligned}
\end{equation}
The expected loss depends on both $\zparam$ and $\yparam$,
and so provides a learning signal to both the encoder and the downstream model.
In particular, some choices of $g$ can correspond to a probabilistic treatment
of $\var{Y}$ as well, making this strategy interesting for generative modelling.
We study methods for probabilistic latent variables in
\cref{chapter:expectation}.
Broadly speaking, these methods tend to require fewer assumptions compared to
deterministic ones, but come at a higher computational cost.

\begin{myremark}

What sets apart a latent representation from an arbitrary ``hidden layer''
is that the former is designed to capture a specific aspect of $x$,
relevant to the modeler.
In this text, we focus on discrete $z$ with structural constraints that can
guide it to take a certain form of interest (\eg, alignments, syntax.)
This is often (but not necessarily) reflected in the more transparent, informed
way in which the way the downstream model $g$ accesses $z$.

\end{myremark}

\section{Further History and Scope}

Latent variable models have a long history in
ML, especially for unsupervised learning.
In this section, we briefly survey this history and clarify the scope of this
work.

\paragraph{Shallow models.}
Many popular models fall under this umbrella, typically with linear $f$ and $g$.
Factor analysis (FA) is an unsupervised representation learning model ($\cY =
\RR^d$, $\cX = \varnothing$) with continuous latent variables
($\cZ = \RR^k$) defined by
\citep[\S 21.1]{barber}
\begin{equation}
   f(\bm{y}, \bm{z}; ~\param) = -\frac{1}{2} (\bm{y} - \bm{Fz} - \bm{\mu})^\top
\bm{\Sigma}^{-1}(\bm{y}-\bm{Fz}
- \bm{\mu}) \,,
\end{equation}
where the covariance $\bm{\Sigma}$ is a diagonal matrix.
If $\bm{\Sigma}$ is further constrained to be isotropic, FA reduces to
probabilistic PCA.\@
The discrete counterpart is the Gaussian mixture model (GMM)
where $\cZ = \{1, 2, \ldots, k\}$ is a discrete variable, and we have
\begin{equation}
   f(\bm{y}, z; ~\param) = -\frac{1}{2} (\bm{y} - \bm{\mu}_z)^\top
\bm{\Sigma}_z^{-1}(\bm{y}-\bm{\mu}_z) \,.
\end{equation}
For supervised regression of continuous $\bm{y}$ given $\bm{x}$, the counterpart
of FA is the linear mixed effect model
\begin{equation}
   f(\bm{y}, \bm{z}, \bm{x}; ~\param) = -\frac{1}{2} (\bm{y} - \bm{Fz} - \bm{Wx})^\top
\bm{\Sigma}^{-1}(\bm{y}-\bm{Fz}
- \bm{Wx}) \,.
\end{equation}
and the counterpart of the GMM is the mixture of linear regressions
\begin{equation}
   f(\bm{y}, z, \bm{x}; ~\param) = -\frac{1}{2} (\bm{y} - \bm{W}_z\bm{x})^\top
\bm{\Sigma}_z^{-1}(\bm{y}-\bm{W}_z\bm{x}) \,,
\end{equation}
corresponding to learning a separate linear regression model for each cluster
component.
All of the above can be fit by expectation-maximization algorithms, with
the notable exception of probabilistic PCA, for which the exact solution can be
found from a single SVD of the design matrix.
Extensions to categorical (\ie, classification) models of $\cY$ are
mostly studied in the context of mixed effects models within the framework of
hierarchical generalized linear models.

\paragraph{Unsupervised linguistic structure discovery.}
An important line of work in natural language processing is the use of
latent structures for language modeling (\ie, learning a distribution over sentences) in a Bayesian setting,
that is by defining a Bayesian network whose observations are sentences and latent variables include structure modeling.
Then, parameter inference from raw texts can provide structured representation of texts.
Although useful for unsupervised and semi-supervised structured prediction, it is important to bear in mind that part of this line of work is also motivated by the goal of automatically discovering structures that may be useful for linguistic research.

Segmentation models are often used for discovering word boundaries
\citep{brent1999segmentation,venkataraman2001segmentation,goldwater2006contextual},
especially
in languages that do not have explicit boundary markers and speech processing
for (non-written) low resource languages \citep{zanon2022segmentation}.
Unsupervised tagging models learn to group similar words in the same class \citep{goldwater2007bayesiantagger}.
They are mainly based on hidden Markov models, possibly with an infinite number of classes \citep{beal2001infinitehmm}.
Syntactic models aim to represent more complex relations between words in a sentence than as a sequence of words.
We often differentiate two types of syntactic structures:
\begin{itemize}
	\item
		Such models are mainly based on latent probabilistic context-free grammars \citep{lari1990insideoutside,clark2001unsupervisedpcfg,johnson2007bayesianpcfg}.
		Phrase structures or constituency trees, that model syntax by grouping words in hierarchical spans.
	\item
		Dependency trees, that model syntax using bilexical dependencies between words.
		The main approach is called \emph{dependency model with valence} \citep{klein2004dmv}.
\end{itemize}
Beyond the sentence level,
previous work considered latent modeling of discourse structures \citep{chen2009latentperm}
and topic segmentation, which aims to model topical changes in a document \citep{eisenstein2008topicsegmentation,du2013topicsegmentation}.

Note that these works are not covered in this manuscript.
\citet{cohen2019bayesian} covers all basic techniques in the purely
probabilistic setting (\eg, parameter inference techniques like Markov chain Monte Carlo and variational inference) including the use of priors to bias models toward linguistically plausible structures.
These approaches exploit probability distribution structures and their (simple) parametrization, which is not possible with the neural network setting that we cover in this manuscript.
We instead focus on techniques for learning neural models in end-to-end approaches with limited assumptions, including but not limited to techniques described by \citet{kim-tutorial}.

\paragraph{Deep models.}
Sigmoid belief networks \citep[SBN,][]{neal1992gibbs} and Boltzmann machines \citep[BM,][]{ackley1985boltzmann} are popular generative neural networks with discrete latent variables that have a long history in machine learning.
They are graphical models (Bayesian network in the case of SBN, factor graph in the case of BM) that use implicit parametrization using a small neural network instead of explicit contingency tables.
SBNs can naturally describe deep architectures with several layers of latent variables whereas RBMs can be stacked to achieve a similar goal \citep{hinton2006dbn,salakhutdinov2009deepbm}.
Straightforward approaches to fit these models are based on Markov chain Monte Carlo estimation of the gradient \citep{neal1992gibbs,hinton2002poe,hinton2006dbn},
which can be slow in practice.
Generalization of the expectation-maximization \citep[EM,][]{dempster1977em} algorithm using mean field theory approximation \citep{parisi1988statistical} allows fast training of these models \citep{peterson1987mft,saul1996mft}.
A downside of EM is that it relies on strong assumption on factors'
parametrization (\ie, simple linear projection),
and therefore does not extend to complex neural parametrization. %
This contrasts with methods studied in this manuscript that focus on techniques for learning discrete latent variables that (1) can learn more complex latent structures than binary variables and (2) are compatible with the modern end-to-end learning framework.
Moreover, some of the techniques we described do not have a probabilistic interpretation of latent variables.

Nonlinear models parametrized by neural networks have proven themselves effective for generative modeling.
Prominent among them is the \emph{variational auto-encoder}
\citep[VAE,][]{kingma2013auto,rezende2014dgm},
which is a Bayesian network where conditional distributions are parametrized by deep neural networks.
This means that variational methods used for SBN are not applicable anymore.
Key to the success of the VAE is the ``evidence lower bound'' (ELBO)
objective
\begin{equation}
	\begin{aligned}
		L(\mbx ; \yparam) &= -\log \EE_{\Pr(\var{Z})}\left[
\Pr(x \mid \var{Z}; \yparam)
\right] \\
		&\leq \textsc{KL}[\Pr(\var{Z} \mid x ; \zparam), \Pr(\var{Z}) ] -
\underbrace{
\EE_{\Pr(\var{Z} \mid x; \zparam)}[\log \Pr(x \mid \var{Z}; \yparam)]
}_{\text{reconstruction term}} \\[-1.1em]  %
		&\coloneqq \bar{L}(\mbx, \yparam, \zparam)\,,
	\end{aligned}
\end{equation}
where $\textsc{KL}$ denotes the Kullback–Leibler divergence and $\Pr(\var{Z} \mid x ; \zparam)$
corresponds to the approximate posterior.
The conditional and approximate posterior distributions are fully specified by the Gibbs distributions
\begin{equation*}
	\Pr(\mbx \mid \mbz; \yparam) \propto \exp f(\mbx; \mbz, \yparam)\,, \qquad
	\Pr(\mbz \mid \mbx; \zparam) \propto \exp g(\mbz; \mbx, \zparam)\,.
\end{equation*}
As such, the reconstruction term of the ELBO is
similar to
\cref{eqn:marginalization}.

In our framework, we may take $\mby=\mbx$ to represent an autoencoding task,
and set, for a Gaussian latent and Gaussian output VAE,
\begin{equation}
\begin{aligned}
f(\mbz; \mbx, \zparam)
&=
\big(\mbz - \bmu_{\mbz}(\mbx; \zparam)\big)^\top
\bm{\Sigma}_{\mbz}^{-1}(\mbx; \zparam)
\big(\mbz - \bmu_{\mbz}(\mbx; \zparam)\big)\,,\\
g(\mbx ; \mbz; \yparam) &=
\big(\mbx - \bmu_{\mbx}(\mbz; \yparam)\big)^\top
\bm{\Sigma}_{\mbx}^{-1}(\mbz; \yparam)
\big(\mbx - \bmu_{\mbx}(\mbz; \yparam)\big)\,,
\end{aligned}
\end{equation}
\ie, a neural network is used to generate the parameters of an observation
distribution and of an approximate posterior; this strategy is known as amortization.

In this text, we focus on deep models with discrete, structured latent variables.
This differs from works that extend the original VAE with richer priors or structured inference networks \citep[amongst others]{johnson2016svae,lin2018variational,pearce2020gpvae,zhao2023rsvae}.
For a tutorial on latent variable learning with a focus on
probabilistic models for language, we refer the reader to
the thorough tutorial by
\citet{kim-tutorial}.

\section{Roadmap}

Before getting into the matter of
discrete latent structure, in
\cref{chapter:struct} we revisit the tools of the trade of (supervised)
structure prediction; they will prove essential for the latent case as well.

The central \cref{chapter:relax,chapter:surrogate,chapter:expectation}
form the main part of our text, covering three different directions to take
for learning deep networks with discrete latent structure.
In \cref{chapter:relax} we explore a deterministic approach to learning latent
structure, using
a fundamental %
\emph{relaxation}
strategy, at the cost of partially abandoning discreteness.
Then, in \cref{chapter:surrogate} we discuss a range of methods that regain
discreteness by introducing a gap between the learning objective and the desired
model. Finally, in \cref{chapter:expectation} we study strategies for
approximately minimizing the true stochastic objective, allowing for the most
flexible latent structure models, at a controllable computational cost.
\Cref{chapter:conclusions} summarizes the field and provides a
table
of various trade-offs and applicability of the discussed
methods, along with pointers to prominent libraries.

\chapter{Structure Prediction Background}\label{chapter:struct}
\section{Overview}

Before handling combinatorial structures as
latent representations, we must establish a formalism for
representing and computationally modelling such structures.
In this section, we provide the necessary background, which draws
from the field of \emph{structure prediction}.
The examples in this section are therefore examples of structured-output
models, not of models with latent structure, but many of them will reappear in
later sections in latent structure form.

\begin{definition}
A structured representation $\z \in \cZ$ is a discrete object consisting of a combination of
interdependent parts $p \in \cP$.
The set of parts is problem-specific.
The set of valid structures
\(\cZ \subseteq \{0,1\}^{|\cP|}\)
can encode global structural constraints on
which parts can be simultaneously present in a structure,
and, like the set of parts, differs based on what is being modeled.
\end{definition}
When convenient, we may see $\z$ as a binary vector indexed by $\cP$, such that
$\zz_p = 1$ if part $p$ is present in the structure $\z$ and $\zz_p=0$
otherwise. From this perspective, $\cZ \subseteq \{0, 1\}^{|\cP|}$.
\footnote{Since the parts are a discrete set, indexing by them is
equivalent to assigning the parts numeric indices \(\{1, 2, \ldots, |\cP|\}\).
We assume an implicit indexing and use either notation interchangeably.}
This is a powerful but abstract definition. ~\Cref{fig:structures}
presents some common and useful types of structures, under our established
notation.
The first two are polar opposite extreme examples,
and we describe them in more detail below.
\begin{myexample}[Unconstrained]\label{ex:unconstrained}
In order to
represent a combination of choices, such as sandwich
toppings,
we show how we represent a global configuration as a vector of binary variables.
The parts are the possible choices, \eg,
\(\cP = \{\) \textsf{lettuce}, \textsf{tomato}, \(\ldots\}\),
and each part is encoded with a binary indicator \(z_p \in \{0,1\}\)
for each \(p \in \cP\). Beyond the binary construction,
the structures are not constrained:

any subset (including no toppings at all) is acceptable.
Thus, $\cZ = \{0, 1\}^{|\cP|}$.
The number of admissible structures is thus exponential in the number of
parts.
When using such an output space for supervised machine learning,
the resulting task is known as multilabel classification \citep{mlclf}.
\end{myexample}
\begin{myexample}[One-of-K]\label{ex:oneofk}
In contrast,
the \textbf{one-of-K} setup forces us to pick exactly one of the %
parts from \(\cP\).
For example, a word in a sentence must have one part-of-speech tag.
In this case, \(\cP=\{\textsf{NOUN}, \textsf{VERB},\ldots\}\) and
\[
\cZ = \left\{ \z \in \{0, 1\}^{|\cP|} : \sum_{i=1}^{|\cP|} \zz_i = 1\right\}
= \{ \bm{e}_1, \ldots, \bm{e}_{|\cP|} \}\,.
\]

The number of admissible configurations is thus exactly the same as the number
of parts.
In supervised learning, this output space leads to the standard multiclass
classification problem \citep[Ch.~10.3]{pml1book}.
\end{myexample}
The next three examples illustrate how structure prediction generalizes
and explores the space between the two extremes encountered in the first two
examples.

\begin{figure}%
\centering\footnotesize%
\begin{tikzpicture}
\node (h1) at (0,0) {\includegraphics[width=1em]{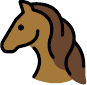}};
\node (h2) at (0,1) {\includegraphics[width=1em]{fig/horse.pdf}};
\node (h3) at (0,2) {\includegraphics[width=1em]{fig/horse.pdf}};

\node (j1) at (1,0) {\includegraphics[width=1em]{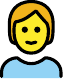}};
\node (j2) at (1,1) {\includegraphics[width=1em]{fig/person.pdf}};
\node (j3) at (1,2) {\includegraphics[width=1em]{fig/person.pdf}};

\draw[thick] (h3) to (j3);
\draw[thick] (h1) to (j2);
\draw[thick] (h2) to (j1);
\end{tikzpicture}%
\qquad%
\begin{dependency}[theme=simple]
\begin{deptext}[column sep=.5ex] Sleep \& the \& clock \& around
 \\ \end{deptext}
\depedge{1}{4}{}
\depedge{4}{3}{}
\depedge{3}{2}{}
\deproot{1}{}
\end{dependency}%
\qquad%
\begin{forest}
  where n children=0{tier=word}{}
  [,parent anchor=south,s sep=0
    [,inner sep=0,outer sep=0[Sleep]]
    [,parent anchor=north,child anchor=north,s sep=0
      [,parent anchor=north,child anchor=north,s sep=0
       [,inner sep=0,outer sep=0,l=0[the]]
       [,inner sep=0,outer sep=0,l=0[clock]]
      ]
      [,inner sep=0,outer sep=0[around]]
    ]
  ]
\end{forest}
\caption{\label{fig:structs}Some example structures. Left: linear assignment
(matching); center: dependency parse tree (directed arborescence);
right: binary constituency parse tree (binary tree).}
\end{figure}

\begin{myexample}[Linear assignment]\label{ex:matching}
The linear assignment (one-to-one, or bipartite matching) setup
\citep{hitchcock1941distribution}
amounts to finding an optimal pairing between two equal-size sets
\(A,B\)
(\eg, horses and jockeys),
such that each horse gets one jockey and each jockey gets one horse
(\cref{fig:structs}, left). Formally,
\[\begin{aligned}%
\cP &=\{i \leftrightarrow j: i \in A, j \in B\},\\
\cZ &= \left\{\z \in \{0, 1\}^{|\cP|}:
\sum_{i\in A} \zz_{i \leftrightarrow j} = 1~\text{for each}~j,
\sum_{j\in B} \zz_{i \leftrightarrow j} = 1~\text{for each}~i
\right\}
\end{aligned}
\]
Writing \(|A|=|B|=n\), notice \(|\cP|=n^2\). By further noting that any
valid assignment is a permutation, we see that \(|\cZ| = n!\).
We will later see how this elementary set already poses challenges,
\ie, full probabilistic modeling over this set is intractable.
\end{myexample}
\begin{myexample}[Non-projective dependency parsing]
This example comes from the domain of natural language syntax
\citep[Ch.~19]{jurafsky-martin}
given a sentence,
search for the best dependency tree to describe the dependency (syntactic
governance) relationships between its
words (\cref{fig:structs}, middle): the parent word is interpreted as a
\emph{head} word and the child is a \emph{modifier}.
A dependency tree is an \emph{arborescence}, that is, a directed rooted subgraph in which
every vertex (\ie, word), has at most one incoming arc and is reachable from the root.
We next show how to represent such dependency trees in our framework.
Given a sentence with words \(w_1, \ldots, w_n\),
we define the parts of a dependency tree as the possible arcs:
\[
\cP = \left\{h \rightarrow m:
h \in \{*, w_1, \ldots, w_n\}, m \in \{w_1, \ldots, w_n\},
h \neq m \right\},
\]
where the asterisk denotes a special token identifying the root.
In order for a set of such arcs to correspond to a valid dependency tree,
it is required that
every node except \(*\) have exactly one parent, and that there be no cycles in
the resulting graph \citep{martins-etal-2009-concise}.
From a generalization of Kirchoff's matrix-tree theorem \citep{Kirchhoff1847}
due to \citet[Thm.~VI.27]{tutte2001graph},
we obtain \(|\cZ|= (n+1)^{n-1}\).

\end{myexample}
\begin{figure}
    \centering
    \small
    \begin{tabular}{p{3.7cm} p{6.1cm}}
    \toprule
    structure & description \\
    \midrule
    \multicolumn{2}{l}{\textbf{Unconstrained (multiple choice)}} \\
    {\tikzset{circ/.style={draw=gray,circle,inner sep=0,minimum size=13pt}}
\begin{tikzpicture}[scale=.6,baseline=(nn.base)]
\small
    \draw[rounded corners,gray!60,thin] (-.5, -.5) rectangle (.5, 3.5) {};
    \node[circ,label=right:{\textsf{red onion}},fill=mylightpurple] at (0, 0) {};
    \node[circ,label=right:{\textsf{corn}}] at (0, 1) {};
    \node[circ,label=right:{\textsf{tomato}},fill=mylightpurple] at (0, 2) {};
    \node(nn) [circ,label=right:{\textsf{lettuce}},fill=mylightpurple] at (0, 3) {};
\end{tikzpicture}}
    &
		$\cP=\{ \textsf{lettuce}, \textsf{tomato}, \ldots \}$;
		Structures can have any combination of parts.
		\newline\newline
		$|\cZ| = 2^{|\cP|}$
		\\ \structspace
    \multicolumn{2}{l}{\textbf{One-of-K (XOR)}} \\
    {\tikzset{circ/.style={draw=gray,circle,inner sep=0,minimum size=13pt}}
\begin{tikzpicture}[scale=.6,baseline=(nn.base)]
\small
    \draw[rounded corners,gray!60,thin] (-.5, -.5) rectangle (.5, 3.5) {};
    \node[circ,label=right:{\textsf{preposition}}] at (0, 0) {};
    \node[circ,label=right:{\textsf{adjective}}] at (0, 1) {};
    \node[circ,label=right:{\textsf{verb}},fill=mylightpurple] at (0, 2) {};
    \node(nn) [circ,label=right:{\textsf{noun}}] at (0, 3) {};
\end{tikzpicture}}
    &
		$\cP=\{ \textsf{noun}, \textsf{verb}, \ldots \}$;
		Structures have exactly one part
		($\bm{1}^\top \z = 1$)
		\newline\newline
		$|\cZ| = |\cP|$
		 \\ \structspace
    \multicolumn{2}{l}{\textbf{Non-projective dependency parsing}} \\
    {\input{fig/structures-03tree.tikz.tex}}
	&
		\sloppypar{%
			$\cP = \{ {\textsf{h} \rightarrow \textsf{m} : \textsf{h} \in S \cup \{*\},}$
			${\textsf{m} \in S,}$
			${\textsf{h}\neq\textsf{m}}\}$
			set of possible arcs. Valid structures are connected and acyclic.
			\newline\newline
			$|\cZ| = (|S| + 1)^{|S| - 1}$
		}
		\\ \structspace
    \multicolumn{2}{l}{\textbf{Linear assignment}} \\
    \raisebox{8pt}{{\tikzset{circ/.style={draw=gray,circle,inner sep=0,minimum size=13pt},
         factor/.style={draw=gray,thin,rounded corners=2pt}}
\begin{tikzpicture}[scale=.6,baseline=(node22.base)]
    \small
    \draw[factor] (-.45, -.6) rectangle (.45, 2.6) {};
    \draw[factor] (.55, -.6) rectangle (1.45, 2.6) {};
    \draw[factor] (1.55, -.6) rectangle (2.45, 2.6) {};
    \draw[factor] (-.6, -.45) rectangle (2.6, .45) {};
    \draw[factor] (-.6, .55) rectangle (2.6, 1.45) {};
    \draw[factor] (-.6, 1.55) rectangle (2.6, 2.45) {};

    \foreach \x in {0, 1, 2}
    \foreach[evaluate={\z = int(\x+3*\y)}] \y in {0, 1, 2}
    {
\pgfmathparse{\z==1 || \z==3 || \z==8 ? "mylightpurple" : "none"}
\edef\myfill{\pgfmathresult}
\node(node\x\y)[circ,fill=\myfill] at (\x,\y) {};
}
\end{tikzpicture}}}
    &
		$\cP = \{ \textsf{i} \leftrightarrow \textsf{j} \} = A \times B$,
		valid structures are exact one-to-one assignments.
		\newline\newline
		$|\cZ| = |A|! = |B|!$
		\\ \bottomrule
\end{tabular}
\caption{\label{fig:structures}Illustration of a selection of useful structures, along with their matrix
representation.}
\end{figure}

The role of structure prediction is to provide a \textbf{prediction rule} for
selecting a best structure, and, if possible, a \textbf{probability distribution}
over all possible structures. In the latter case, we would have a
\emph{probabilistic} structure prediction model.
This is challenging because the total number of possible structures can grow
very quickly (often exponentially) with the size of the problem, see~\cref{fig:structures}.
Given a specified probabilistic structure prediction model $\Pr(\z \mid \x)$,
in this book, we consider three challenging problems, which will serve as
\textbf{computational building blocks} for learning and prediction.
\begin{itemize}
\item \textbf{Maximization:} find $\argmax_{\z \in \cZ} \Pr(\z \mid \x)$;
\item \textbf{Expectation:} compute $\bbE_{\var{Z} \sim \Pr(\var{Z}\mid\x)}\big[h(\var{Z})\big]$ for
some function $h$;
\item \textbf{Sampling:} draw a stochastic $\z \sim \Pr(\var{Z} \mid \x)$.
\end{itemize}
These building blocks are well studied in the structured output prediction
setting, which we review in this chapter. However, in the next chapters
we will see that structured latent variable learning also builds upon the same
building blocks. Moreover,
for latent variable learning, it is usually not enough to just have access to
these methods, but we also need to be able to compute or approximate their gradients,
with respect to some parameters of the distribution; this is sometimes called
``differentiating over the structure'' and is the main topic of
\cref{sec:struct-relax}.

In this chapter, we review the two main strategies classically employed
for structure prediction: incremental prediction and global prediction.
Both of them rely on breaking down
structures into the constituent parts, but they differ in the trade-offs between
representation power and optimality.

\section{Incremental Prediction}

To model a structured variable $\z$, possibly conditioned on some input $\x$,
we may employ a probabilistic model $\Pr(\z \mid \x)$, which must, of course, obey
the global constraints (\ie, it must assign zero probability to invalid
configurations). The set of possible configurations $\cZ$ is discrete and
finite. Note that $\cZ$ may depend on $\x$ --- \eg, for part-of-speech tagging,
the length of a sentence $|\x|$ gives the number of tags in the output sequence.

The probability chain rule gives a way to factor the distribution of $\z$ without any loss of expressiveness:
\begin{equation}\label{eqn:increm}
\Pr(\z \mid \x) =
\prod_{i=1}^{|\mathcal{P}|}
\Pr(z_i \mid z_1, \ldots, z_{i-1}, \x)
\end{equation}
This decomposition suggests a method for building up a structure by \emph{incrementally
adding parts}.

\paragraph{Connection to multi-label classification.}
The multi-label setting is a particular instance of unconstrained structure prediction
(\cref{ex:unconstrained}). In this context, incremental prediction is known as
the  \emph{classifier chain} method \citep{read2011classifier},
which amounts to training $|\cP|$ binary classifiers applied in
sequence, each classifier having access to the already-predicted labels.

\paragraph{Blockwise prediction.}

A commonly-occurring pattern involves predicting multiple
interdependent \emph{one-of-K} choices (\cref{ex:oneofk}).
If a structure consists of $L$ assignments, each one with $K_i$ choices, for $i
= 1,\ldots,L$, then
the vector $\z$ then can be seen as a concatenation
\[ \z = [\z_1, \ldots, \z_{L}]\, \]
where each $\z_i \in \{\bm{e}_1, \ldots, \bm{e}_{K_i}\}$ is an indicator vector,
so $\cZ = \{\bm{e}_1, \ldots, \bm{e}_{K_1}\} \times \ldots \times \{\bm{e}_1, \ldots, \bm{e}_{K_L}\}$.
The dimensionality of $\z$ is thus $|\cP| = \sum_{i=1}^L K_i$, and the number of
possible configurations is $|\cZ| = \prod_{i=1}^L K_i$.
We may incrementally predict $\z$ block-wise, using
\begin{equation}\label{eqn:block-increm}
\Pr(\z \mid \x) = \prod_{i=1}^L \Pr(\z_i \mid \z_1, \ldots, \z_{i-1}, \x)\,.
\end{equation}
The only difference from \cref{eqn:increm} is that this time we have a chain of
\emph{multi-class} classifiers rather than binary classifiers.

\begin{myexample}[Incremental part-of-speech tagging.]
Here, the input $\x$ is a sentence of length $L$, and we seek to generate a sequence $\z$ consisting of
$L$ one-of-K assignments: for instance, the sentence ``the dog barked''
should be tagged as \textsf{determiner, noun, verb}. The possible tags are the
same for each word, so $K_i = K$, so $|\cP|=LK$ and
$|\cZ|=K^L$.
\end{myexample}

Despite its simplicity, this paradigm has gained popularity with the rise of
neural networks. A particularly powerful extension is to allow variable-length
target structures,
$\cZ = \bigcup_{L \in \mathbb{N}} \{\bm{e}_1, \ldots, \bm{e}_K\}^L$.
This can be achieved by designating a specific variable assignment as a
``stop'' control token and imposing that $\Pr(\z_i \mid \z_{i-1}=\textsf{STOP})=0$.
The next example illustrates this generalization and underscores many successful
recent models in NLP.

\begin{myexample}[Conditional language generation and seq-to-seq]
Sentences in natural language can be seen as a sequence of discrete tokens (\eg,
words). We may incrementally build up a sentence by appending tokens selected from a
\emph{vocabulary} of \(K\) words, mapped to \(\cZ=\{\mbe_1, \ldots, \mbe_K\}\),
one of which is the special \textsf{STOP} token.

Let $\bm{V} \in \mathcal{R}^{d \times K}$ denote a matrix of token embeddings,
and a neural network \(\bm{h}(\z_1, \ldots, \z_{i-1}, \x, \param) \in \RR^d\).
We can parametrize the probability of the next word as
\[\Pr(\bm{z}_i \mid \z_1, \ldots, \z_{i-1}, \x) \propto \exp
f(x, \z_1, \ldots, \z_{i-1}, \z_i)
\,,\]
where
\[
f(x, \z_1, \ldots, \z_i) \defeq \DP{\bm{z}_i}{\bm{V}^\top\bm{h}(\z_1, \ldots, \z_{i-1}, \x)}
\,.
\]
This is a neural conditional language generation model, where $\x$ represents an
input to be converted to a sentence. For image caption generation, $\x$ is an
image; for speech-to-text transcription, $\x$ is a sound, and for translation,
$\x$ is a sentence in a foreign language. When $\x$ is a sequence, such models
are often called sequence-to-sequence or seq-to-seq.
The parametrization of $\bm{h}$ is crucial as it needs to
handle variable-length input. Common
choices include recurrent networks and transformers
\citep{seq2seq_sutskever,seq2seq_cho,transformers}. The parameter sharing
between predictors at different positions allows the models to generalize,
including to variable-length sequences.
This strategy is sometimes called
\emph{autoregressive} generation, drawing parallels with time series analysis.
\end{myexample}

\paragraph{Transition-based systems, constraints, and automata.}
The extension of incremental prediction to more complex and constrained
structured problems is enabled by transition-based encodings.
Here, the assignments $\z_i$ are interpreted as \emph{actions} or
\emph{transitions}. For instance, the path of a robot through a grid can be
seen as $L$ steps, or actions, chosen among $\{\textsf{up}, \textsf{down}, \textsf{left}, \textsf{right}\}$.
Incremental prediction is popular in parsing, where it is known as
transition-based parsing, action-based parsers, or shift-reduce parsing.
For parsing, alignments, and other similar more complex structures,
the set of admissible actions may be constrained based on the actions already
taken.

In general, an arbitrary constraint $h(\z)=0$ cannot be handled by
incremental prediction, because there may be no admissible actions left.
Any finer-grained early detection of a violation is only possible for certain
forms of constraints, and for well-chosen prediction orderings.
Fortunately, a principled way to incorporate constraints into incremental
prediction is provided by automata theory.
By building an automaton with
accepting set equal to the desired feasible set, it is possible to
detect wrong paths early.

The next example is a celebrated construction from syntactic parsing of natural
language sentences.

\begin{myexample}[Shift-reduce parsing]
One flavor of syntactic analysis of language is provided by constituency
parsing, where a sentence is recursively divided into smaller phrasal units
(\cref{fig:structs}).
A constituency parse can be represented by a binary tree. We may build up this
tree by choosing between two actions \citep{sagae2005shiftreduce}:
\begin{itemize}
\item {S}hift: add a new leaf node;
\item {R}educe: merge the two most recently produced nodes.
\end{itemize}
The process, shown in~\cref{table:shiftreduce},
corresponds to an encoding of candidate binary parse trees for a sentence of length $N$
as bit vectors of length $d=2N-1$, with $z_i=1$ denoting a \emph{shift} and $z_i=0$
a \emph{reduce}.
In order to represent an admissible tree, there are non-trivial constraints.
Firstly,
reducing is only allowed when there are two nodes available for merging.
Equivalently,
up to any position $k$, the shifts must strictly outnumber the reduces, or,
$\sum_{i=1}^{k} z_i > k/2$.
In particular, this means the first two actions must always be shifts.
Secondly, exactly $N$ shift operations must be performed, since all words must
be included in the tree. This can be written as
$\sum_{i=1}^{d} z_i = N$.

The first constraint is naturally phrased in a way that can be checked at every
iteration, to decide whether a reduce action is allowed. It is thus already in
an FSA form. The second constraint appears to depend
on the entire sequence, so we would have to defer its checking until the end.
However, if we incrementally satisfy the first constraint, we can never have
$\sum_i z_i < N$, so we only have to watch out, incrementally, whether we
already ``used up'' all of the N shifts. If this happens, the algorithm must
terminate, filling up the remaining structure with reduce actions.
\end{myexample}

\begin{table}
	\caption{Example of incremental construction of vector $\z$ by a shift-reduce parser. Action \texttt S (respectively \texttt R) corresponds to a shift (respectively reduce). \label{table:shiftreduce}}
	\small
	\begin{tabular}{llrl}
		Action&$\z$&Node stack&Input buffer\\
		\hline
		&&& Sleep the clock around \\
		\texttt S & $[1]$ & Sleep &the clock around \\
		\texttt S & $[1~1]$ & Sleep the & clock around \\
		\texttt S & $[1~1~1]$ & Sleep the clock & around \\
		\texttt R & $[1~1~1~0]$ & Sleep (the clock) & around \\
		\texttt S & $[1~1~1~0~1]$ & Sleep (the clock) around &  \\
		\texttt R & $[1~1~1~0~1~0]$ & Sleep ((the clock) around) &  \\
		\texttt R & $[1~1~1~0~1~0~0]$ & (Sleep ((the clock) around)) &
	\end{tabular}
\end{table}

\begin{figure}
\includegraphics[width=.5\textwidth]{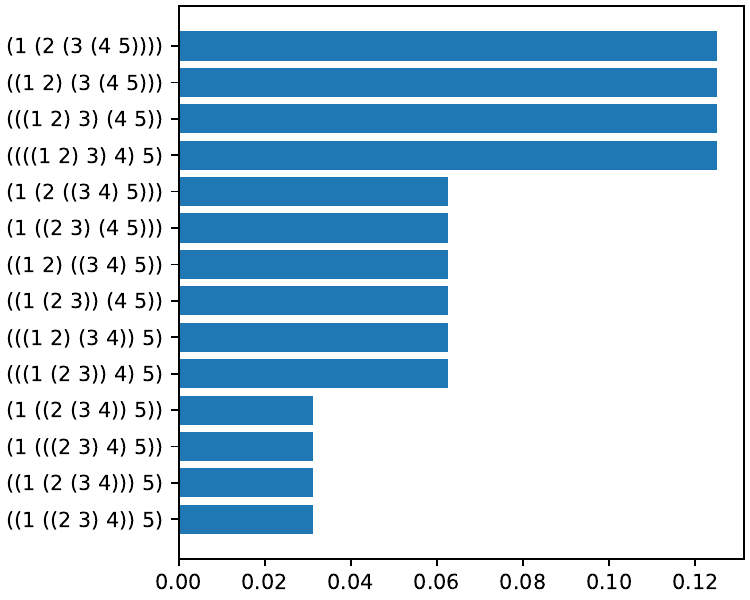}
\includegraphics[width=.5\textwidth]{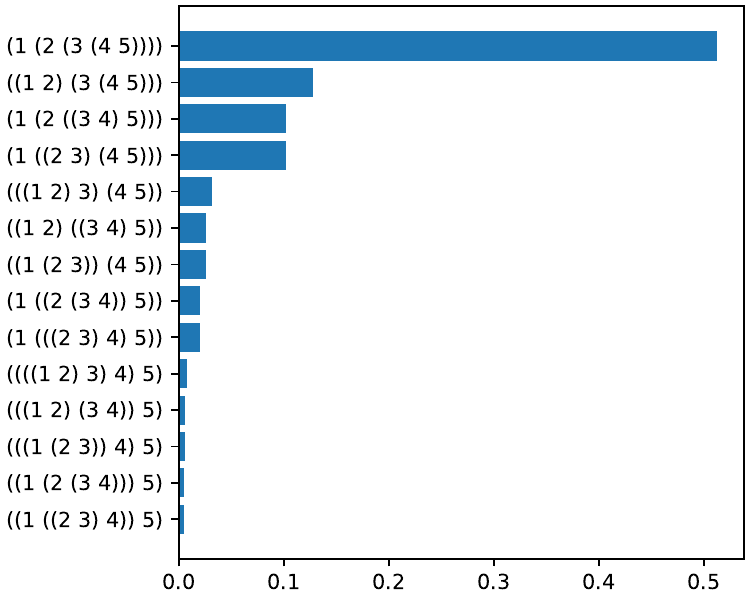}
\caption{\label{fig:shift-reduce-skew}Shift-reduce probabilities for binary trees, with shift probability
$.5$ (left) and $.8$ (right) modulo constraints.}
\end{figure}

We note that the choice of transition system for a structured problem is not
unique, and moreover, every choice may introduce specific subtle biases.
We highlight the bias intrinsic to the shift-reduce model
\citep{drozdov_coadapt}: consider an ``uninformed'' shift-reduce model that assigns
to $\Pr(z_k \mid z_{1:z-1})$
a uniform distribution over the admissible choices at that point.
One might mistakenly expect this to correspond to a uniform distribution over
all possible binary trees, but this is not the case:
the uninformed shift-reduce model prefers highly skewed structures.
If instead of uniform probability we introduce a slight preference to shifts,
this skew is accentuated, as demonstrated in \cref{fig:shift-reduce-skew}.

A further complication posed by transition-based systems is that, in some
models, multiple sequences of transitions can lead to the same target structure.
This phenomenon is known as \emph{spurious ambiguity} \citep{cohen-spurious}.

\subsection{Computation.}

We next describe the algorithms available for maximization, sampling, and
computing expectations, in incremental structure models such as transition-based
models.

\begin{algorithm}[t]
\caption{\label{alg:beam}%
	Beam search for conditional language generation and seq-to-seq.
	Note that, in practice, it is usual to have maximum sequence length to assert that the algorithm will eventually terminate even if not all sequences in the beam reached the \textsf{STOP} token.
	Moreover, the top-k computation usually renormalizes scores by hypothesis lengths.
}
\small
\begin{algorithmic}
	\LeftComment{Approximate maximization using a beam of size $k$}
	\Function{BeamSearch}{$k, g, \x$}
		\State $B = \{ \emptyset \}$
		\Repeat
			\State $B' = \emptyset$
			\For{$[\z_1 ... \z_i] \in B$}
				\If{$\z_i = \textsf{STOP}$}
					\LeftComment{Don't expand sequences whose last token is \textsf{STOP}}
					\State $B' \gets B' \cup \{ [\z_1, ..., \z_i] \}$
				\Else
					\State $\bm{w} = \bm{V}^\top\bm{h}(\z_1, \ldots, \z_{i}, \x)$
					\For{$\z_{i+1} \in \{\e_1, ..., \e_K\}$}
						\State $\operatorname{score}( [\z_1, ..., \z_{i+1}])
\gets \operatorname{score}([\z_1, ..., \z_{i}]) + \log \frac{\exp \DP{\z_{i+1}}{\bm{w}} }{\sum_i \exp w_i}$
						\State $B' \gets B' \cup \{ [\z_1, ..., \z_{i+1}] \}$
					\EndFor
				\EndIf
			\EndFor
			\LeftComment{Compute the $k$-best sequences in $B'$}
			\State $B = \operatorname{top}_{(k)}(\operatorname{score}, B')$
		\Until{$\forall [\z_1 ... \z_i] \in B: \z_i = \textsf{STOP}$}
	\EndFunction
\end{algorithmic}
\end{algorithm}

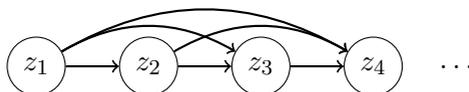
\begin{figure}\centering%

\begin{tikzpicture}[scale=2]
\node[circle,draw] at (0, 0) (a) {$z_1$};
\node[circle,draw,right=20pt of a] (b) {$z_2$};
\node[circle,draw,right=20pt of b] (c) {$z_3$};
\node[circle,draw,right=20pt of c] (d) {$z_4$};
\node[right=10pt of d] (e) {$\ldots$};
\draw[thick,->] (a) -- (b);
\draw[thick,->] (a) to[bend left] (c);
\draw[thick,->] (a) to[bend left] (d);
\draw[thick,->] (b) -- (c);
\draw[thick,->] (b) to[bend left] (d);
\draw[thick,->] (c) -- (d);
\end{tikzpicture}
\caption{\label{fig:incgm}The graphical model of incremental prediction is a fully
connected acyclic graph. The combinatorial explosion challenge
is still present, but delegated to the incremental
predictors
$\Pr(z_k \mid z_1, \ldots, z_{k-1})$.}
\end{figure}

\paragraph{Maximization.}
The reformulation in \cref{eqn:increm} does not change the fact that there are
$\mathcal{O}(2^\szP)$ possible paths to consider.
As such, exactly finding the highest-probability structure
is not feasible in general.
A popular heuristic algorithm is the \emph{beam search} algorithm, which keeps
track of the $k$-best partial prefixes at each time step and seeks to expand
them (see~\cref{alg:beam}). The particular case of $k=1$ is known as
\emph{greedy search} and
corresponds to directly choosing the most likely transition at each step.
The suboptimality of such heuristics is apparent:
if the optimal structure starts with an action that seemed unlikely early on,
the optimal structure cannot be recovered.
The chosen ordering therefore also
may impact performance.
As such, even if beam search has a long history \citep{harpyspeech}, it is still an active research area \citep[amongst others]{pmlr-v97-kool19a,meister-etal-2020-best,meister-etal-2020-beam,meister-etal-2021-conditional,kasai-etal-2024-call}.
If further factorization assumptions are imposed on the transition model, exact
maximization is possible; this is the scenario considered in
\cref{sec:global-str}.

\paragraph{Sampling.} The incremental parametrization of $\Pr(\z \mid \x)$ as in
\cref{eqn:increm} can be seen as a dense graphical model, illustrated in
\cref{fig:incgm}. This suggests we may conditionally sample from this distribution by
sampling $z_1, \ldots, z_\szP$ sequentially: this is known as ancestral
sampling.
Moreover, samples \emph{without replacement} can be drawn using a \emph{stochastic
beam} algorithm \citep{kool-jmlr}.
Algorithms have been proposed to perform approximate maximization by way of
sampling \citep[\eg,][]{keith-etal-2018-monte}.

\paragraph{Expectations.} For the same considerations as maximization,
it is in general not computationally feasible to compute expectations exactly.
However, since exact sampling is available, Monte Carlo methods provide
a good approximation. Given samples $\z^{(1)}, \ldots, \z^{(s)}$, we have
\[
\sum_{\z \in \cZ} h(\z) \Pr(\z\mid\x)
\approx
\frac{1}{s} \sum_{i=1}^{s} h(\z^{(i)})\,,
\]
for any function $h$.

\section{Global Prediction}\label{sec:global-str}

In order to build structured models with global optimality guarantees, and thus
open an avenue toward exact maximization and expectations, we will need to
sacrifice some expressiveness.

Since global prediction models are not always probabilistic, we will consider,
for more generality,
the \emph{score} of a structure
$f : \cZ \times \cX \to \RR$,
with $f(\z, \x; \zparam)$
measuring the compatibility of the structure
represented by $\z$ with the input (context) $\x$.
We may then use the induced Gibbs probability distribution,
\begin{equation}
\Pr(z \mid x) \propto \exp f(\z,\x; \zparam)\,,
\end{equation}
but other distributions are also possible, and will be discussed later on.

For global prediction, we will make the following assumption.
\begin{assumption}[Decomposition into parts]
\label{assumption:parts}
We assume that the score of a
structure can additively decompose as:
\begin{equation}
\begin{aligned}
f(\x,\z, \zparam)
&= \sum_{\substack{p \in \cP \\ z_p = 1}} [\s(\x,\zparam)]_p \\
&= \DP{\z}{\s(\x; \zparam)}
\,,
\end{aligned}
\end{equation}
where $\s(~\cdot~, \zparam) \in \RR^{|\cP|}$ is a vector
of scores for each possible part, and the score of the structure $\z$ is
the sum of scores of the parts that make it up.
\end{assumption}

The additive decomposition of the score turns into a multiplicative
factorization of the probability, as \(\Pr(z \mid x) \propto \prod_{p \in P} z_p
\exp [\s(\x,\zparam)]_p\). For this reason, \cref{assumption:parts} is sometimes
called a \emph{factorization} assumption. Since the probabilistic interpretation
is not necessary, we avoid this term. As we shall see, this assumption
removes some but not all computational limitations.

\begin{myexample}[Arc-factored dependency parsing]
For the non-projective dependency tree example, the parts
$\cP$ are the possible arcs $\mathsf{h} \to \mathsf{m}$.
Following \citet{kg}, consider encoding each word in a sentence of length $L$ with a neural
network (\eg, a bidirectional LSTM) with parameters $\widetilde{\zparam}$,
 giving vectors $\{ \bm{v}_j : j = 1, \ldots,
L \}.$ We may parametrize an arc-factored dependency parser as
\begin{equation}
[\s(\x)]_{\mathsf{h} \to \mathsf{m}}
= \DP{\bm{w}}{\tanh\big([\bm{v}_{\mathsf{h}}; \bm{v}_{\mathsf{m}}]\big)}\,.
\end{equation}
The parameters of the arc-factored parser are thus $\zparam = \{\bm{w}\} \cup \tilde{\zparam}$.
\end{myexample}

\begin{myexample}[Markov sequence tagging]\label{ex:markovseq}
For sequence tagging
\citep{lafferty2001conditional},
given a sentence of length $L$ and a tag set
of size $T$, we may build a globally-optimal model by imposing a Markov
assumption, \ie, scoring \emph{pairs} of consecutive assignments.
We may encode this by using \emph{bigram parts}:
$\cP = \{ (i, t, t') : 1 \leq i < L, 1 \leq t, t' \leq T \}$ where $t$ and
$t'$ are consecutive tags at position $i$ and $i+1$
(subject to a specific handling for the last word transition).
One way to parametrize this model  is
\begin{equation}
    [\s(\x)]_{i, t, t'}
    = \DP{\mbv_i}{\mbw_t} + b_{t,t'}\,,
\end{equation}
where $\mbv_i$ are vector representations for each word (\eg, from some neural
net module), matrix $\mbW \in \RR^{T \times H}$ is an ``output layer''
containing embeddings of each tag, and $\mbB \in \RR^{T \times T}$ is a pairwise
matrix of transition scores, \ie, $b_{\textsf{det}, \textsf{noun}}$ is a bonus
reward for tagging a determiner right before a noun.
The parameters are thus $\zparam = \{\bm{W}, \bm{B} \} \cup \widetilde{\zparam}$.
\end{myexample}

\subsection{The Marginal Polytope}
The decomposition assumption allows us to recast structure prediction problems
as optimization over
the convex hull of possible structures
\begin{equation}\label{eqn:margpoly}
\cM = \conv(\cZ)\,.
\end{equation}
This set is often called the \emph{marginal polytope} \citep{wainwright}.
In a sense, the simplest possible marginal polytope is the convex hull of the
one-of-K indicator vectors $\cZ = \{\bm{e}_1, \ldots, \bm{e}_K\}$, which
is the simplex $\simplex_K$.
For any other finite $\cZ = \{\mbz_{1}, \ldots, \mbz_{|\cZ|}\}$ the marginal
polytope can be written as a linear transformation of the $|\cZ|$-simplex,%
\footnote{Some authors call every such convex hull a $|\cZ|-$simplex, and
$\simplex_{|\cZ|}$ the \emph{canonical} or \emph{probability} simplex.}
\begin{equation}
\conv(\cZ) =
\left\{ \sum_{i=1}^{|\cZ|} \alpha_i \mbz_i : \bm{\alpha} \in \simplex_K \right\}\,.
\end{equation}
The (possibly impractically long) vectors $\bm{\alpha} \in \simplex_{|\cZ|}$
can be interpreted as distributions over the set of structures $\cZ$,
so any point $\zsurr \in \conv(\cZ)$ can be
seen as an \emph{expected structure}:
\begin{equation}
\zsurr \in \conv(\cZ)  %
\quad \iff \quad
\zsurr = \bbE_{\var{Z} \sim \bm{\alpha}} \big[ \var{Z} \big]
\,,
\end{equation}
for at least one $\bm{\alpha} \in \simplex_{|\cZ|}$.

\subsection{Computation}

We describe here the computational challenges of global structure prediction. A
summary of key algorithms for structures of interest is presented in
\cref{tab:max_marg_samp_examples}.
\paragraph{Maximization.}
Finding the highest-scoring structure, \ie, solving
$\argmax_{z \in \cZ} f(z; x)$,
is a difficult combinatorial optimization problem.
But, for many specific problems of interest, specialized, case-by-case algorithms are
available.
For  example, for sequence tagging, the Viterbi algorithm retrieves the global optimum \citep{forney1973viterbi}.
Under \cref{assumption:parts}, finding the highest-scoring structure is equivalent to
maximizing a linear function over a polytope,
\begin{equation}
\max_{\z \in \cZ} f(\z;\x)
= \max_{\z \in \conv(\cZ)} \DP{\z}{\s(\x)}\,,
\end{equation}
because the maximum of a linear function over a polyhedral domain is always
attained at a vertex (a fact that will be revisited in \cref{chapter:relax}). For this reason,
general structured maximization can thus be tackled via linear
programming, but describing $\conv(\cZ)$
may require exponentially many constraints. One strategy for approaching complex
structured problems is therefore developing compact constrained formulations
\citep{punyakanok2004semantic,roth2005integer,andre-concise},
either exact or approximate.
Problems involving intersections of tractable substructures can be approached via
\emph{dual decomposition}, an approach that tackles an outer relaxation of the
linear programming formulation
\citep{rush,komodakisrc,ad3}.

\paragraph{Expectations.} Under a parts-factored Gibbs distribution, computing
$ \zsurr = \bbE[ \var{Z} ] $
is known as ``marginal inference'', because each coordinate of the result is the
marginal probability of a part being selected: $[\zsurr]_r = \Pr(\var{Z}_r = 1
\mid \x)$.
Like maximization, this is intractable in general, but
efficient exact algorithms are available for many structures of
interest; summarized in \cref{tab:max_marg_samp_examples}.
Even when a marginal inference algorithm is available, computing general expectations
$\bbE[h(\var{Z})]$ is not necessarily possible unless $h$ is affine.

Marginal inference is also the gradient of the log-normalizing constant
of the Gibbs distribution with respect to the factorized scores \citep{wainwright,eisner2016tutorial}:
\begin{equation}\label{eq:grad_logsumexp}
\pd{}{\mbs} \log \sum_{\z \in \cZ} \exp \DP{\z}{\mbs}
= \bbE[\var{Z}]\,.
\end{equation}
Marginal inference is therefore necessary in maximum likelihood learning
for supervised structure prediction, and, as we shall later see, it also
provides ways of learning structured latent variables.

\begin{figure}
\centering\includegraphics[width=.5\textwidth]{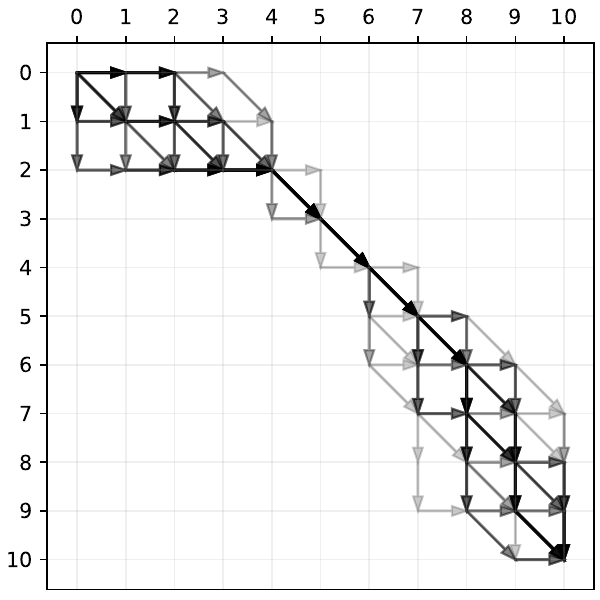}
\caption{\label{fig:nwsamples}Samples drawn from a Needleman-Wunsch alignment
model using a dynamic programming \emph{Forwards Filtering, Backwards Sampling}
strategy.}
\end{figure}

\paragraph{Sampling.} In contrast to incremental models, exact sampling from
$\Pr(\z \mid x)$ is more difficult.
For structures tractable with dynamic programming, we describe an algorithm
in \cref{sec:dag}.
\Cref{fig:nwsamples} illustrates samples drawn using the aforementioned method
from a Needleman-Wunsch model often used in DNA sequence alignment.
For non-projective dependency trees, \citet{zmigrod-etal-2021-efficient} and \citet{stanojevic2022sampling} propose an algorithm involving
repeated calls to marginal inference.
A powerful general framework called \textbf{perturb-and-MAP}
\citep{papandreou2011perturb,hazan2012partition}
attempts to reduce
sampling to
maximization of $ \DP{\z}{\s + \var{U}}$ where $\var{U}$ is a perturbation
drawn from a noise distribution; in general, such samples do not come from the exact Gibbs
distribution, but have proven useful for learning and uncertainty
quantification \citep{caio-iclr,caio-acl}.

\subsection{Structures Encoded as Paths in a
DAG.}\label{sec:dag}

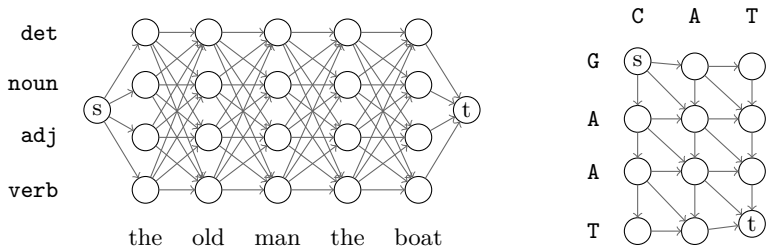
\begin{figure}[t]\footnotesize\centering
\newcommand\pos[1]{\texttt{#1}}
\def\dagcolsep{1.2em}
\begin{tikzpicture}[
    every node/.style={inner sep=0pt,outer sep=0pt},
    nd/.style={circle,draw,minimum size=10pt},
    ed/.style={->,black!60,font=\scriptsize}
]

\matrix[matrix of nodes, column sep=\dagcolsep, row sep=1em,
ampersand replacement=\&,
row 1/.style={nodes={nd}},
row 2/.style={nodes={nd}},
row 3/.style={nodes={nd}},
row 4/.style={nodes={nd}},
] (trellis)
{
{} \& {} \& {} \& {} \& {} \\
{} \& {} \& {} \& {} \& {} \\
{} \& {} \& {} \& {} \& {} \\
{} \& {} \& {} \& {} \& {} \\
the \& old \& man \& the \& boat \\
};
\node[left=3em of trellis-1-1] {\pos{det}};
\node[left=3em of trellis-2-1] {\pos{noun}};
\node[left=3em of trellis-3-1] {\pos{adj}};
\node[left=3em of trellis-4-1] {\pos{verb}};
\node[nd,below left=.25em and \dagcolsep of trellis-2-1] (s) {s};
\node[nd,below right=.25em and \dagcolsep of trellis-2-5] (t) {t};
 \foreach \curr in {2,...,5} {
    \foreach \c in {1,...,4} {
    \foreach \d in {1,...,4} {
        \pgfmathtruncatemacro\prev{\curr - 1}
        \draw[ed] (trellis-\c-\prev) -- (trellis-\d-\curr);
    }
    }
 }

\foreach \c in {1,...,4} {
  \draw[ed] (s) -- (trellis-\c-1);
  \draw[ed] (trellis-\c-5) -- (t);
}
\end{tikzpicture}
\qquad\qquad
\begin{tikzpicture}[
    every node/.style={inner sep=0pt,outer sep=0pt},
    nd/.style={circle,draw,minimum size=10pt},
    ed/.style={->,black!60,font=\scriptsize}
]

\matrix[matrix of nodes, column sep=\dagcolsep, row sep=1em,
ampersand replacement=\&,
row 2/.style={nodes={nd}},
row 3/.style={nodes={nd}},
row 4/.style={nodes={nd}},
row 5/.style={nodes={nd}},
] (trellis)
{
\pos{C} \& \pos{A} \& \pos{T} \\
{s} \& {} \& {} \\
{} \& {} \& {} \\
{} \& {} \& {} \\
{} \& {} \& {t} \\
};
\node[left=1em of trellis-2-1] {\pos{G}};
\node[left=1em of trellis-3-1] {\pos{A}};
\node[left=1em of trellis-4-1] {\pos{A}};
\node[left=1em of trellis-5-1] {\pos{T}};
\foreach \jj in {2,...,3} {
\foreach \ii in {3,...,5} {
    \pgfmathtruncatemacro\iprev{\ii - 1}
    \pgfmathtruncatemacro\jprev{\jj - 1}
    \draw[ed] (trellis-\ii-\jprev) -- (trellis-\ii-\jj);
    \draw[ed] (trellis-\iprev-\jj) -- (trellis-\ii-\jj);
    \draw[ed] (trellis-\iprev-\jprev) -- (trellis-\ii-\jj);
}
}
\draw[ed] (trellis-2-1) -- (trellis-2-2);
\draw[ed] (trellis-2-2) -- (trellis-2-3);
\draw[ed] (trellis-2-1) -- (trellis-3-1);
\draw[ed] (trellis-3-1) -- (trellis-4-1);
\draw[ed] (trellis-4-1) -- (trellis-5-1);
\end{tikzpicture}
\caption{\label{fig:dags}Representing Markov sequence tagging (left) and monotonic sequence
alignment (right) as paths in DAGs.
Edge weights correspond to part scores.}
\end{figure}

A specific case in which global structure prediction is tractable is when
we can represent structures as paths from a source \(s\) to a target \(t\) in a weighted directed acyclic graph (DAG) \(\cG = (\cV,
\cE, w)\); in this case,
all three computational primitives discussed can be computed in time
\(\Theta(|\cV|+|\cE|)\)
using straightforward variants of the Viterbi algorithm
\citep{huang-2008-advanced}.\footnote{In fact,
\citet{huang-2008-advanced} generalize this even further to \emph{hyper}graphs;
with algorithms essentially unchanged. We present the simpler case for clarity.}
\Cref{fig:dags}
shows how we may represent Markov sequence tagging
\citep{lafferty2001conditional}
and monotonic sequence alignments \citep{vintsyuk1968speech,needleman,wagner}
as paths in a DAG. The edges in the DAG correspond to parts of the structure,
and the edge weights to the part scores; therefore the total weight of a path
is the score of a structure.
We give a general form of the algorithms in \cref{alg:dag}.
Such algorithms are instances of dynamic programming
\citep{bellman}
on semirings \citep{mohri}
and were independently (re)discovered a number of times.
The maximization algorithm is credited to
\citet{viterbi},
the algorithm for log-normalization is due to \citet{baum},
and the sampling algorithm, known as \emph{forward filtering, backward
sampling} (FFBS) is due to \citet{ffbs}.

\begin{algorithm}[t]
\caption{\label{alg:dag}Viterbi algorithms for maximization,
log-normalizer,
and sampling paths from \(s\) to \(t\) in a weighted DAG \(\cG=(\cV,\cE,w)\)
\citep{huang-2008-advanced}.
The marginals \(\EE[\var{Z}]\) can be computed by automatic differentiation
of the log-normalizer \cref{eq:grad_logsumexp}.}%
\noindent\hspace{-.2cm}%
\begin{algorithmic}
\Function{Maximize}{$\cG$}
\Comment{Viterbi algorithm}
    \For{\(v \in \operatorname{topologic\_order}(\cV)\)}
    \State \(d_v \gets \displaystyle\max_{u: uv \in \cE} d_u + w(uv)\)
    \State \(p_v \gets \displaystyle\argmax_{u: uv \in \cE} d_u + w(uv)\)
    \EndFor

    \State \(v \gets t; \text{path}=()\)
    \While{\(v \neq s\)}
        \State \(\text{path} \gets p_v v + \text{path}\)
        \State \(v \gets p_v\)
    \EndWhile
    \Return \(d_t, \text{path}\)
    \Comment{\(
d_t = \displaystyle\max_{\mbz\in\cZ} f(\mbz);
\text{path} = \argmax_{\mbz\in\cZ} f(\mbz)
\)}
\EndFunction

\Function{LogNormalizer}{$\cG$}
\Comment{Forward algorithm}
    \For{\(v \in \operatorname{topologic\_order}(\cV)\)}
    \State \(c_v \gets \displaystyle
    \log\sum_{u: uv \in \cE}\exp  d_u + w(uv)\)
    \EndFor
    \Return \(c_t\)
    \Comment{\(
c_t = \displaystyle\log\sum_{\mbz\in\cZ}\exp f(\mbz);
\)}
\EndFunction

\Function{Sample}{$\cG$}
\Comment{FFBS algorithm}
    \For{\(v \in \operatorname{topologic\_order}(\cV)\)}
    \State \(c_v \gets \displaystyle
    \log\sum_{u: uv \in \cE}\exp  d_u + w(uv)\)
    \EndFor

    \State \(v \gets t; \text{path}=()\)
    \While{\(v \neq s\)}
    \State sample \(u\) w.p.
    \(
    \begin{cases}
    \exp( c_u - c_v + w(uv)),& uv \in \cE \\
    0,& uv \not\in \cE \\
    \end{cases}
    \)
    \State \(\text{path} \gets uv + \text{path}\)
    \State \(v \gets u\)
    \EndWhile

    \Return \(\text{path}\)
    \Comment{path is a sample from \(\Pr(\var{Z}=\mbz) \propto \exp f(\mbz)\)}
\EndFunction
\end{algorithmic}
\end{algorithm}

\begin{table}
\caption{%
	Summary of algorithms for maximization and marginal inference,
	for a selection of structures of interest.
	\label{tab:max_marg_samp_examples}%
}
\small
\begin{tabular}{p{2.2cm} p{4cm} p{4cm}}
\toprule
	&\textbf{Maximization} & \textbf{Marginals} \\
\midrule
	\textbf{Sequence tagging}
	& Viterbi \newline \citep{viterbi,Rabiner1989}
	& Forward-Backward \newline\citep{baum,Rabiner1989}
	\\	%
\midrule
	\textbf{Constituency trees}
	& CKY \newline\citep{kasami,younger,cocke}
	& Inside-Outside\newline \citep{insideoutside}%
	\\	%
\midrule
	\textbf{Sequence alignments}
	&
\citep{vintsyuk1968speech,needleman,wagner,dtw}
	& Soft-DTW\newline \citep{softdtw}
	\\	%
\midrule
	\textbf{Non-projective dependency trees}
	& Maximum Spanning Arborescence\newline \citep{Chu1965,Edmonds1967}
	& Matrix-Tree\newline \citep{Kirchhoff1847}
	\\
\midrule
	\textbf{Linear\newline assignment}
	& Kuhn-Munkres \newline \citep{hungarian},\newline
      \citet{lapjv},\newline
      Auction \citep{bertsekas1988auction}
	& intractable, \#P-complete\newline \citep{valiant,garey1979computers,taskar-thesis}
	\\
\bottomrule
\end{tabular}
\end{table}

\section{Summary} %

In this chapter, we have provided the formalisms and algorithms needed for
computations with structures, as studied for predicting structured outputs. We
have covered two very different directions. The first is an incremental one,
allowing expressive parametrizations, at the cost of tractability of
maximization and marginalization.
The second is a global one, limiting the possible parametrization of structure
scores, but resulting often in tractable algorithms.
In the next chapters, we switch gears to latent structure, and we shall see how
the discussed algorithms show up again as the required computational building
blocks for learning and prediction with structured latent representations.

\chapter{Continuous Relaxations}\label{chapter:relax}

\section{Challenges of Deterministic Choices}%
\label{section:challenges_deterministic}

Deterministic latent representations are
a straightforward way to define the mapping from $x$ to a representation
$\hat{\z}$ as a function:
\begin{equation}
\widehat{\z}(x, \zparam)\,.
\end{equation}
Given such a decision rule, the end-to-end downstream model is obtained through
function composition of \(g\) with \(\widehat{\z}\):
\begin{equation}
\widetilde{g}(x,y,\zparam,\yparam) = g(x, y, \widehat{\bm{z}}(x, \zparam) ;
\yparam)\,.
\end{equation}
This strategy
resembles a pipeline, but instead of using an off-the-shelf
frozen model $\hat{\z}(x, \zparam)$ to predict the latent representation,
we will aim to train $\zparam$ and the downstream $\yparam$ together.

Notice that $\widetilde{g}(x,y)$ depends on both $\zparam$ and $\yparam$.
The gradient with respect to the decoder parameters
poses no problems:
\[ \pd{\widetilde{g}}{\yparam}(x,y,\zparam,\yparam)
= \pd{g}{4}(x,y,\widehat{\bm{z}}(x), \yparam)\,. \]
In contrast, the dependency on $\zparam$ is indirect, requiring application of
the chain rule through $\widehat{\bm{z}}$ (\cf \cref{eq:chainrule-def}):
\begin{equation}\label{eq:chainrule}
\pd{\widetilde{g}}{\zparam}(x,y,\zparam,\yparam) =
\underbrace{\pd{\widehat{\bm{z}}}{2}(x, \zparam)}_{%
\text{grad.\ through encoder}
}
\circ
\overbrace{\pd{g}{3}(x, y, \widehat{\bm{z}}(x), \yparam)}^{%
\text{downstream grad.\ \wrt}~\hat{\z}}
\,.
\end{equation}

The breakdown in \cref{eq:chainrule}
highlights that deterministic end-to-end learning has the following
requirements driven by the two terms.
In particular, the following assumption constrains the possible downstream
models $g(x,y,\z,\yparam)$ that can be used with deterministic latents.
\begin{assumption}[Continuous decoder]\label{assumption:relaxed_decoder}
Gradient-based learning of deterministic latent variables requires that:

\begin{enumerate}
\item The latent values $\mbz \in \cZ$ live in a vector space,
\ie, $\cZ \subset \RR^d$.
\item The downstream $g$ accepts weighted averages of structures, \ie, it is defined
at least on $\conv(\cZ)$.
\item $g$ is almost everywhere differentiable w.r.t.\ $\z$.
\end{enumerate}
\end{assumption}
Without this assumption, the downstream gradient $\pd{g}{3}$ cannot be defined:
this partial derivative quantifies the response of $g$ for small changes in its
third argument (\ie, $\z$), but if $\z$ is not encoded as a continuous vector
(and is instead, \eg, a string label), then small changes cannot be quantified.

\paragraph{Differentiating through discrete mappings.}
Even if \cref{assumption:relaxed_decoder} is satisfied,
the first term in \cref{eq:chainrule} may impede learning.
Consider the argmax mapping
\(\zsurr(x,\zparam)\defeq\argmax_{\z \in \cZ} f(x,z,\zparam)\).
If \(\cZ\) is a discrete set, this mapping is discontinuous and flat almost
everywhere. The discontinuity is due to ``jumps'' between discrete values, and
if the changes are small enough to not warrant jumps then
we would get no change at all, \ie,
\(\pd{\widetilde{g}}{\zparam}=\bm{0}\).
There are two main ways to address this challenge.
In this chapter, we consider \textbf{continuous, differentiable deterministic mappings}
$\widehat{\bm{z}}$, allowing exact calculation of \cref{eq:chainrule} for
end-to-end training. In \cref{chapter:surrogate}, we shall see another approach.

To ease into
our construction, consider the rule that assigns the \textbf{average representation}:
\begin{equation}
\hat{\bm{z}}_H(x; \zparam) \defeq \sum_{\bm{z} \in \cZ} \alpha(\bm{z}, x;
\zparam) \bm{z}
\end{equation}
where the weights $\alpha$ are defined as
\begin{equation}
\alpha(\bm{z}, x; \zparam) \defeq \frac{\exp f(\bm{z}, x;
\zparam)}{\sum_{\bm{\zeta \in \cZ}} \exp f(\bm{\zeta}, x, \zparam)}\,.
\end{equation}
The weights $\alpha$ are differentiable in
$\zparam$, and therefore so is $\widehat{\bm{z}}_H$.

This construction is related to the probabilistic latent variable formulation
(\cref{eqn:avg-loss}):
the weights $\alpha$ can be seen as a distribution of $\Pr(\var{Z} \mid x)$,
thus $\widehat{\bm{z}}_H = \EE_{\Pr(\var{Z} \mid x)}[\var{Z}]$ is the ``expected assignment''.
Therefore, this relaxation can be seen as moving the expectation inside the model,
replacing $\EE[g(\var{Z})]$ by $g(\EE[\var{Z}])$. The latter is often much
easier to compute.\footnote{An additional insight is that
if $g$ is linear in $\z$, the two are equal, and if $g$ is convex in $\z$, then the
relaxation yields a lower bound, by Jensen's inequality. However, in
deep learning, $g$ is usually non-convex in $\z$.}

The following examples demonstrate the implication of relaxations.

\begin{myexample}{\bfseries Sigmoid Belief Networks.}
Consider a Sigmoid Belief Network \citep{neal1992gibbs}: a probabilistic model defined
with observed random variables \var{X} (input) and \var{Y} (output)
and latent variables \var{Z} as follows:
\begin{align*}
	\Pr(\var{Z}_i = 1 \mid \mbx) &= \sigmoid(\DP{\mbw_i}{\mbx}), \\
	\Pr(\var{Y}_j = 1 \mid \mbz) &= \sigmoid(\DP{\mbv_j}{\z}).
\end{align*}
where $\var{Y}$ and $\var{Z}$ are vectors concatenating the corresponding
Bernoulli variables.
It is usual to train using the negative log-likelood loss,
\begin{align*}
	L(\mbx, \mby)
	&= - \log \EE_{\Pr(\var{Z}_1,\ldots,\var{Z}_d \mid \mbx)}
\left[ \Pr(\var{Y}=\mby \mid \var{Z}_1, \ldots, \var{Z}_d) \right] \\
\intertext{However, this objective is intractable as the conditional
distribution $\Pr(\var{Y} \mid \var{Z}_1, \ldots,  \var{Z}_d)$ is parameterized by a shallow neural network.
We can approximate this loss function via the following term:}
	&\approx - \log \Pr(\var{Y} \mid \hat{z}_1(\mbx),\ldots,\hat{z}_d(\mbx)) \\
\intertext{where $\zhat_i(x)$ is the expected value of latent variable $\var{Z}_i$:}
	\zhat_i(\mbx) =& \EE_{\var{Z}_i} [\var{Z}_i | \mbx]
	= 1 \cdot \Pr(\var{Z}_i = 1 \mid \mbx) + 0 \cdot \Pr(\var{Z}_i = 0 \mid \mbx) \\
	=& \sigmoid\big(\DP{\mbw_i}{\mbx} \big).
\end{align*}
In other words, the relaxation yields a standard deterministic feed-forward
neural network with one hidden layer with sigmoid activations.
Note that this is different from the mean field theory approach that relies on approximate posterior inference in a generalized EM algorithm \citep{saul1996mft}, and which is therefore highly dependent on the specific parameterization of the distribution $\Pr(\var{Y} \mid \mbz)$.
\end{myexample}

The next two examples illustrate
cases where relaxations cannot be applied because \cref{assumption:relaxed_decoder}
does not hold.

\begin{myexample}{\bfseries Coarse module selection.}
Consider a model with a binary latent variable $\var{Z}$ and a decoder defined as
follows:
\begin{equation}
g(x, y, z) =
\begin{cases}
\operatorname{LSTM}(x, y),& z = 0; \\
\operatorname{transformer}(x, y),& z = 1.\\
\end{cases}
\end{equation}
In other words, $\var{Z}$ determines what architecture to use for prediction.
As given, $g$ is undefined for relaxed $z\in(0,1)$.
In order to relax $g$, we would need to come up with a meaningful interpolation
between LSTMs and transformers, to support evaluation at, \eg, $\hat{z} = 0.5$.
Due to the heterogeneous nature of the two possible modules, no such relaxation
seems any more useful or efficient than explicitly computing $\EE[g(x; y,
\var{Z})]$, which requires running both modules, defeating the entire purpose
of the relaxation approach.
\end{myexample}

\begin{myexample}{\bfseries Program induction.}
Consider the task of inducing latent programs in the form of code,
given expected outcomes $y$ \citep{liang2010learning}, possibly with
natural language descriptions $x$ \citep[\eg,][]{wong2021leveraging}.
Let $\var{Z}$ be a latent program string, and
$g(x,y,\var{Z})$ be computed by compiling and executing $\var{Z}$ as code. As
compilers cannot interpolate between discrete programs, relaxations of $\var{Z}$
are not applicable with this choice of $g$.
\end{myexample}

In this section, we present relaxation methods for latent structure.
All presented methods share
this intuition of returning an average latent choice, including
generalizations toward structured and sparse representations.

\section{Regularized Argmax Operators}\label{sec:regularized_argmax}

If our model were a well-trained pipeline and we had to make an assignment
to the latent variable during the forward pass, a sensible choice is the
highest-scoring one:
\begin{equation}
\begin{aligned}
\hat{\bm{z}}_{0}(x; \zparam) &\defeq \argmax_{\z \in \cZ} f(x, \z, \zparam)\\
                             &=  \argmax_{\z \in \cZ} \DP{\z}{\s(x, \zparam)}\,,
\end{aligned}
\end{equation}
where the second equality comes from \cref{assumption:parts}.
In the sequel, we shall omit the dependency of $\bm{s}$ on $x$ and $\zparam$
and just study the mapping $\bm{s} \to \hat{\z}$.
If we can obtain a continuous, differentiable function $\bm{s} \to \hat{\z}$, then
the composition rule (chain rule) gives us gradients \wrt $\zparam$ by composing
$\zparam \to \bm{s}(x, \zparam) \to \hat{\z}(x, \zparam)$.
We thus zoom in on
\[ \hat{\bm{z}}_{0}(\bm{s}) \coloneqq \argmax_{\z \in \cZ} \DP{\z}{\bm{s}}. \]
The argmax in the right-hand side is a \textbf{discrete} optimization problem,
and, as such, the mapping $\zparam \to \hat{\z}_0$ is discontinuous.
Moreover, this mapping is also almost everywhere flat, because
so is the mapping $\bm{s} \to \hat{\z}_0$. The following example
illustrates this.

\begin{myexample}[Single Coinflip]
Take $\cZ = \{ 0, 1 \}$ and let
$s \in \bbR$. The problem $\hat{z}(s)=\argmax_{z \in \{0, 1\}} zs$ can be solved by
considering the sign of $s$.
If \underline{$s>0$}, then $1s > 0s$ so $z=1$.
If \underline{$s<0$}, then $1s < 0s$ so $z=0$.
If \underline{$s=0$}, we have $1s = 0s$ so either choice for $z$ is optimal.
This means the mapping $\hat{z}(s)$ is not always a function, since at $s=0$
it is multivalued. Even ignoring this pathology,
if we vary $s$ continuously, the maximizer $z$ almost
never changes, except abruptly at $s=0$.
In other words, the derivative
$\pd{\hat{z}}{}(s)$ is zero almost everywhere,
preventing any learning with gradient backpropagation.
\end{myexample}

\Cref{chapter:surrogate} is dedicated to heuristic methods that replace this flat
gradient with an informative (but wrong) one.
Here, instead, we designed the assignment $\hat{\bm{z}}_H$ which circumvents
these issues.

\paragraph{Continuous relaxation.}
The first step in our construction is to apply the result
\begin{equation}\label{eq:relax_continuous}
\begin{aligned}
\hat{\bm{z}}_{0}(\mbs)  &=  \argmax_{\z \in \cZ} \DP{\z}{\bm{s}}\\
&\subseteq \argmax_{\zrel \in \conv(\cZ)} \DP{\zrel}{\bm{s}}\,.
\end{aligned}
\end{equation}
This is a consequence of a theorem of \citet{dantzig}, fundamental to the field of
\emph{linear programming}: the maximum of an optimization problem over a
polytope is always achieved at one of the vertices (but, possibly, at more than one.)
This result lets us relax a discrete problem to a continuous but constrained
one, at the cost of allowing spurious solutions at threshold points.

\addtocounter{myexample}{-1}
\begin{myexample}[Single Coinflip, continued]
Returning to the one-dimensional case, the convex hull of $\{0, 1\}$ is the
closed interval $[0, 1]$. The cases $s>0$ and $s<0$ still yield the optimum
$\zzrel=1$ and $\zzrel=0$ respectively. However, the case
$s=0$ now admits any $\zzrel \in [0, 1]$ as solution. And, promised by
\cref{eq:relax_continuous}, this example verifies $\{0, 1\} \subset [0, 1]$.
\end{myexample}

At first, this might not seem like a step forward, as we made the problem
harder by increasing the search space. However,
the continuous formulation allows us to flexibly ``reconfigure'' the
maximization problem to obtain desired properties, including continuity and
differentiability of $\bm{s} \to \hat{\z}_0$.

\paragraph{Smoothing.}
The second step relies on the insight that the argmax mapping of an optimization
problem with \emph{strongly concave} objective over a convex domain is
a smooth function.
The objective of \cref{eq:relax_continuous} is linear, thus concave
but not strongly so. We then perform a further relaxation by adding a
\textbf{strongly concave regularizer} $G(\zrel)$ to the objective, yielding:
\begin{equation}\label{eqn:relax_smooth}
\hat{\bm{z}}_{G} \coloneqq \argmax_{\zrel \in \conv(\cZ)} \DP{\zrel}{\bm{s}} +
G(\zrel)\,.
\end{equation}
Strong concavity, along with non-emptiness of $\cZ$, ensure
that the maximizer exists and is unique, so
the mapping $\bm{s} \to \hat{\z}_G$ is well-defined, Lipschitz
continuous, and thus differentiable almost everywhere
(\cite{zalinescu}, Corollary 3.5.11; \cite{kakade_jmlr}, Theorem 3).

We first illustrate a popular choice of $H$ in the one-dimensional example
above. Then, in \cref{sec:cat-relax}, we study more relaxations in the
categorical case, followed by a discussion of the structured generalizations in
\cref{sec:struct-relax}.

\addtocounter{myexample}{-1}
\begin{myexample}[Single Coinflip, continued]
Let $H_{FD}(\zzrel) \coloneqq -\zzrel \log \zzrel - (1-\zzrel) \log (1-\zzrel).$
This function is sometimes called the Fermi-Dirac entropy \citep[][Section 3.3]{borwein_lewis_convex_analysis}.
We then have the optimization problem
\[\argmax_{0 \leq \zzrel \leq 1} \zzrel s + H_{FD}(\zzrel)\,.\]

Ignoring the constraints and taking the gradient of the objective w.r.t.\
$\zzrel$
gives the unconstrained optimality condition
$s + \log(1-\zzrel) - \log(\zzrel) = 0$, with unique solution
$\zzrel=(1+\exp(-s))^{-1}=\sigmoid(s)$,
the logistic sigmoid function.
We can now check that for any $s$,
this solution also verifies the constraints, therefore it is correct.
We thus have the relaxation
$\hat{z}_{H_{FD}}(s)=\sigmoid(s)$.
This result agrees with the theoretical guarantee that
$s \to \sigmoid(s)$ is single-valued, continuous, and differentiable
almost everywhere (in this case, it turns out it is differentiable everywhere).
\end{myexample}

\section{Categorical Relaxation and Attention}\label{sec:cat-relax}

Let us now consider the categorical (one-of-K) case,
\ie, $\cZ = \{\bm{e}_1, \ldots, \bm{e}_K\}$,
as in \cref{ex:oneofk},
and thus $\conv(\cZ) = \simplex^K$.

Points $\zzrel \in \simplex^K$ can be interpreted as discrete probability
distributions (categoricals), suggesting that a meaningful regularization might
be the Shannon entropy:
\begin{equation}\label{eq:shannon}
H_1(\zrel) = -\sum_{i=1}^K \zzrel_i \log \zzrel_i\,.
\end{equation}
This function is strongly concave over the simplex, and the resulting relaxation
turns out to correspond to another familiar function:
\begin{equation}\label{eq:softmax}
\hat{\z}_{H_1}(\mbs)
\coloneqq
\argmax_{\zrel \in \simplex^K} \DP{\zrel}{\bm{s}} + H_1(\zrel) \eqqcolon \softmax(\bm{s})\,.
\end{equation}
We can thus regard softmax as a differentiable relaxation of the one-of-K argmax
mapping. Since softmax is defined using the exponential function, the output of
softmax is \textbf{fully dense}: $\hat{\z}_{H_1}$ is elementwise strictly
positive.
In other words, there is no possible score vector $\bm{s} \in \mathbb{R}^K$ such
that $[\hat{\z}_{H_1}(\bm{s})]_j = 0$ for some $j$.
And since the vertices are one-hot vectors, this also rules them out:
there is no $\bm{s} \in \mathbb{R}^K$ such that $\hat{\z}_{H_1}(\bm{s}) \in \cZ$.

Starting from this motivation, \citet{sparsemax} proposed \emph{sparsemax}
as a sparse alternative to softmax.
It corresponds to regularizing the argmax with the Gini entropy
\begin{equation}\label{eq:gini}
H_2(\zrel) = \frac{1}{2} \sum_{i=1}^K \zzrel_i (1-\zzrel_i) =
-\frac{1}{2}\|\zrel\|^2 + \frac{1}{2}\,,
\end{equation}
where the rightmost equality uses $\sum_i \zzrel_i = 1$. Thus,
\begin{equation}\label{eq:sparsemax}
\begin{aligned}
\hat{\z}_{H_2}(\mbs)
&\defeq
\argmax_{\zrel \in \simplex^K} \DP{\zrel}{\bm{s}} + H_2(\zrel)\\
&= \argmin_{\zrel \in \simplex^K} \frac{1}{2} \| \bm{s} - \zrel \|^2
\eqqcolon \sparsemax(\bm{s})\,.
\end{aligned}
\end{equation}
Contrary to softmax, sparsemax does not rule out sparse points and can even
return a vertex, provided the difference between the highest and second-highest
score is large enough---this behavior is analogous to the \emph{margin
property} of max-margin losses applied in a latent setting \citep{fylosses}.
Yet, as a construction of the form of \cref{eqn:relax_smooth}, it is
continuous and differentiable almost everywhere.

More generally, \citet{fylosses} identify \emph{generalized entropies} that lead
to similar properties as softmax and sparsemax. We highlight one such useful
construction,  the $\alpha$-entmax family of mappings recovers softmax and
sparsemax, and arises from regularization with the Tsallis entropy,
\begin{equation}
H_\alpha(\zrel) = \sum_i\frac{\zzrel_i - \zzrel_i^\alpha}{\alpha(\alpha-1)}.
\end{equation}
Solutions can be written in a thresholded form,
\begin{equation}
\hat{\z}_{H_\alpha}(\bm{s}) = [(\alpha-1)\bm{s} - \tau\bm{1}]_+^{\nicefrac{1}{\alpha-1}},
\end{equation}
allowing efficient computation via reduction to one-dimensional root finding.
Moreover, for a few $\alpha$ values, there exist exact algorithms
based on sorting or median-finding
algorithms: for $\alpha=2$, such algorithms are known since
\citet{Held1974,Brucker1984} and recently revisited by
\citet{Condat2016,duchi}. For $\alpha=1.5$, a similar quasilinear algorithm was
derived by \citet{sparseseq}. These algorithms rely on closed-form solutions to
systems of linear, and, respectively, quadratic equations and are thus
not readily extensible to other $\alpha$ values.

\begin{algorithm}
\caption{\label{alg:entmax-bisect}%
Computing $\alpha$-entmax by root finding via bisection.
The algorithm finds $\tau$ approximately by identifying its initial bounds based
on the maximum value in $\bm{s}$, then recursively halves the search interval.}
\begin{algorithmic}
\Function{entmax}{$\bm{s} \in \bbR^d, \alpha \in (1, \infty), T \in \bbN$}
\State $\tau_L\gets \alpha \max(\bm{s}) - 1$
\State $\tau_R\gets \alpha \max(\bm{s}) - d^{1-\alpha}$
\Loop~$T$ times
\State $\tau_M \gets (\tau_L + \tau_R)/2$
\State $\tilde{\z} \gets [(\alpha-1)\bm{s} - \tau_M\bm{1}]_+^{\nicefrac{1}{\alpha-1}}$
\If{$\DP{\bm{1}}{\tilde{\z}} < 1$}
\State $\tau_L \gets \tau_M$
\Else
\State $\tau_R \gets \tau_M$
\EndIf
\EndLoop
\State \textbf{return} $\tilde{\z}$
\EndFunction
\end{algorithmic}
\end{algorithm}

\paragraph{Gradients of relaxed mappings.}
Recall that our main motivation for the relaxations discussed so far is to
obtain a differentiable mapping with nontrivial gradient $\pd{\hat{\mbz}(\mbs)}{}$.
For the softmax case, the gradient can be found by routine application of vector
calculus:
\begin{equation}\label{eq:gradsoftmax}
\pd{\hat{\mbz}_{H_1}(\mbs)}{} = \operatorname{diag}(\softmax(\mbs)) -
\softmax(\mbs)\softmax(\mbs)^\top.
\end{equation}
For the general $\alpha$-entmax family,
\citet{sparseseq} show that a (generalized) Jacobian of the mapping
$\bm{s} \to \hat{\z}_{H_\alpha}$ has the expression
\begin{equation}
\pd{\zsurr_{H_\alpha}}{}(\bm{s}) =
\operatorname{diag}(\bm{q}) - \frac{\bm{qq}^\top}{\bm{1}^\top\bm{q}},
~\text{where}~
q_i = \begin{cases}
(\zzsurr_i)^{2-\alpha}, & \zzsurr_i > 0; \\
0, & \text{otherwise}, \\
\end{cases}
\end{equation}
and $\zsurr = \zsurr_{H_\alpha}(\bm{s})$ is shorthand for the solution vector.
When $\alpha=1$ this yields the softmax Jacobian; when
$\alpha>1$ it is a sparse matrix with zeros for all rows and columns outside of
the support.

\citet{correia2019adaptively} further derive derivatives
with respect to $\alpha$ in order to tune this additional parameter via
gradient methods.

\begin{figure}\centering
\includegraphics[width=.6\linewidth]{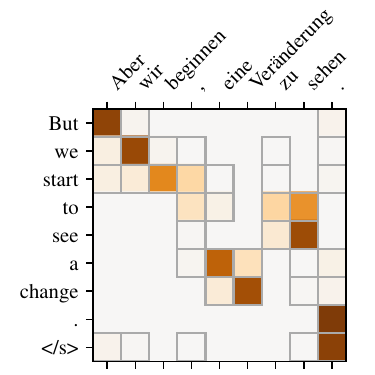}
\caption{\label{fig:sparseattn} Sparse attention in a sequence-to-sequence
machine translation model. Each row is an independent sparse and differentiable
selection $\widehat{\mbz}$ over the German source sentence. Reproduced with permission from \citet{sparseattn}.}
\end{figure}

\begin{myexample}{\bfseries One-of-K and neural
attention.}\label{ex:attn_relax}
Attention mechanisms in deep learning are differentiable components that are
employed for extracting a single contextual representation of a set of $K$ objects (\eg,
words) individually encoded as vectors.
In particular, in the key-value attention formulation \citep{transformers},
each object is encoded by a key vector $\mbk_i$ and a value vector
$\mbv_i$, row-stacked in the matrices $\mbK$ and $\mbV$.
Given a query $\mbq$, attention is
\[ \operatorname{Attn}(\mbq,\mbV,\mbK) = \sum_{i=1}^{K} \alpha_i \mbv_i,\quad
\text{where~} \bm{\alpha}=\softmax\left(\mbK\mbq\right)\,.\]
Consider a deterministic key-value lookup, in which we would simply
retrieve the value corresponding to the key with the highest dot product with the
query. Writing $s_i = \DP{\mbk_i}{\mbq}$,
\[
\operatorname{Lookup}(\mbq,\mbV,\mbK) = \mbv_{i^\star},\quad
\text{where~} i^\star=\argmax_{i=1,\ldots,K} s_i\,.
\]
Using the set of one-of-K indicator vectors $\cZ = \{ \mbe_1,
\ldots, \mbe_K\}$, we have
\[
\operatorname{Lookup}(\mbq,\mbV,\mbK) = \mbV^\top \hat{\mbz}_0,
\text{where~} \hat{\mbz}_0 = \argmax_{\mbz \in \cZ} \DP{\mbz}{\mbs}\,.
\]
Attention can then be seen to be the entropy-regularized relaxation of lookup,
replacing $\hat{\mbz}_0$ with $\hat{\mbz}_{H_1}=\softmax(\mbs)$ as in
\cref{eq:softmax}.
The entropy regularization forces all attention weights to be nonzero. However,
sparse attention using $H_\alpha$, as shown in \cref{fig:sparseattn},
can bridge this gap and induce combinations between only a few objects.
\end{myexample}

\section{Global Structured Relaxations and Structured Attention}\label{sec:struct-relax}

In general structure prediction settings, $\conv(\cZ)$ is a polytope that lacks a
compact description.
There are two ways to generalize softmax and sparsemax--style mappings
in the structured case: \emph{probabilistic regularization}, in which we rely on
the interpretation of $H$ as  an entropy, measuring the flatness of
probability distributions, and \emph{mean regularization}, in which we think of
$H$ just as a regularizer on relaxed marginal assignments of parts.

\paragraph{Probabilistic regularization.}

Consider the structured entropy
\begin{equation}\label{eq:structured_entropy}
H(\zrel) = \sup_{\bm{\alpha}} \left\{ -\sum_i \alpha_i \log \alpha_i : \bm{\alpha} \in
\simplex^{|\cZ|}, \bbE_{\bm{\alpha}}[\var{Z}] = \zrel \right \}\,.
\end{equation}
$H$ measures the highest entropy among
all decompositions of $\zrel$ as convex
combinations of the elements of $\cZ$. If $\zrel$ is a vertex $\z_0$, then the
unique $\bm{\alpha}$ is an indicator vector and $H(\zrel)=0$. If $\zrel$ is at
barycenter of the polytope, the max-entropy decomposition is $\bm{\alpha} = [1/|\cZ|, \ldots,
1/|\cZ|]$, but other decompositions with lower entropy also may exist.
The main motivation for using this regularizer is that the solution
$\bm{z}_{H}$ is none
other than the average structure under the Gibbs distribution,
and computing it is known as the \emph{marginal inference} problem,
which we have seen in \cref{sec:global-str}
as a fundamental primitive for probabilistic structured output prediction,
and now we see it also plays a central role in relaxed learning of latent
structures:
\begin{equation}
\widehat{\z}_{H} = \bbE[\var{Z}]
\quad \text{under} \quad
\Pr(\z\mid x) \propto \exp \DP{\z}{\bm{s}}
\,.
\end{equation}
In the one-of-K model, this recovers softmax, because $\cZ=\{\mbe_1, \ldots,
\mbe_K\}$,
so $H(\zrel)=-\sum_i \zzrel_i \log \zzrel_i$.
More generally, the solution takes the form $\hat{\z}_{H} =
\mbE_{\var{Z}\sim\bm{\alpha}_\star}[\var{Z}]$, where
$\bm{\alpha}_\star$ is the maximizer inside \cref{eq:structured_entropy}.
The distribution $\bm{\alpha}_\star$ cannot usually be computed, or even stored
in memory, however, the mean $\hat{\z}_{H}$ can be computed efficiently for
some select models of interest,
\eg, the ones in \cref{tab:max_marg_samp_examples}.

The marginal \(\hat{\zrel}_H\) is continuously differentiable as a function of
\(\mbs\): this follows from the fact that \(\hat{\zrel}_H\) is itself
the gradient of the normalization constant \(\log\sum_{\z in \cZ} \exp
\DP{\z}{\mbs}\) (\cref{eq:grad_logsumexp}), which is analytic.
The gradient is thus available through automatic differentiation whenever
marginal computation is available: as we have seen from \cref{chapter:struct}
this is the case for global structure models representable as paths in
(hyper)graphs, and for a select number of other global structure models.
The next paragraph surveys research using this type of relaxation.

\paragraph{Structured attention networks.}
\citet{structured_attn} made use of the differentiability of marginal inference
to propose structured attention mechanisms, including an attention mechanism with
first-order Markov potentials to encourage contiguous selection,
as well as a tree-structured attention mechanism demonstrated on arithmetic
expression parsing. Both models are based on differentiating through dynamic
programs \citep{Li2009,arthurdp}.
\citet{lapata} employ a similar strategy to learn tree-structured document
representations, this time using unconstrained arborescences via the matrix-tree
theorem. \citet{liu-etal-2018-structured} develop a structured attention network
for alignment using a variant of an inside-outside parsing algorithm.

\section{Mean Structure Regularization: Sinkhorn and SparseMAP}
\label{sec:meanreg}

In some cases, marginal inference is not available for the structure we
are interested in modeling.
A more tractable regularizer in structured case can be obtained by
applying a penalty directly on the marginal vector $\zrel$, rather than on the
decomposition $\bm{\alpha}$.  While this approach abandons probabilistic
interpretations in terms of maximum entropy, it leads to useful algorithms.

\paragraph{Linear assignment and Sinkhorn.}
For certain structured representation, one may apply an entropy-inspired regularizer onto
the coordinates of $\zrel$ directly:
\begin{equation}
S(\zrel) = -\sum_r \zzrel_r \log \zzrel_r\,.%
\end{equation}
If the representation $\zrel$ consists of a concatenation of one-of-K
variables, this can be seen as regularizing the (independent) entropies of
each of those variables. This coordinate-wise application of entropy
must not be confused with the Shannon entropy
regularizer of \cref{eq:structured_entropy}; in general $S(\zrel) \neq
H(\zrel)$.

The main example for which this construction leads to efficient computation is
linear assignment (matching), introduced in
\cref{ex:matching,fig:structs,fig:structures}, which we now formalize
as a special case of the transportation problem
\citep{hitchcock1941distribution,kanto,compot}.

Given two sets of size $m$ with pairwise affinities between them $s_{i
\leftrightarrow j}$
the task is to
find a maximum-scoring assignment that maps each object in a
set to exactly one counterpart in the other.
Solutions can be represented as \emph{permutation matrices}:
\begin{equation}
\cZ = \{ \mbz = \operatorname{vec}(\mbZ) :
\mbZ \in \{0, 1\}^{m \times m} : \bm{1}^\top \mbZ = \mbZ \bm{1} =
\bm{1} \}\,,
\end{equation}
and the marginal polytope is known as the Birkhoff polytopes of bistochastic
matrices \citep{birkhoff}:
\begin{equation}
\conv(\cZ) = \{ \zrel = \operatorname{vec}(\mbZ):
\mbZ \in \bbR_+^{m \times m}, \bm{1}^\top \mbZ = \mbZ \bm{1} =
\bm{1} \}\,.
\end{equation}
It is possible to generalize this formulation to allow \emph{unbalanced}
assignments between different-sized sets, by replacing the equality constraint
along the larger dimension with an inequality constraint.
Generalizing the row and column sums to distributions yields the transportation
problem.

As per \cref{tab:max_marg_samp_examples},
the maximization problem
\begin{equation}
\argmax_{\mbz \in \cZ} \DP{\mbz}{\mbs}
\end{equation}
can be solved in polynomial time by algorithms such as
Kuhn-Munkres, Jonkers-Volgenant, auction, however, the
corresponding marginal inference problem is \#P-complete
\citep{garey1979computers}.
However, the corresponding mean-regularized problem
\begin{equation}
\argmax_{\zrel \in \conv(\cZ)} \DP{\zrel}{\mbs} + S(\zrel)
\end{equation}
can be efficiently computed by an iterative renormalization algorithm due to
\citet{sinkhorn1964relationship}, popular in optimal transport
\citep{cuturi2013sinkhorn}.

\citet{adams2011ranking} differentiate through Sinkhorn
for learning to rank. For neural latent structures, we highlight
\citet{mena}, who employ Sinkhorn within an approximate
strategy for probabilistic permutations learning
and \citet{tay2020sparse}, who employ Sinkhorn attention for reordering inside
the transformer architecture.

\paragraph{Quadratic regularization: SparseMAP.}

Taking $Q(\zrel) = -\frac{1}{2}\|\zrel\|_2^2$ yields a projection onto the
marginal polytope:
\begin{equation}
\widehat{\bm{z}}_Q =  \argmax_{\zrel \in \conv(\cZ)} \DP{\zrel}{\bm{s}} -
\frac{1}{2}\|\zrel\|_2^2 = \argmin_{\zrel \in \conv(\cZ)} \| \zrel - \bm{s} \|_2^2 \,.
\end{equation}
This quadratic penalty was used by \citet{smooth_and_strong} as a loss
function.

The SparseMAP strategy for structure learning \citep{sparsemap,lp-sparsemap}
relies on this quadratic regularization, along with an efficient
general-purpose algorithm for computing both the solution and its
gradients, and observes that, analogously to sparsemax in the unstructured
case,
solutions tend to be sparse, \ie, consist of combinations of only a few global
structures.
For computing the solution, \citet{sparsemap}
apply the general-purpose \emph{active set} algorithm
\citep[Ch.~16.4 \& 16.5]{vinyes,nocedalwright},
also known as Wolfe's \emph{min-norm point} algorithm,
\citep{mnp}, part of a Frank-Wolfe family of algorithms
\citep{fw,cg} which do not require any specific form of $\conv(\cZ)$ but instead only
interact with the constraint via an \emph{linear oracle}
$\argmax_{\zrel \in \conv(\cZ)} \DP{\zrel}{\mbs}$.
In the context of structured prediction, this linear oracle corresponds to
finding the highest-scoring structure, and thus algorithms are often
well-studied (\cref{tab:max_marg_samp_examples}).
The backward pass has an efficient low-rank structure
that benefits from sparsity.

\begin{figure}
\includegraphics[width=.99\linewidth]{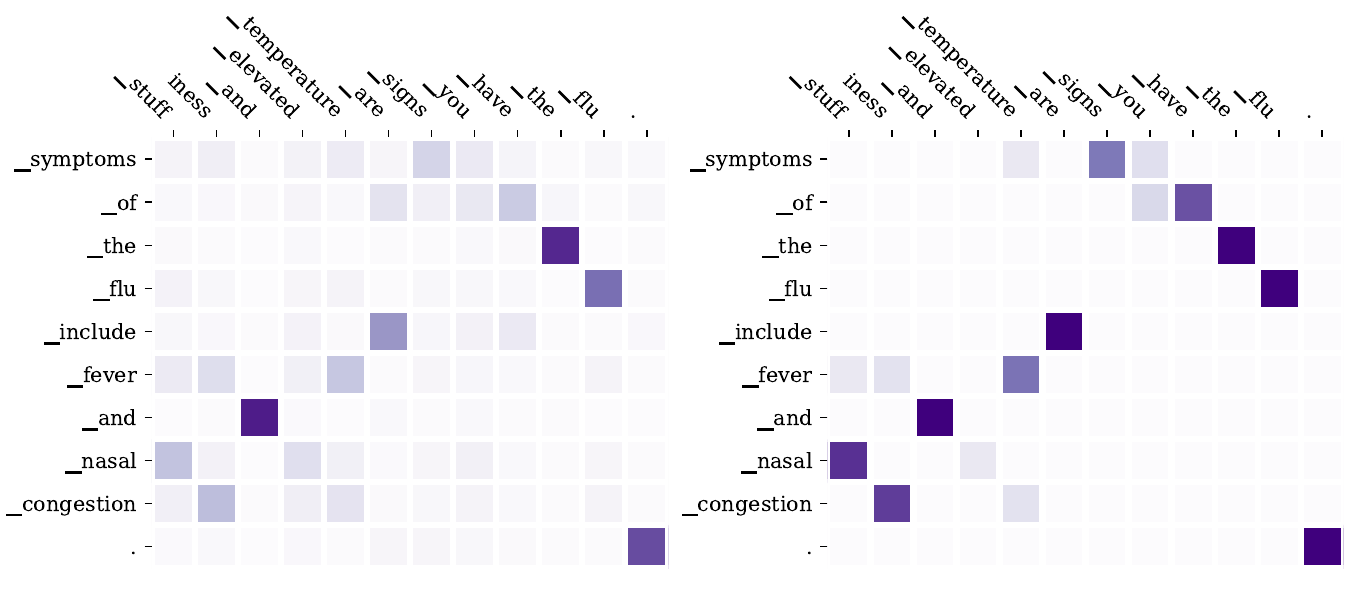}
\caption{\label{fig:matching} Relaxed bipartite matching (linear assignment) between
tokens in two English sentences. Left: Sinkhorn, right: SparseMAP.}
\end{figure}

\begin{myexample}[Dense vs.\ sparse linear assignment]
\Cref{fig:matching} shows ``soft matchings'' between two English paraphrases, as
induced by Sinkhorn and by SparseMAP. In both cases, the pairwise affinity scores
are computed as
\begin{equation}
[\mbs(x, \param_f)]_{i \leftrightarrow j} =
\Big\langle
{[\mbh(x_1; \param_f)]_i},
{[\mbh(x_2; \param_f)]_j}
\Big\rangle\,,
\end{equation}
where $x=(x_1, x_2)$ is the tuple of input sentences, and $h$ is a neural
network encoder; in this case, an \texttt{albert-base} pretrained transformer
\citep{albert}.
Both induced alignments are differentiable. Analogous to the unstructured case,
SparseMAP tends to make ``hard assignments'' for unambiguous variables; in this
case, for identical words.
\end{myexample}

\paragraph{Implicit perturbation-based regularization.}
In some situations, the choice of regularizer itself is not
necessarily important. \citet{berthet2020learning} propose to implicitly induce
a regularized argmax from the smoothing generated by a random noise variable
\(\var{U}\), defining
\begin{equation}\label{eq:berthet_perturb}
 \widehat{\mbz}_{\Omega_{\var{U}}} (\mbs) =
\bbE_{\var{U}}\left[\argmax_{\z\in\cZ} \DP{\mbz}{\mbs +
\var{U}}
\right]\,.
\end{equation}
The expression of \(\Omega_{\var{U}}\) depends on the distribution of
\(\var{U}\) and is not generally available, but nevertheless
\(\zsurr_{\Omega_{\var{U}}}\) and its gradient can be estimated from samples.
Despite the stochastic nature of this method, we catalog it as a relaxation
since it is based on computing \(g(\bbE[\var{Z}])\) and not \(\bbE[g(\var{Z})]\):
the downstream objective is not optimized in expectation.
Like SparseMAP, this strategy only requires  calls to the
maximization oracle $\argmax_{\z \in \cZ} \DP{\z}{\s}$.
Unlike SparseMAP, the maximizations can be performed in parallel,
possibly incurring substantial speed benefits on modern hardware;
in exchange, one loses finite convergence properties and adaptive sparsity (more
samples can always be drawn to improve the estimate).
This strategy is also related to marginal inference, as for
certain structures \(\var{U}\) can be chosen such that
\(\Pr(\zsurr_{\Omega_\var{U}}(\s)) \propto \exp \DP{\z}{\s}\): in the case of the one-of-K
model \(\var{U}\) would have to be the standard Gumbel variable, a connection
discussed in more detail in \cref{chapter:expectation}.

\paragraph{Picking a regularizer.}
When an efficient and numerically stable algorithm for marginal inference
is available, and sparsity of the induced structures is not desired,

marginal inference (\ie, the structured entropy regularizer)
is a good choice.
When only a maximization algorithm is available, or sparsity is desired,
or marginal inference is numerically unstable,
SparseMAP provides a solution.
If not even a maximization algorithm is available,
\citet{ves} and \citet{domke} study
differentiation of approximate marginal inference via belief propagation for
pairwise Markov random fields,
and
\citet{lp-sparsemap} propose an extension of SparseMAP to arbitrary factor
graphs, supporting coarse decomposition when parts of the problem are
tractable.

\section{Summary}

In this chapter we explored ways to relax a discrete variable selection into a
continuous, differentiable mapping, while still retaining some of the desired
properties from the discrete mapping. We have seen how results from convex
analysis allow us to frame this relaxation as a smoothing, and lead to exact
gradient computations.
Examples of techniques falling into this class include attention mechanisms and
their structured counterparts, sparse transformations such as sparsemax and
$\alpha$-entmax, as well as their structured counterparts, such as
Sinkhorn-based methods and SparseMAP, and perturbation-based regularization
methods.

When a relaxation compromise is feasible, due to the model design or the
application scenario, relaxation methods work generally well and are among the
most stable to train, as they are essentially just standard deterministic deep
models. While sparsity can bridge the gap between discrete and continuous and
lead to integral outputs some of the time, in some applications discrete
mappings are strictly necessary. The next two chapters focus on methods that do
not compromise on discreteness.

\definecolor{downstreamcol}{RGB}{192,41,66}
\definecolor{intermcol}{RGB}{83,119,122}
\tikzset{%
    >=latex,
    nd/.style={font=\footnotesize},
    downstream/.style={downstreamcol,very thick},
    interm/.style={intermcol, very thick}
}
\newcommand{\scaffold}{
    \node[nd] at (0, 0) (x) {$\strut \mbx$};
    \node[nd] at (3, 0) (s) {$\strut \mbs(\mbx; \zparam)$};
    \node[nd] at (6, 0) (z) {$\strut \widehat{\mbz}$};
    \node[nd] at (9, 0) (y) {$\strut g(\mbx, \mby, \widehat{\bm{z}}; \yparam)$};

    \draw[->] (x) -- node[midway] (xs) {} (s);
    \draw[->] (s) -- node[midway, above] (am) {\strut \footnotesize argmax} (z);
    \draw[->] (z) -- node[midway] (zy) {} (y);

    \node[above=3pt of zy] (yparam) {$\yparam$};
    \node[below=3pt of xs] (zparam) {$\zparam$};
}

\newcommand{\infobox}[1]{
    \vspace{\baselineskip}
    \setlength{\fboxrule}{1pt}
    \noindent\cfbox{gray}{
    \parbox{\linewidth}{
    #1
    }}
    \vspace{\baselineskip}
}

\newcommand{\zvector}{\bm{v}}

\chapter{Surrogate Gradients}\label{chapter:surrogate}

\section{Straight-Through Gradients}\label{sec:ste}

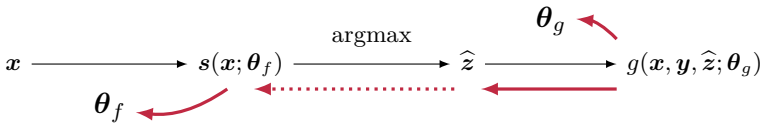
\begin{figure}
\centering
\begin{tikzpicture}[baseline]

\scaffold
\draw[->,downstream] (y.north west) to[bend right=20] (yparam);
\draw[downstream,->] (y.south west) to ([xshift=5pt]z.south);
\draw[downstream,dotted,->] ([xshift=-5pt]z.south) to ([xshift=5pt]s.south);
\draw[downstream,->] ([xshift=-5pt]s.south) to[bend left=20] (zparam);

\end{tikzpicture}
\caption{Selecting a hidden representation \(\widehat{\z}\) using a discrete
argmax introduces a break (dotted line) in the backpropagation chain (thick
purple arrows). The surrogate gradient approach keeps the forward pass
unchanged, but replaces the dotted arrow with a different gradient, such as an
identity function.}
\end{figure}

In \cref{section:challenges_deterministic}, we have seen that
a discrete latent variable assignment
\(\widehat{\z}(x, \zparam)\)
cannot simply be plugged into a downstream model
\[\widetilde{g}(x,y,\zparam,\yparam) = g(x, y, \widehat{\bm{z}}(x, \zparam) ;
\yparam)\,,\]
as the gradient through a discrete encoder (\cref{eq:chainrule}) is
either null or undefined, preventing gradient-based learning.

As before, invoking
\cref{assumption:relaxed_decoder},
we can focus on a model whose encoder
generates factorized scores \(\mbs(\mbx, \zparam)\),
and the discrete variable assignment is
\(
\hat{\bm{z}}_{0}(\bm{s}) = \argmax_{\z \in \cZ} \DP{\z}{\mbs}\).
Instead of relaxing \(\widehat{\z}_0\), as in \cref{chapter:relax}, in this
chapter we keep the discrete encoder in the forward pass,
but in the backward pass we replace its problematic derivative with
a different function.
A possible surrogate gradient is the identity that leads to the following mapping \(\widehat{\z}\) \citep{hinton2012coursera,conditional_bengio}:
 \begin{equation}
 \begin{aligned}\label{eq:surrogategrad}
 \z_\text{ST}(\mbs) & =
 \argmax_{\mbz \in \cZ} \DP{\mbz}{\mbs}\,,\\
 \pd{\z_\text{ST}}{}(\mbs) &\defeq \id \,.
 \end{aligned}
 \end{equation}
That is, the backward pass multiplies by an identity matrix regardless of the
value of \(\mbs\).
This is the origin of the name
``straight-through,'' as during backpropagation the downstream gradient
passes directly through the argmax mapping as if it weren't there.
The second line in \cref{eq:surrogategrad} is not an assertion,
but a construction: we replace the gradient of a function with another function.
From an optimization point of view, this is incorrect, minimizing a
different function than the intended objective.
However, the update can be informative enough for effective learning
in practical applications, and straight-through strategies are easy to
implement.

\section{Straight-Through Variants}\label{sec:st-variants}
\paragraph{Softmax-ST.}

In the unstructured model where \(\cZ=\{\mbe_1, \ldots, \mbe_K\}\),
recall that softmax gives a differentiable relaxation of the argmax operator,
per \cref{sec:cat-relax}.
Therefore, as an alternative of using the identity matrix as the Jacobian in the
backward pass, the derivative of softmax is a sensible alternative,
corresponding to ``pretending'' we had used softmax in the forward pass instead
of argmax \citep[Sec.~2.2]{gumbel_softmax}:
\[
\begin{aligned}\label{eq:softmax-st}
\widehat{\z}_\text{Softmax-ST}(\s) &\defeq \argmax_{\z \in \cZ} \DP{\z}{\s}\,, \\
\pd{\widehat{\mbz}_\text{Softmax-ST}}{}(\s)
&\defeq \pd{\softmax}{}(\s) \\ &= \diag(\softmax(\s)) -
\softmax(\s)\softmax(\s)^\top\,,
\end{aligned}
\]
see \cref{eq:gradsoftmax}.
Since softmax can be seen as a regularized argmax,
this may be interpreted to only turning on the regularization in the backward
pass, and so it may lead
to a smaller mismatch between the forward and backward pass.
In particular, if the maximum value is reached by a single element of $\s$,
$\lim_{\gamma \to 0} \softmax(\s/\gamma) = \widehat{\z}_\text{Softmax-ST}(\s)$.

When tackling latent structured representations, surrogate straight-through
methods fit well within the incremental formalism, as each step in the
prediction is a one-of-K choice.
For learning latent global structure,
a natural extension of Softmax-ST to the structured case is given by the
marginals \(\widehat{\z}_H = \bbE[\var{Z}]\),
since, as discussed in
\cref{sec:struct-relax},
the softmax is the expected vector for the case of the one-of-K structure.
As we know from \cref{chapter:struct},
computing the marginals is an important computational primitive
for structured models, but it is not always tractable.
In cases when marginals are not available, alternative relaxations such as
SparseMAP
(\cref{sec:struct-relax}) can be used, or one may consider the approaches
described next.

\paragraph{Structured Projection of Intermediate Gradients (SPIGOT).}

Motivated by
allowing hard decisions in the hidden layers of neural
networks, while respecting the constraints of a structured $\argmax$ operation
in the backward pass,
SPIGOT \citep{spigot}
computes an intermediate approximation of the gradient that it then projects
onto the marginal polytope:
\begin{equation}\label{eq:spigot}
\begin{aligned}
\widehat{\z}_\text{SPIGOT}(\mbs) &\defeq \widehat{\z}_0(\mbs) = \argmax_{\z \in \cZ} \DP{\z}{\mbs}\,,\\
\pd{\widehat{\z}_\text{SPIGOT}}{}(\mbs)(\zvector)
&\defeq
\widehat{\z}_0(\mbs) - \proj{\conv(\cZ)}(\widehat{\z}_0(\mbs) - \eta\zvector)\,,
\end{aligned}
\end{equation}
where $\eta$ is the step size.
We provide an intuitive explanation of the above update, unifying it with STE,
in \cref{section:pulled_back_interpretation}.
In the work originally proposing SPIGOT, \citet{spigot} compute the projection
approximately using a further relaxation.
\citet{mihaylova-etal-2020-understanding} remark that this projection is the
same computational primitive as the one in the SparseMAP relaxation
(\cref{sec:struct-relax}) and propose to compute it using the active set
algorithm.

\paragraph{Linear Interpolation (LI).}

\citet{diffbb} study a surrogate gradient defined as a linear interpolation of
the downstream value at two points:
\begin{equation}\label{eq:surr-li}
\begin{aligned}
\widehat{\z}_{\text{LI}}(\mbs)&\defeq \widehat{\z}_0(\mbs) \\
\pd{\widehat{\z}_\text{LI}}{}(\mbs)(\zvector)
&\defeq
\frac{1}{\eta} \left(
\widehat{\z}_0(\mbs + \eta\zvector) - \widehat{\z}_0(\mbs)
\right)\,.
\end{aligned}
\end{equation}
Notice that in the limit $\eta\to0$ this recovers the definition of the
directional derivative of $\widehat{z}_0$, which in this case would be
ill-behaved. However, \citet{diffbb} study this mapping in the regime of
substantially larger \(\eta \gg 0\) and argue for its desirable smoothing effect.

Unlike STE and Softmax-STE, SPIGOT and LI result in a derivative that is not
a linear operator, it is therefore not the derivative of any function. For this
reason we must write it as a nonlinear function of the
direction vector \(\zvector\) and not as a Jacobian
(during backpropagation, \(\zvector\) is the downstream
derivative of the loss \wrt \(\widehat{\z}\)).

\paragraph{Implementation Details.}

Surrogate gradient methods, by definition, override the actual gradient
computation with something else. In PyTorch \citep{pytorch}, at present
we recommend an implementation using custom functions (inheriting from
\texttt{torch.autograd.Function}). Such objects implement a
\texttt{forward(self, s)} method which should return \(\widehat{\z}_0(\mbs)\),
and a
\texttt{backward(self, v)} method which returns the vector-Jacobian product at
\(\mbs\) (saved from the forward pass) with \(\mbv\),
\ie, it returns our construction of choice
\(\pd{\widehat{\z}_\text{ST}}{}(\mbs)(\mbv)\),
\(\pd{\widehat{\z}_\text{SPIGOT}}{}(\mbs)(\mbv)\), etc.
Such an implementation cannot be tested using automatic gradient checks, since
the gradient will be technically incorrect. We therefore recommend using custom
unit tests against the theoretically expected answers, \eg, using an identity encoder
\(\mbs(x,\zparam)\defeq\zparam\).

\section{Quantization: Straight-Through Friendly Models}

In certain applications, it may be possible to rearrange the architecture of a
neural network with discrete latent structure so that the
discrete mapping
is in some sense
``close'' to linear, thereby making the STE a
better
choice. While these
approaches require specific model choices, these choices can contain beneficial
inductive biases and perform well.

\subsection{Rounding}

Consider a discrete latent variable whose domain is the integers, \ie, \( \cZ =
\bbZ \defeq \{\ldots, -2, -1, 0, 1, \ldots \}\). Our usual construction that
assigns a score to each integer is infeasible since there are infinitely many
options. However, the natural ordering of \(\bbZ\) suggests we can use a single
real-valued score \(s(x, \zparam) \in \bbR\) alongside a discretizing operation:
\[ \widehat{z}_\text{int}(x, \zparam) = \lfloor s(x, \zparam) \rfloor\,. \]
Applying the straight-through idea here results in ignoring the floor operation
in the backward pass:
\[
\begin{aligned}
\widehat{z}_\text{int-ST}(s) &\defeq \lfloor s \rfloor \\
\pd{}{s} \widehat{z}_\text{int-ST}(s) &\defeq 1\,. \\
\end{aligned}
\]
The floor function is always within \(\pm1\) of the identity function, and this
strategy has proven useful in constructing integer normalizing flow models
\citep{idf} as well as in learning efficient neural networks with fixed
precision weights \citep{quantize}. Note, however, that arbitrarily close
functions may have arbitrarily far away derivatives.

\subsection{Vector Quantization}

Extending the scalar idea to a vector space leads to the VQ-VAE construction
introduced by \citet{vqvae}.
This model draws ideas from vector quantization, as
used in clustering and compression. It similarly uses an ``encoder-decoder''
architecture, but unlike our framework, the VQ-VAE supports only discrete
categorical variables represented by embeddings $\{ \mbz_i \in \bbR^d : i = 1,
\ldots, K \}$.  In its general form, the VQ-VAE uses an encoder
$\widetilde{f} : \cX \to \bbR^D, \widetilde{f}(\mbx) = \mbz$
and a decoder
$\widetilde{g}: \cX \times \cY \times \bbR^D \rightarrow \bbR$
where for instance $g(\mbx, \mbz, \mby)$ models the probability $\Pr(\var{Y}=\mby \mid
\var{X}=\mbx, \var{Z}=\mbz)$.
But instead of directly computing $\widetilde{g}\left(\mbx, \mby,
\widetilde{f}(\mbx)\right)$, \ie,
letting $\mbz$ be a continuous hidden vector, we instead employ a
quantization step:
\begin{equation}
\widetilde{g}\left(\mbx, \mby, Q(\widetilde{f}(\mbx))\right)
\quad\text{where}\quad
Q(\mbs) = \argmin_{z \in \cZ} \| \mbs - \mbz
\|_2\,.
\end{equation}
The mapping $Q : \bbR^D \to \cZ$ essentially \emph{snaps-to-grid} the encoder
output, by returning the nearest neighbor anchor point.
Importantly, the decoder's third argument (corresponding to $\mbz$) is
parametrized by a neural network that can accept any continuous embedding in
$\bbR^D$, and it is therefore possible to apply the STE to
the quantization mapping $Q$. Additionally, the VQ-VAE loss includes a term
$\| \widetilde{f}(x) - Q\left(\widetilde{f}(x)\right) \|^2$
which simultaneously learns the embeddings and encourages the encoder to output
values as close as possible to the set of allowed embeddings \(\cZ\): this term
can be seen as a penalty on the STE approximation error, and if it were zero,
the STE would be exact.

\section{Interpretation via Pulled-Back Labels}
\label{section:pulled_back_interpretation}

STE and its variants are mainly motivated as a practical solution for learning
problematic models with null gradients. In this section, we provide a possible
interpretation of the end-to-end objective optimized when using STE or SPIGOT.

\begin{figure}
\centering

\begin{tikzpicture}[baseline]
\centering
\scaffold
\draw[->,downstream] (y.north west) to[bend right=20] (yparam);
\node[interm,below of=z,anchor=west] (iloss) {$L_{\var{Z}}(\mbs, \z)$};
\draw[interm,->] ([xshift=-5pt]iloss.west) to[bend right=10] (zparam);
\end{tikzpicture}

\vspace{\baselineskip}

\begin{tikzpicture}[baseline]
\scaffold
\draw[->,downstream] (y.north west) to[bend right=20] (yparam);
\node[interm,below of=z,anchor=west] (iloss) %
  {$L_{\var{Z}}(\mbs, \textcolor{downstreamcol}{\z_\star})$};
\draw[->,downstream,dotted] (y.south west) to[bend right=20] %
  ([xshift=-5pt,yshift=-1pt]iloss.north east);
\draw[interm,->] ([xshift=-5pt]iloss.west) to[bend right=10] (zparam);
\end{tikzpicture}

\caption{
Top: If ground truth supervision were available for the latent variable
 $\z$, $\zparam$ could be trained jointly with an auxiliary loss \(L_\var{Z}\).
Bottom: As such supervision is not available, we induce a best-guess
\emph{pulled-back label} $\z_\star$ by pulling back the downstream loss.
This strategy recovers the STE and SPIGOT estimators.}
\label{fig:pullback-loss}
\end{figure}
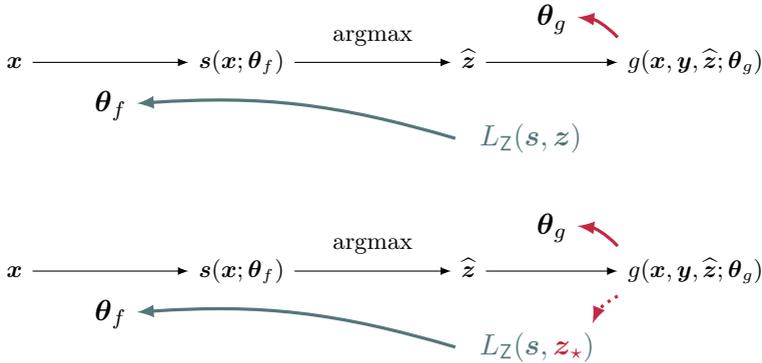

Consider a multitask learning scenario \citep{caruana1997multitask},
where we predict the latent variable as a separate task in addition to the
 downstream task. In this case, $\z$ would be fully observed, \ie, we would
have supervised data in the form of triplets \((\mbx, \mby, \z)\). We could
therefore, in addition to the downstream loss on $\mby$, also incur a loss
$L_{\var{Z}}(\mbs,\z)$ which would allow us to update the parameters $\zparam$.
We will focus for now on the simple perceptron loss:
\begin{equation}\label{eq:perceptron}
L_{\var{Z}}(\mbs,\z) \defeq \DP{\mbs}{\z} - \argmax_{\z' \in \cZ}
\DP{\mbs}{\z'}\,.
\end{equation}
In the latent variable case, we do not have access to this supervision and
instead induce a \emph{pulled-back label} as a best-effort guess at what $\z$
should be, given the available data, as illustrated in
\cref{fig:pullback-loss}.

Let us define the pulled-back label as a sensible approximation to the value for
$\z$ that minimizes the downstream loss:
\begin{equation}
\z_\star :\approx \argmin_{\z \in \conv(\cZ)} \dolo(x, y, \z)\,.
\end{equation}
Notice that the optimization is with respect to \(\z\) and not the model
parameters; the encoder $f$ therefore does not enter this problem. Note also
that this is a continuous optimization problem due to our choice of using the
convex hull of \(\cZ\) as domain. Searching only over the discrete set \(\cZ\)
would be intractable in the structured case; the continuous version can be
approximated with iterative methods.

Still, this optimization problem is challenging and non-convex for many
interesting models, such as deep neural networks.
We may approach it using the \emph{projected gradient algorithm}
\citep{goldstein1964convex,levitin1966constrained}, which takes a step in the
direction of the gradient (with step size \(\eta\)), and then projects the
updated guess to the constraint set, yielding the iteration
\[\mbz^{(t+1)} = \proj{\conv(\cZ)}(\mbz^{(t)} - \eta \pd{\dolo}{3}(x,y,\z^{(t)}))).\]

If \(\dolo\) were convex \wrt its third argument, we could repeat this iteration until
convergence to the optimal \(\mbz_\star\), regardless of initialization.
However, this is infeasible and may be an unwise use of resources, at least
until the model is trained well.

Consider instead starting from the sensible initialization
\(\mbz^{(0)}\defeq
\widehat{\z}_0(\mbs) =
\argmax_{\z \in \cZ} \DP{\z}{\mbs}\)
and applying a single iteration of projected gradient, giving:
\begin{equation}
\begin{aligned}
\z_\star \defeq \proj{\conv(\cZ)}(\widehat{\z}_0 - \eta\pd{\dolo}{3}
(x,y,\widehat{\z}_0))\,.
\end{aligned}
\end{equation}
Using this pulled-back label to update the encoder by the perceptron
loss from \cref{eq:perceptron}, we get the following gradient \wrt \(\mbs\):
\begin{equation}
\begin{aligned}
\pd{}{\mbs}L_{\var{Z}}(\mbs, \z_\star)
&=\widehat{\z}_0 - \z_\star\\
&=\widehat{\z}_0 - \proj{\conv(\cZ)}(\widehat{\z}_0
  - \eta \pd{\dolo}{3}(x,y,\widehat{\z}_0))\\
&=\pd{\widehat{\z}_\text{SPIGOT}}{}(\mbs) \circ
\pd{\dolo}{3}(x,y,\widehat{\z}_\text{SPIGOT}(\mbs))\\
&=\pd{\dolo}{\mbs}(x,y,\widehat{\z}_\text{SPIGOT}(\mbs))\,,
\end{aligned}
\end{equation}
recognizing the SPIGOT surrogate gradient from
\cref{eq:spigot} and undoing the chain rule.
This shows that
SPIGOT minimizes the \textbf{perceptron loss} \wrt a pulled-back label
computed by \textbf{one projected gradient step} on the downstream loss
starting at
\(\widehat{\z}_0\).%

\paragraph{Relaxing the constraints.}

We can relax the constraints for $\z_\star$ and instead optimize
\begin{equation}
\z_\star :\approx \argmin_{\z \in \bbR^{D}} \dolo(x, y, \z)\,.
\end{equation}
This optimization problem can again be solved iteratively, but without the
projection step. One step of gradient descent starting at \(\widehat{\z}_0\)
yields:
\begin{equation}
\begin{aligned}
\z_\star \defeq \widehat{\z}_0 - \eta \pd{\dolo}{3}(x,y,\widehat{\z}_0)\,.
\end{aligned}
\end{equation}
Applying the perceptron loss again, assuming $\eta=1$,
\begin{equation}
\begin{aligned}
\pd{}{\mbs} L_{\var{Z}}(\mbs, \z_\star)
&=\widehat{\z}_0 - \z_\star\\
&=\widehat{\z}_0 - \left(\widehat{\z}_0 - \pd{\dolo}{3}(x,y,\widehat{\z}_0)\right)\\
&=\pd{\dolo}{3}(x,y,\widehat{\z}_0) \\
&=\pd{\widehat{\z}_\text{STE}}{}(\mbs)
 \circ \pd{\dolo}{3}(x,y,\widehat{\z}_{\text{STE}}(\mbs))\\
&=\pd{\dolo}{\mbs}(x,y,\widehat{\z}_{\text{STE}}(\mbs))\,.
\end{aligned}
\end{equation}
This reveals that indeed STE can also be interpreted as minimizing a perceptron
loss on a pulled-back label, and that SPIGOT is indeed a version of STE that
takes into account the structure of the marginal polytope \(\conv(\cZ)\).

This view was proposed by \citet{mihaylova-etal-2020-understanding} and
employed to develop new surrogate gradient methods by changing other components,
such as the loss function (from perceptron to
cross-entropy loss) or the optimization algorithm (mirror descent instead of
projected gradient).

\section{Summary}

Surrogate gradient methods provide a way to train deterministic models without
compromising on the use of discrete mappings in the forward pass (\eg, argmax or
quantization layers). Whereas in \cref{chapter:relax} we relaxed the forward
pass mapping and computed its gradient exactly, in this chapter we leave the
forward pass unchanged, but only employ a relaxation in the backward pass.
While strictly speaking incorrect, surrogate gradients can often be justified in
principled ways and can work well in practice.
\citet{mihaylova-etal-2020-understanding} empirically compared surrogate gradient
methods against relaxations on a task where \(\z\) is known,
and found that both classes of methods are able to reach similar downstream
performance. Relaxation methods seem to converge faster and are more robust to
initialization, but are worse than straight-through methods at recovering a
discrete latent \(\z\) that agrees with the experiment design.

All methods seen so far require
\cref{assumption:relaxed_decoder}:
continuity of the decoder $g$ \wrt
\(\mbz\). Even if surrogate gradient methods use only discrete \(\mbz\) in the
forward pass, the backward pass requires the gradient
\(\pd{g}{3}(x,y,\widehat{\z})\).
In the next chapter, we give the discrete latent a
probabilistic treatment, allowing us to have correct (unbiased) estimation
without compromising discreteness and without any requirements on the form of \(g\).

\chapter{Probabilistic Latent Variables}\label{chapter:expectation}

\section{Formulating the Probabilistic Model.}
In this section, we directly tackle the probabilistic form of discrete latent
variable learning problem.
Instead of
the downstream model seeing a
 single choice of \(\zhat\), we will
instead consider the expectation of the downstream model
over all possible assignments of the latent variable:
\begin{equation}\label{eqn:explicit_redux}
\begin{aligned}
\bar \dolo(x, y)
  \defeq&~\bbE_{\var{Z}}\left[\dolo(x, y, \var{Z})\right] \\
       =&~\sum_{\z \in \mathcal{Z}} \dolo(x, y, \z) \Pr(\var{Z}=\z \mid x), \\
\end{aligned}
\end{equation}
The distribution $\Pr(\var{Z}=\z \mid x)$ is defined in terms of the
encoder $f(x, \z)$. A standard way is $\Pr(\z \mid x) \propto \exp f(x, \z)$, but
we shall see in this chapter that other ways to define this distribution are possible.

In \cref{chapter:relax}, we have avoided the challenging sum
and relaxed the problem by pushing the expectation inside:
$\bbE_{\var{Z}} \left[ \dolo(x, y, \var{Z}) \right] \approx
\dolo\left(x, y, \bbE[\var{Z}]\right)$.
In \cref{chapter:surrogate}, we eschewed any probabilistic interpretation and
instead tried to approximate the discrete mapping
$\dolo\left(x, y, \left( \argmax_{\z\in\cZ} f(x, \z) \right) \right)$. We now finally
come face-to-face with this computation.

The probabilistic framework is challenging because, per \cref{chapter:struct},
\(\cZ\) may be an exponentially large combinatorial set. In general, this comes
at a cost; the payoff however is that we can get well-principled models with
much fewer assumptions on the downstream model \(g\).

While many implementations of probabilistic models rely on stochastic computation
(sampling), this is not necessarily the case.
The first and the last strategy we present in this chapter are deterministic.

\section{Explicit Marginalization by Enumeration}

In the case where $|\cZ|$ is of manageable size and when the decoder is not too
expensive, it is of course possible to tackle \cref{eqn:explicit_redux}
explicitly by enumerating all possibilities, which requires evaluating the decoder $|\cZ|$ times.
This is for instance possible for categorical (one-of-$K$) discrete latents with
$K$ not too large, or for bit vectors of manageable size. For example,
$\cZ = \{0,1\}^D$ has $2^D$ elements, but for $D=3$ it is not necessarily
prohibitive to evaluate the decoder 8 times.

At this point we may see why explicit marginalization does not require
\cref{assumption:relaxed_decoder},
and $\z$ can be a completely discrete index: gradients of the form
$\pd{\dolo}{3}(x, y, \z)$ are not required or used, because in
\cref{eqn:explicit_redux} $\z$ is not a transformed function of any parameters:
it is an abstract summation index.

When one can afford to explicitly compute this objective, results tend to be
unsurprisingly good in comparison to approximate approaches or relaxations.
Of course, for large $\cZ$ enumeration is prohibitive.
\begin{myexample}{\bfseries Neural attention as latent alignment.}
Recall from \cref{ex:attn_relax} that the neural attention mechanisms can be
seen as relaxed Categorical (one-of-K) variable.
Using a Categorical latent variable instead is known as hard attention and was
proposed by
\citet{xu-hardattn} for image captioning. \citet{deng2018latent}
observed that in the context of machine translation the model can be tackled
with explicit enumeration. They proposed an explicitly-marginalized variational
attention which performed the best on machine translation.
\end{myexample}

\begin{myexample}{\bfseries Variational Auto-Encoder with a discrete latent variable.}
\citet{kingma2014semisup} propose a method for semi-supervised learning based on a
generative neural network. Their model includes a categorical latent variable representing
the target class. To train the model, they build a Variational Auto-Encoder, that is the
posterior distribution over latent variables is approximated with a learned encoder network.
To ensure that the encoder is correctly trained, they explicitly marginalize over
the categorical latent variable, in contrast to the standard approach in this setting
(reparametrization trick for continuous latent variables).
\end{myexample}

Even when not affordable on realistic problems, we recommend implementing explicit
marginalization for testing purposes.

\section{Monte Carlo Gradient Estimation}

When $\Pr(\z \mid x) \propto \exp f(x, \z)$
and an algorithm is available to obtain samples from
$\Pr(\z \mid x) \propto \exp f(x, \z)$, expectations such as \cref{eqn:explicit_redux} can be
approximated via the Monte Carlo method \citep{montecarlo}:
\begin{equation}\label{eqn:mc}
\bbE_{\var{Z}} [ \dolo(x, y, \var{Z}) ] \approx \frac{1}{S} \sum_{i=1}^{S} \dolo(x, y,
\z^{(i)}).
\end{equation}
For learning the decoder parameters, we can directly differentiate
\cref{eqn:mc}, since the expectation operator is linear and so
is the differentiation operator \citep[\S~6.4.1]{mml}.
For any function \(F(\z, \param)\) and distribution
\(\Pr(\var{Z}=\z)\) that does not depend on \(\param\), we have
\begin{equation}
\begin{aligned}
\pd{}{\param} \EE_{\var{Z}} [F(\var{Z}, \param)]
&= \pd{}{\param} \sum_{\z \in \cZ} F(\z, \param) \Pr(\z) \\
&= \sum_{\z \in \cZ} \left(\pd{F}{\param}(\z, \param)\right) \Pr(\z) \\
&= \EE_{\var{Z}} \left[ \pd{F}{\param}(\var{Z}, \param)\right].
\end{aligned}
\end{equation}
However, if $\Pr(\var{Z}\mid \param)$ does depend on
$\param$ (even if \(F\) does not) then differentiating is harder, as
simply applying the chain rule does not directly lead to an expectation that can
be estimated from samples $\z^{(1)},\ldots,\z^{(S)}$:
\begin{equation}\label{eqn:mc_grad_fail}
\begin{aligned}
\pd{}{\param} \EE_{\var{Z}\sim \Pr(\z \mid \param)} [F(\var{Z})]
&= \pd{}{\param} \sum_z F(\z) \Pr(\z \mid \param) \\
&= \sum_{\z \in \cZ} F(\z) \left(\pd{}{\param} \Pr(\z \mid \param)\right). \\
\end{aligned}
\end{equation}
For this reason, applying the Monte Carlo method to gradient estimation requires
nontrivial techniques. The next two sections describe two main strategies used
for Monte Carlo gradient estimation with discrete and structured
latent variables. A more general and thorough review can be found in~%
\citep{monte_carlo_grad}.

\section{Path Gradient Estimator (The Reparametrization Trick)}
\label{sec:reparam}

Estimating gradients of expectations is challenging not only in the discrete
setting, but also in the continuous one.  An alternative to the SFE is given by
the \emph{Path Gradient Estimator} (PGE),
introduced by \citet{ho1983optimization} and
also known as the

push-out gradient method \citep{rubinstein1992sensitivity},
process derivative \citep{pflug2012optimization},
perturbation analysis \citep{glasserman1990gradient},
or the
reparametrization trick \citep{rezende2014dgm,kingma2013auto,titsias2014varbayes}.

The challenge, identified in \cref{eqn:mc_grad_fail},
 stems from the fact that the distribution of the latent
variable, over which the expectation is taken, depends on the parameters
$\param$. The PGE circumvents this problem by
extracting the source of randomness into
a parameter-free random variable $\var{U}$, using a
transformation $\phi$ that ensures $\var{Z}=\phi(\var{U}, \param)$,
where equality is in distribution.
For any function \(F(\z)\)
we may then write
\citep[Def.~6.3]{mml}
\[
\EE_{\var{Z}\sim\Pr(\var{Z} \mid \param)} [F(\var{Z})] =
\EE_{\var{U}\sim\Pr(\var{U})} \left[
F\left(\phi(\var{U}, \param)\right) \right].
\]
The right-hand side is an expectation over a distribution that does not depend
on $\param$: we have moved the parameter into the argument of the expectation.
We can then obtain the Monte Carlo gradient estimator:
\begin{equation}\label{eqn:pge}
\pd{}{\param} \EE_{\var{Z}\sim\Pr(\var{Z} \mid \param)} [F(\var{Z})] =
\EE_{\var{U} \sim
\Pr(\var{U})} \left[ \pd{F}{\param} \left(\phi(\var{U}, \param)\right) \right].
\end{equation}
If $\var{U}$ is a continuous random variable, the right-hand-side
expectation is an integral. In order for the exchange of integration and
differentiation to be valid, the following conditions must hold
\citep{lecuyer}:
\begin{itemize}
\item
\(\pd{}{\param} F(\phi(\mbu,\param))\)
exists;
\item \(\EE_{\var{U}}[F(\phi(\var{U},\param))] < \infty \)
for any \(\param\).
\item There exists an integrable function \(h(\mbu)\) such that
\[ \sup_\param \|\Pr(\mbu) \cdot
\pd{}{\param} F(\phi(\mbu,\param))
\| \leq
h(\mbu)\,.\]
\end{itemize}
These assumptions are for instance satisfied if \(F \circ \phi\) is Lipschitz.

The PGE finds many successful applications in continuous variables. A prominent
example is the reparametrization of a Gaussian random variable: If $\var{Z} \sim
\operatorname{Normal}(m, \sigma^2)$ then $\var{Z}$ is identical in
distribution to a shifted and scaled standard normal:
\[ \var{Z} {=} \phi(\var{U}, m, \sigma) = m + \sigma \var{U},
\quad\text{where}\quad
\var{U} \sim \operatorname{Normal}(0, 1).
\]

\paragraph{The Gumbel reparametrization.}
In the discrete setting, applying the PGE hits further roadblocks. However,
this line of thinking opens the door to some valuable approximations.
A fundamental result for the reparametrization of discrete random variables is
the Gumbel-argmax construction for the categorical distribution.
Taking $\cZ = \{ \mbe_1, \ldots, \mbe_K \}$, the categorical distribution
parametrized by a score vector $\mbs \in \bbR^K$ is
\[ \Pr(\var{Z}=\z) = \exp \DP{\z}{\mbs} / \sum_j \exp \DP{\mbe_j}{\mbs}, \]
in other words, it is the distribution induced by the \emph{softmax} of the
scores. It turns out that $\var{Z}$ is identical in distribution to the perturbed argmax \citep{gumbel1954statistical,yellott1977relationship}:
\begin{equation}\label{eq:perturbed_map}
\var{Z} \stackrel{d}{=}
\phi(\var{U},\mbs) = \argmax_{\z\in\cZ} \DP{\z}{\mbs + \var{U}},
\end{equation}
where $\var{U}$ is a vector of i.i.d.\ standard Gumbel random variables
$\var{U}_i \sim \operatorname{Gumbel}(0)$, and \(\cZ=\{\mbe_1,\ldots,\mbe_K\}\).
This reparametrization is, alas, not suitable for the PGE as in \cref{eqn:pge},
because $\phi$ is discontinuous and therefore so is $F \circ \phi$.
However, this reformulation paves the way for a relaxation we discuss
next.

\paragraph{Gumbel-Softmax and the Concrete distribution.}
As in \cref{chapter:relax}, we may relax the argmax in the Gumbel-Max
reparametrization to a softmax mapping,
yielding the Gumbel-Softmax \citep{gumbel_softmax}, also known as the
Concrete distribution \citep{concrete_distribution}.
Consider the parametrization
\[
\phi_{\gamma H_1}(\var{U},\mbs) \defeq
\widehat{\z}_{\gamma H_1}(\mbs+\var{U})
= \softmax\left((\mbs + \var{U}) / \gamma \right)\,,
\]
which is a perturbed version of the softmax relaxation from \cref{sec:cat-relax}
where $\gamma > 0$ is a regularization strength, or ``temperature'' hyperparameter.
Let $\var{Z}_{\gamma H_1} \stackrel{d}{\defeq} \phi_{\gamma H_1}(\var{U}, \mbs)$.
Since $\var{Z}_{\gamma H_1}$ takes values in $\operatorname{relint} \conv(\cZ)$
and $\var{Z}$ takes values in $\cZ$, \ie, their supports do not overlap,
the two are not only distinct, but difficult to compare. (We revisit this issue
at the end of this section.)
Nevertheless, the approximation becomes exact in the $\gamma \to 0$ limit,
suggesting a meaningful approximation.\footnote{
This temperature-based
argument requires care. Indeed, in the $\gamma\to 0$ limit, $\var{Z}_{\gamma
H_1}\to\var{Z}$, but
also in this limit $\phi$ approaches a threshold function, so its gradients approach zero (or
infinity near discontinuities). In practice, $\gamma$  must be set far away from
zero to ensure numerical stability and meaningful gradients, but not too far, or
else the approximation gap is too big, as reported, \eg, by
\citet{baziotis-etal-2019-seq}.
}
Plugging in the Gumbel-Softmax approximation into our framework, we could fully
embrace the relaxation and consider a continuous latent variable model:
\[
\begin{aligned}
\bar{\dolo}(x,y,\zparam,\yparam)
&=\EE_{\var{Z}_{\gamma H_1}}\left[ \dolo(x, y, \var{Z}_{\gamma H_1}, \yparam) \right]
\\&=\EE_{\var{U}}\left[ \dolo(x, y, \widehat{\mbz}_{\gamma
H_1}(\mbs(x,\zparam)+\var{U}), \yparam) \right]
\end{aligned}
\]
using the PGE.

For global-structured latent variables $\var{Z}$ whose structure is characterized or representable by
a dynamic program, \ie, equivalent to paths in a (hyper)graph
(\cref{sec:dag}), sampling from $\Pr(\z \mid x) \propto \exp f(x, \z)$
is possible via the Forward-Filtering Backward-Sampling (FFBS) algorithm
\citep{ffbs}.
In this case, the Backward-Sampling pass can be modified with a Gumbel-Softmax
distribution, leading to differentiable samples \citep{fu2020gumbelcrfs}.
Similarly, in the temperature limit, the approximation becomes exact.

\paragraph{Structured perturbation methods.}
Extending the Gumbel reparametrization to structures is
challenging:
the direct approach would require perturbing the score of each structure
independently, which is intractable. For directed models (\eg, incremental
structure prediction), \citet{kool2020topk} propose a top-down sampling
approach based on sampling Gumbels conditional on their sum.
For undirected models (MRF), \citet{hazan2013gibbs} propose an
algorithm based on factor-decomposed perturbations and Gibbs sampling.

The Perturb-and-MAP approach
\citep{papandreou2011perturb}
instead applies perturbations directly to the scores $\mbs$ of an arbitrary
structured model: this induces a random field $\var{Z}$ but the probability
distribution is not the Gibbs one. \citet{hazan2012partition}
derive some bounds on the relationship between Perturb-and-MAP and the true
log-partition function.
However, in many cases
it is not essential to prescribe $\var{Z}$ to have the Gibbs distribution, and
the Perturb-and-MAP approach can thus be very useful.
To ensure differentiability, the role of the softmax is taken by marginal
inference for structured latent variables, and we can therefore use this
strategy whenever an algorithm for marginal inference is available \citep{caio-iclr,caio-acl}.

\paragraph{Straight-Through Gumbel (Gumbel-ST).}
Combining ideas of the PGE with the surrogate gradient strategies of
\cref{chapter:surrogate}, we may use the true discrete $\var{Z}$ in the forward pass,
and only apply a reparametrized relaxation in the gradient estimation for the
backward pass.
This strategy, known as the Straight-Through Gumbel-Softmax estimator
\citep{gumbel_softmax}, corresponds to using the construction
\(\widehat{\z}_\text{Softmax-ST}\) from \cref{eq:softmax-st}:
\[
\var{Z}_\text{Softmax-ST} \defeq
\widehat{\z}_{\text{Softmax-ST}}(\mbs+\var{U})\,,
\]
leading to the following approximate gradient:
\[
\begin{aligned}
\pd{}{\zparam} \bar{\dolo}(x,y,\zparam,\yparam)
&\approx\pd{}{\zparam}\EE_{\var{Z}_\text{Softmax-ST}}\left[ \dolo(x, y,
\var{Z}_\text{Softmax-ST}, \yparam) \right] \\
&=\pd{}{\zparam}\EE_{\var{U}}\left[ \dolo(x, y,
\widehat{z}_\text{Softmax-ST}(\mbs(x, \zparam) + \var{U}), \yparam) \right] \\
&=\EE_{\var{U}}\left[ \pd{}{\zparam} \dolo(x, y,
\widehat{z}_\text{Softmax-ST}(\mbs(x, \zparam) + \var{U}), \yparam) \right]\,. \\
\end{aligned}
\]
This time, in the forward pass only discrete values \(\z\in\cZ\) are passed to
the downstream model. Still, due to the use of relaxations in both Gumbel-Softmax
and Gumbel-ST,
\cref{assumption:relaxed_decoder}:
 is still necessary.
\citet{paulus2020gradient} catalog and study a number of structured
Gumbel-based models, and \citet{huijben2022review}
provide a broader survey of gradient estimation with
Gumbel-Max and Gumbel-Softmax is available
The Gumbel-ST approximation is well-suited and popular for learning
models with incremental latent structure, for instance
\emph{sequence-to-sequence-to-sequence} models \citep{baziotis-etal-2019-seq}
and related \citep{xu2022latentqueries}.

\paragraph{Implicit Maximum Likelihood Estimation (I-MLE).}

One interesting approach is to combine the structured perturbation methods
described above with surrogate gradient approaches
(\cref{section:pulled_back_interpretation}). An implicit maximum likelihood
estimation approach with perturbation-based implicit differentiation has been
proposed by \citet{niepert2021implicit} which combines these two ideas.
I-MLE uses a surrogate gradient construction defined as
\begin{equation}
\begin{aligned}
\zsurr_\text{I-MLE}(\s) &\defeq \zsurr_0(\s) = \argmax_{\z \in \cZ}
\DP{\s}{\z}\,, \\
\pd{\zsurr_\text{I-MLE}}{}(\s)(\zvector) &\defeq \zsurr_0(\s) -
\zsurr_0(\s-\eta\zvector)\,.
\end{aligned}
\end{equation}
This is similar to the linear interpolation surrogate of \citet{diffbb}, see \cref{eq:surr-li},
with \(\eta\) absorbed into the global learning rate.
In addition, I-MLE uses a perturb-and-MAP  construction
\[ g(x,y,\zsurr_\text{I-MLE}(\s + \var{U}))\,, \]
where \(\var{U}\) is a noise variable.
\citet{niepert2021implicit} point out that if \(\var{U}\) is taken to always
be zero, I-MLE recovers the deterministic linear interpolation approach; in
their experiments, perturbations improve substantially.
Remark that the noise sample from
\(\var{U}\) appears in both terms of the surrogate gradient.

\paragraph{Mixed Random Variables.}

As pointed out above, one disadvantage of the Gumbel-Softmax construction is
that, for any temperature $\gamma > 0$, the resulting random variable
$\var{Z}_{\gamma H_1}$ is continuous, taking values in $\operatorname{relint}
\conv(\cZ)$, whereas the discrete random variable that one would like to
approximate, $\var{Z}$, takes values in $\cZ$---their supports have empty
intersection. Although in the zero temperature limit ($\gamma \rightarrow 0_+$)
$\var{Z}_{\gamma H_1}$  approaches a categorical distribution, in practice it is
hard to learn with low temperatures, as the gradients will be very small and
numerical instabilities often arise.

\begin{figure}\centering
\includegraphics[width=.66\textwidth]{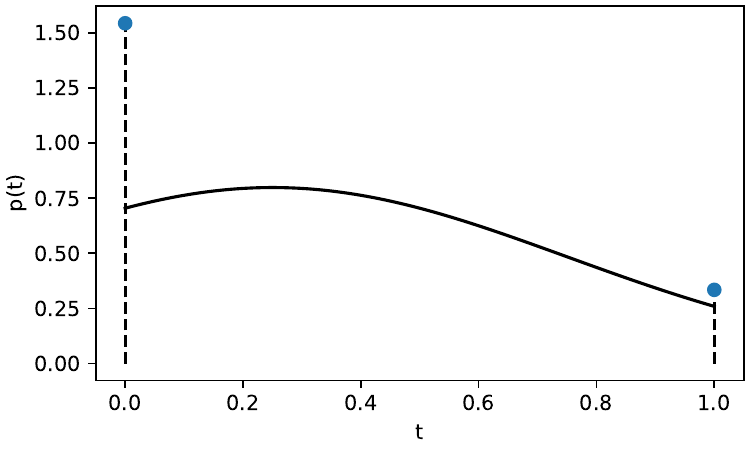}
\caption{\label{fig:truncdist}Projected distribution onto the unit interval. In
this case, the base distribution is Gaussian.}
\end{figure}

This observation led several researchers to propose \emph{truncated}
distributions which can be regarded as hybrids of discrete and continuous random
variables, called \emph{mixed} random variables \citep{farinhas2022sparse}. For
the binary case ($|\cZ| = 2$), the probability simplex is isomorphic to the unit
interval, $\triangle_1 \simeq [0,1]$, so a simple procedure to obtain a mixed
random variable $\var{Z}$ is first sampling a continuous variable $\var{U}$
using an auxiliary distribution $p(\var{U})$ whose support contains $[0,1]$
properly, and then defining $\var{Z}$ as the projection onto the unit interval,
\begin{equation}\label{eqn:hard_sigmoid}
\begin{aligned}
\var{Z} = \max\{0, \min\{1, \var{U}\}\}.
\end{aligned}
\end{equation}
The resulting probability density function of $\var{Z}$ will be the combination
of two Dirac delta functions (point masses) placed at $0$ and $1$ with the
probability density function of $\var{U}$ in $]0,1[$, as shown in
\cref{fig:truncdist}
An example of this construction is the \emph{Hard Concrete
distribution} \citep{louizos2018learning}, where $\var{U}$ is a ``stretched''
Concrete distribution and $\var{Z}$ is defined as in \cref{eqn:hard_sigmoid}. Of
course, this rectification can be applied to other continuous distributions
besides the (stretched) Concrete and has in fact been so; a simple choice is the
Gaussian distribution, to which one-sided \citep{hinton1997generative} and
two-sided rectifications \citep{palmer2017methods} have been proposed. Another
example if the Hard-Kuma distribution \citep{bastings2019interpretable}, which
uses the Kumaraswamy \citep{kumaraswamy1980generalized} as the base
distribution. These ``stretch-and-rectify'' techniques enable assigning
probability mass to the boundary of $\triangle_1$ and are similar in spirit to
the spike-and-slab feature selection method
\citep{mitchell1988bayesian,ishwaran2005spike} and for sparse codes in
variational auto-encoders \citep{rolfe2016discrete,vahdat2018dvae++}. Since they
apply to binary random variables, they can be used to approximate ``bit-vector''
latent variables, where $\cZ = \{0,1\}^K$ for some $K \in \mathbb{N}$.

\begin{figure}[t]
\begin{center}
\includegraphics[width=0.75\columnwidth]{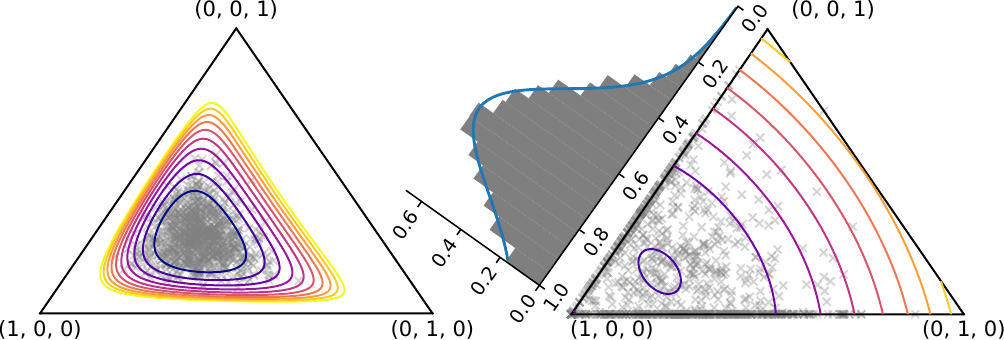}
\caption{Multivariate distributions over $\triangle_{K-1}$. Standard
distributions, like the Logistic-Normal (left), assign zero probability to all
faces except the maximal one $\operatorname{relint}(\triangle_{K-1})$. In contrast, mixed
distributions support assigning probability to the \textit{full} simplex,
including its boundary: the Gaussian-Sparsemax (right) induces a distribution
over the 1-dimensional edges (shown as a histogram), and assigns $\Pr\left(\{(1,
0, 0)\}\right) = .022$. Reproduced with permission from \citet{farinhas2022sparse}.}
\label{fig:multivar-gaussian-spmax}
\end{center}
\end{figure}

We show next how to generalize the idea above to other choices of $\cZ$, for example to approximate a multivariate categorical variable $\var{Z}$.
We note first that the hard sigmoid transformation of \cref{eqn:hard_sigmoid} is
just a particular case of the sparsemax transformation when $|\cZ|=2$. This immediately suggests an extension of the Hard Concrete and rectified Gaussian distributions for the multivariate case, which has been proposed and studied in detail by \citet{farinhas2022sparse} under the names Gumbel-Sparsemax and Gaussian-Sparsemax, respectively.  Figure~\ref{fig:multivar-gaussian-spmax} shows illustrates the latter---the Gaussian-Sparsemax can also be seen as a ``mixed'' counterpart of the multivariate logistic normal distribution, where the softmax is replaced by a sparsemax transformation.
It has been shown by \citet{farinhas2022sparse} that these mixed distributions
have solid mathematical properties which make them amenable for
information-theoretic treatment, if one chooses as the underlying base measure
the ``direct-sum measure''---a combination of the counting measure (which underlies discrete variables) with Lebesgue-Borel measures of multiple dimensionalities, associated to each face of the probability simplex.
By doing so, one avoids the cumbersome practice (which is commonplace when the
Gumbel-Softmax relaxation is used with discrete latent variables) of
treating a distribution as discrete or continuous as convenient in different
computations. Indeed, differential entropy and discrete Shannon entropy have
different properties, starting with the fact that the former can be negative.
The full effect of this mismatch is not clear, but
\citet{farinhas2022sparse} report improved performance when
using mixed random variables under the appropriate direct-sum measure.

\section{Score Function Estimator}

The Score Function Estimator
\citep[SFE,][]{rubinstein1976monte,paisley2012variational}, dubbed in some contexts
REINFORCE \citep{Williams1992}
or the likelihood ratio method \citep{glynn1990likelihood},
relies on the fact that $(\log t) ' = 1/t$ implies that $f'(t) = f(t) (\log f(t))'$ to
rewrite the expectation in \cref{eqn:mc_grad_fail} as
\begin{equation}
\begin{aligned}
\pd{}{\param} \EE_{\var{Z}\sim \Pr(\z \mid \param)} [F(\var{Z})]
&= \sum_{\z \in \cZ} F(\z) \left(\pd{}{\param} \Pr(\z \mid \param)\right) \\
&= \sum_{\z \in \cZ} F(\z) \left(\Pr(\z \mid \param) \pd{}{\param} \log \Pr(\z \mid \param)\right) \\
&= \EE_{\var{Z}} \left[ F(\var{Z}) \left(\pd{}{\param} \log \Pr(\var{Z} \mid \param)\right)
\right]. \\
\end{aligned}
\end{equation}
In this case, since \(\var{Z}\) is discrete there are no requirements on \(F\)
needed to swap differentiation and expectation.\footnote{The SFE can also be
applied in continuous cases, and in that case the requirements are more
stringent \citep[\S4.3]{monte_carlo_grad}.}
This means we can use a Monte-Carlo approximation to train both the encoder and
the decoder in a discrete latent variable model.
With the notation
$\bar{\dolo}(x, y%
) \defeq \EE_{\var{Z} \sim \Pr(\var{Z} \mid x;
\zparam)} \left[ \dolo(x, y, \var{Z}; \yparam) \right]$, we have
\begin{equation}\label{eqn:mc_grad}
\begin{aligned}
\pd{}{\yparam} \bar{\dolo}(x, y)
&\approx \frac{1}{S} \sum_{i=1}^{S} \pd{}{\yparam} \dolo(x, y,
\z^{(i)}, \yparam), \\
\pd{}{\zparam}
\bar{\dolo}(x, y)
&\approx \frac{1}{S} \sum_{i=1}^{S} \dolo(x, y,
\z^{(i)}, \yparam) \left(\pd{}{\zparam} \log \Pr(\z^{(i)} \mid x; \zparam)
\right). \\
\end{aligned}
\end{equation}

Altogether, this leads to a strategy for gradient-based learning with discrete or structured
latent variables, with no constraints at all on the dependency of $\dolo$ on $\z$.
This explains the origins and success of the SFE in the reinforcement learning
community, when often the rewards are computed via simulations and interactions
with an environment that cannot always be relaxed.

Moreover, in contrast to \cref{chapter:relax,chapter:surrogate}, even if
\cref{eqn:mc,eqn:mc_grad} are approximate, they are \textbf{unbiased
estimators} of the underlying stochastic quantities: they are equal to the true
values in expectation (w.r.t.\ the sampling process).

The most prominent limitation is that SFE has rather high variance; the
remainder of this section discusses variance reduction strategies. For
structured latent variables, there is another, less-discussed limitation: the
SFE has among the strictest computational requirements, needing a sampling
oracle for $\Pr(\var{Z} \mid x)$, as well as access to
$\pd{}{\param} \log \Pr(\var{Z} \mid x, \param)$.

\paragraph{Implementation details.}
The SFE is
by definition a custom gradient estimator, and thus not immediately compatible
with the automatic differentiation language of current deep learning frameworks,
in which we specify a computation (or estimator of) the loss, and the gradients
are handled for us. The most common approach is to define a surrogate function
$\dolo_\text{SFE}(x, y)$ which, when automatically differentiated, yields the appropriate
gradients from \cref{eqn:mc_grad}. This can be accomplished using the
$\operatorname{stopgrad}$ (stop gradient) operator, which returns its argument in the
forward pass, but a zero gradient in the backward pass. Then, the following
surrogate function is suitable:
\[
\begin{aligned}
\dolo_\text{SFE}(x, y) =~&
\frac{1}{S} \sum_{i=1}^{S} \Big[
\dolo(x, y, \z^{(i)}, \yparam) \\
&+ \operatorname{stopgrad}\left(\dolo(x, y, z^{(i)}, \yparam)\right)
\log \Pr(z^{(i)} \mid x; \zparam)
\Big]\,.\\
\end{aligned}
\]
Since the value  of $\dolo_\text{SFE}$ in
the forward pass is arbitrary,
it should not be used for reporting or monitoring performance.
An alternative and perhaps more transparent implementation strategy can be to
define a custom operation within the deep learning framework by explicitly
specifying the forward and backward passes, as described in
\cref{chapter:surrogate}.

\subsection{Variance Reduction}
A drawback of the SFE is its high variance: unless many samples are taken, the
estimator might be too noisy for effective learning. As a result, there is
a substantial body of work devoted to variance reduction techniques.
We briefly discuss three main directions and their application for discrete
latent variables.

\paragraph{Control variates.}
The control variate construction
\citep[\S 7.4]{monte_carlo_grad} is one of the most fruitful variance reduction
strategies. Given some $h : \cZ \to \bbR$ whose expectation
under $\Pr(\z \mid \param)$ is known, and some $\beta \in \bbR$, we can construct
\[ F_{\beta,h}(\z) = F(\z) - \beta (h(\z) - \bbE_Z [h(\var{Z})])\,. \]
Taking expectations,
\[ \bbE_{\var{Z}} [ F_{\beta,h}(\var{Z}) ] = \bbE_\var{Z}[ F(\var{Z}) ] \,,\]
so any unbiased gradient estimator will remain correct. It remains to choose $h$ and
$\beta$ may lead to lower variance.
We may take $h$ to be the score function $\pd{}{\param} \log\Pr(\z \mid \param)$,
because
\[
\begin{aligned}
\bbE \left[ \pd{}{\param} \log\Pr(\var{Z} \mid \param) \right] &=
\sum_{\z \in \cZ} \Pr(\z \mid \param)
\frac{\pd{}{\param} \Pr(\z \mid \param)}{\Pr(\z \mid \param)}\\
&= \pd{}{\param} \sum_{\z \in \cZ} \Pr(\z \mid \param)\\
&= \pd{}{\param} 1 = 0\,.
\end{aligned}
\]
This leads to the estimator
\begin{equation}
\pd{}{\param} \bbE_{\var{Z}\sim \Pr(\z \mid \param)} [F(\var{Z})]
= \bbE_{\var{Z}} \left[ (F(\var{Z}) - \beta) \left(\pd{}{\param}
 \log \Pr(\var{Z} \mid \param)\right) \right].
\end{equation}
A simple but effective way to choose $\beta$ is as a value close to
$F(Z)$, in order to keep the magnitude of the loss term small. This
includes \textbf{running average} and \textbf{exponentially moving average}
(EMA) of previously-seen values of $F(\var{Z})$ during learning. Running averages give
equal weight to all values, while EMA can prioritize recent observations, which
is useful in a machine learning context where the loss should decrease.
From the same motivation, another good choice is the model prediction itself
$\beta=F(\z')$ for some $\z'$ that does not depend on $\var{Z}$. Using the
highest-scoring $z'$ is known as \textbf{self-critic
baseline} \citep{rennie2017self}, another good choice is to draw $\z'$ as an
independent sample from $\Pr(\var{Z} \mid \param)$.
These strategies are easy to implement and in practice
tend to make a substantial difference compared to no variance reduction at all.
We recommend never using SFE without at least one of these simple variance
reduction strategies.

\paragraph{Learned data-driven baselines and NVIL.}
The EMA estimator for $\beta$ corresponds to learning
$\beta$ as a scalar parameter using gradient descent with a squared loss
$(\beta - F(\z))^2$.
If the problem also has some contextualizing input $x$ then one may consider
a \textbf{learned data-driven baseline} $\beta(x; \param_\beta)$, with its
own parameters to learn \citep{Mnih2014}.

\paragraph{Relaxation-based control variates.} MuProp \citep{MuProp} approximates the objective using
a deterministic, relaxed network (replacing $\var{Z}$ with $\bbE[\var{Z}]$);
but, in contrast to \cref{chapter:relax}, this relaxation is only used for
variance reduction when optimizing the true stochastic model. The derivation is
based on a Taylor expansion around $\bbE[\var{Z}]$.
REBAR \citep{Tucker2017} uses Gumbel-Softmax relaxations (\cref{sec:reparam}) to compute a stochastic but
reparametrizable baseline. RELAX \citep{RELAX} uses a trained differentiable network as a surrogate,
trading off additional parameters for a smaller approximation gap.
These methods require continuity and differentiability of the objective
(or, in the case of RELAX, of the surrogate network) w.r.t.\ the latent
variable. MuProp seems applicable to globally structured latent variables, but
REBAR and RELAX rely on a deterministic mapping between the relaxed
continuous variable (\eg, Gumbel-Softmax) and the corresponding discrete sample.
For binary and one-of-K variables such a mapping exists, but in the structured
case this is not available.

\paragraph{Rao-Blackwellization.} This strategy relies on tractable conditioning
as a way to reduce variance \citep{casella1996raoblackwellisation}, and has been used for black box variational inference \citep{ranganath2014blackboxvi,titsias2015blackboxvi}.
One particularly successful example is the
sum-and-sample estimator \citep{RB19}, which provides
a trade-off between explicit marginalization and the SFE by summing over some
values $\bar{\cZ} \subset \cZ$ and compensating with a sample from the
remainder:
\[
\begin{aligned}
\bbE\left[\pd{}{\param} F(\var{Z})\right] =~&
\pd{}{\param} \left(
\sum_{\z \in \bar{\cZ}}
\Pr(\z \mid \param) F(z)
\right)\\
&+ (1-\Pr(\bar{\cZ} \mid \param)) \bbE_{\Pr_{\bar{\cZ}}(\var{Z} \mid
\param)}\left[ F(\var{Z}) \pd{}{\param} \log
\Pr_{\bar{\cZ}}(\var{Z} \mid \param) \right]
\end{aligned}
\]
where we defined the renormalized tail distribution
obtained after discarding the
summed-out values \(\bar{\cZ}\):
\[\Pr_{\bar{\cZ}}(\z) \propto
\begin{cases}
0,& \z \in \bar{\cZ},\\
\Pr(\z \mid \param),& \z \not\in \bar{\cZ},\\
\end{cases}\]
A good choice for $\bar{\cZ}$ is a set of
highest-probability values.
This strategy does not efficiently translate to the general structured
case, as the second term
requires computing and sampling from the renormalized tail distribution.
Another example of Rao-Blackwellized estimators is the
sample-without-replacement estimator \citep{Kool2020Estimating}.

\section{Sparsifying the Distribution}
The first method discussed in this section, exact marginalization, is entirely
deterministic; there is no sampling and all computations are exact, but it is
costly or even intractable for large $\cZ$. The next two methods invoke Monte
Carlo estimation to achieve tractability, but this introduces randomness, hence
variance, which may be undesirable and lead to unstable learning. In this
section we present middle ground strategies based on sparsity.

Until now, we have assumed that $Z$ follows a Gibbs distribution,
\begin{equation}\label{eqn:gibbs_reprise}
\Pr(\var{Z}=\z \mid x; \zparam) \propto \exp f(x, \z, \zparam).
\end{equation}
This choice can be motivated by a maximum-entropy principle,
as discussed in
\cref{sec:regularized_argmax}

Unlike the Monte Carlo strategies, which are
based on sampling, we now seek a way to deterministically compute the
full expectation over a discrete set,
\begin{equation}
\EE[g(x, y, \var{Z})] = \sum_{\z \in \cZ} g(x, y, \z) \Pr(\z \mid x, \zparam),
\end{equation}
by choosing a parametrization in which $\Pr(\z \mid x, \zparam)$ is set to
exactly zero for many $\z \in \cZ$.
When this happens, the sum and its gradient can be computed efficiently with
much fewer calls to the downstream model $g$. Importantly, the Gibbs parametrization in
\cref{eqn:gibbs_reprise} can never give us this property, since $\exp(\cdot) >
0$.

In \cref{sec:regularized_argmax},
we have explored sparse counterparts to the softmax transform
in the context of replacing $\var{Z}$ by a relaxed mean counterpart
$\hat{\z} = \EE[\var{Z}]$,
and using the continuity of $\hat{\z}$ to learn end-to-end models with a single call to a continuous
decoder.
In this section, we show how sparse mappings can also induce a sparse
distribution over $\var{Z}$, leading to efficient marginalization, as proposed by
\citet{correia2020EfficientMarginalizationDiscrete}.

\paragraph{One-of-K (categorical) case.}
Let us set
\begin{equation}\label{eqn:structured_sparsemax}
\Pr(\var{Z}=\z \mid x) = [f(x, \z) - \tau]_+,
\end{equation}
with $\tau$ such as the distribution sums to 1.
This corresponds to applying $\sparsemax$ to the vector of scores $\mbs$ (where
$s_{\z} = f(x, \z)$) to set the probability distribution of
$\var{Z}$, therefore it will assign zero probabilities to many choices.
The vector
$\bm{\alpha}_\star = \sparsemax(\mbs)$
can be computed in average time $\Theta(K)$; typical implementations
$\Theta(K \log K)$ are sometimes empirically faster
\citep{Condat2016}.
We then can treat such a sparse \(\bm{\alpha}\) as a discrete probability distribution,
setting \(\Pr(\var{Z})=\bm{\alpha}\), or more explicitly,
\(\Pr(\var{Z}=\mbe_i) = \alpha_i\).
We can then compute $\EE_{\bm{\alpha}}[g(x, y, \var{Z})]$ and its gradients with only
$\|\bm{\alpha}_\star\|_0$ calls to $g$. For large enough $K$ and/or expensive $g$,
this can be substantially faster in practice than $K$ calls,
while performing comparably on downstream tasks,
as the following example due to
\citet{correia2020EfficientMarginalizationDiscrete}
shows.
\begin{myexample}{\bfseries Single-token communication game.}
Emergent communication between interacting agents
can be formulated as a game, or equivalently
a latent variable model \citep{Lazaridou2017,Havrylov2017}:
given a dataset of labeled
images, player 1 and player 2 are both shown the same set of $N$ images, but
arranged in a different order. Player 1 picks an image and then a code word from
a vocabulary of size $K$;
they send the code word to player 2. The
communication is successful if player 2 identifies the correct image, and
unsuccessful otherwise.
\citet{Lazaridou2017} observe near-perfect communication success with SFE and
straight-through estimators for the case $N=2$.
\citet{Havrylov2017} find that with more distractor images and a larger
vocabulary, performance degrades. In particular,
\citet{correia2020EfficientMarginalizationDiscrete} find that
with $N=16,K=256$
none of the benchmarked MC-based methods solve the game.
(The best performing MC model
is based on SFE, reaching 55.36\% communication success rate.
Explicit marginalization can reach 93.37\% but requires 256 decoder calls.
Marginalization over the sparesmax distribution of
\cref{eqn:structured_sparsemax}, however, reaches 93.35\% with only 3.13 decoder
calls on average, illustrating the potential of sparse marginalization for
learning.
\end{myexample}

Despite in practice demonstrating very sparse distributions on average, notice that when $f(x, \mbz)=
f(x, \mbz')$ for any $\mbz,\mbz' \in \cZ$, the vector $\mbs$ is constant and
so $\bm{\alpha}_\star = \frac{1}{|\cZ|} (1, \ldots, 1)^\top$.
Thus, in the worst case, it is possible for sparsemax
to be as dense as softmax, rendering this method slow or even intractable for
large or structured $\cZ$. Below we discuss two generalizations that help in
this scenario.

\paragraph{Top-k sparsemax.}
A way to bound the sparsity of sparsemax solutions is by introducing a
non-convex $\ell_0$ constraint:
\begin{equation}
\sparsemax_{(k)}(\bm{s}) \coloneqq
\argmin_{\bm{\alpha} \in \simplex^K, \|\bm{\alpha}\|_0 \leq k} \frac{1}{2} \|
\bm{s} - \bm{\alpha} \|^2.
\end{equation}
Despite the non-convexity, \citet{kyrillidis2013sparse} show that an optimal solution can be
found by applying sparsemax to the $k$-dimensional vector of the highest scores.
Formally we may define an operator $\operatorname{top}_{(k)} : \bbR^K \to \bbR^K$ such that
$[\operatorname{top}_{(k)}(\mbs)]_i = s_i$ if $i$ is among a set of $k$
indices with score greater than or equal to all others,
and $-\infty$ otherwise. But in practice all but the top $k$ indices can be
ignored during computation.
In the unstructured case, finding the top $k$ indices can be done with a
selection algorithm in $O(n + k \log k)$. In the structured case, there is no
general strategy, but specialized algorithms are available for many structures of interest.

Crucially, if the $\ell_0$ constraint is
loose, (\ie, the solution has strictly fewer than $k$ nonzeros), then for this
particular $\mbs$, $\sparsemax_{(k)}(\mbs)=\sparsemax(\mbs)$, and thus we have found an optimal solution much faster.
Remarkably, in the structured case, this means we can sometimes compute exact
expectations of arbitrary functions over exponentially many structures.

On one hand, this observation can lead to another algorithm for sparsemax
that is more efficient in the average case: start with $k$ set to a guess of
what $\|\sparsemax(\mbs)\|_0$ might be, and compute
$\bm{\alpha} =\sparsemax_{(k)}(\mbs)$. If $\|\bm{\alpha}\|_0<k$, we found the
solution. Otherwise, increase $k$ (\eg, doubling) and repeat.
This strategy was proposed in the unstructured case by \citet{sparseseq}.

On the other hand,
\citet{correia2020EfficientMarginalizationDiscrete} propose to
directly use $\sparsemax_{(k)}(\mbs)$ for marginalizing over latent variables,
without increasing $k$ in the case of suboptimality. This way we may control
the computational complexity, at the cost of biasing the estimation toward the
top. In practice, \citet{correia2020EfficientMarginalizationDiscrete} find
that, provided sufficiently high $k$, after some training iterations, the encoder $f$ becomes more confident
and the estimation transitions from approximate to exact.
This algorithm can be applied to structured latent variables as well, as long as
a structured top-k oracle is available.

\paragraph{SparseMAP marginalization.}
Under
\cref{assumption:parts},
recall we have
\[
f(x, \z) = \DP{\z}{\mbs(x; \zparam)}
\]
where the vector-valued function $\mbs$ returns a vector of part scores.
As described in \cref{sec:meanreg}, SparseMAP uses a quadratic regularization
on the mean vector $\bmu = \bbE_{\var{Z}\sim\bm{\alpha}}[\var{Z}]$,
interpreting \(\bm{\alpha} \in \simplex_{|\cZ|}\) as the distribution of
\(\var{Z}\).
We may equivalently rewrite the SparseMAP optimization problem in terms of
$\bm{\alpha}$:
\begin{equation}
\begin{gathered}
\sparsemap(\mbs) = \bbE_{\var{Z}\sim\bm{\alpha}_\star}[\var{Z}],
\\ \text{where} \quad
\bm{\alpha}_\star \in
\argmax_{\bm{\alpha} \in \simplex^{|\cZ|}}
 \DP{\bbE_{\var{Z}\sim\bm{\alpha}}[\var{Z}]}{\bm{s}}
- \frac{1}{2} \| \bbE_{\var{Z}\sim\bm{\alpha}}[\var{Z}] \|^2
\,.
\end{gathered}
\end{equation}
The active set algorithm \citep{nocedalwright} proposed by \citet{sparsemap}
for SparseMAP returns not only the unique solution $\sparsemap(\mbs)$
but also a corresponding
sparse $\bm{\alpha}_\star$ and a sparse generalized Jacobian of
\(\bm{\alpha}_\star\) \wrt
\(\mbs\),
\citep{correia2020EfficientMarginalizationDiscrete}, allowing the use
SparseMAP instead of sparsemax for sparse marginalizing over $\bm{\alpha}_\star$.
The non-uniqueness of $\bm{\alpha}_\star$ leads to
a bias inherent to the deterministic active set algorithm,
but this does not seem to impact performance. Alternatively,
\citet{correia2020EfficientMarginalizationDiscrete} propose
mitigating the bias by turning active set into a stochastic algorithm by
initializing it with a random structure instead of the MAP structure, leading
to a (non-uniform) sample from the set
$\{ \bm{\alpha} : \bbE_{\var{Z}\sim\bm{\alpha}}[\var{Z}] = \sparsemap(\mbs) \}$.
If we denote by \(\Pr(\bm{\alpha})\) the implicit sampling distribution of the
Active Set algorithm, we can then optimize
\[
\EE_{\bm{\alpha}\sim\Pr(\bm{\alpha})} \EE_{\var{Z}\sim\bm{\alpha}} [ g(x, y, Z) ]\,,
\]
with a sampled \(\bm{\alpha}\) from the outer expectation and an explicit sum
over the nonzero \(\z\) for the inner one. The two strategies perform comparably.
Like the top-k sparsemax, SparseMAP exhibits adaptive sparsity:
when the models are very confident, SparseMAP and MAP coincide and then
there is only one possible $\bm{\alpha}_\star$ which is a one-hot vector.
The set of possible $\bm{\alpha}$ is larger earlier in training when models are less certain.
A more specific characterization of the set of possible $\bm{\alpha}$ for a
given $\bm{s}$ in general seems elusive but may be possible for specific
structured models.

Note that in the categorical (one-of-K) case both sparsemax and SparseMAP are
equivalent, since the possible values of $\var{Z}$ are one-hot vectors, so
$\EE_{\var{Z}\sim\bm{\alpha}}[\var{Z}]=\bm{\alpha}$

\section{Summary}
The probabilistic formulation allows us to handle discrete latent structure in
a way that is both fundamentally principled, as well as very flexible, leading to
some of the methods with the lowest assumptions in our text. There is a price to
pay in complexity, as we must always consider entire distributions and not just
punctual values. This translates either to computational complexity (for exact
marginalization) or sample complexity (in the case of Monte Carlo methods).
\citet{correia2020EfficientMarginalizationDiscrete} compare variance-reduced
SFE, Gumbel-ST, top-k sparsemax and SparseMAP on a variational autoencoder with a bit-vector latent
variable, finding relatively similar results in terms of heldout
log-likelihood, but a different distribution in terms of distortion/rate
decomposition in which SFE-based methods perform worse. When applicable,
Gumbel-ST and related methods seem to perform very well in practice.
However, bit vectors are one of the few types of structure for which these
methods all apply. \citet{zantedeschi2022dag} study permutation latent
variables, for which neither SFE nor Gumbel-ST apply, but SparseMAP does.

\chapter{Conclusions}\label{chapter:conclusions}

\section{Overview}
We survey in this text several strategies for learning models with discrete latent
structure.
This sort of structure is relevant for many fields and applications, including natural language processing, computer vision, and bioinformatics,
where data is often well represented by compositional structures
such as trees, sequences, or matchings.
Latent structure models are a powerful tool for learning to extract or induce such
representations, offering a way to incorporate structural bias, discover insight
about the data, and interpret decisions.

We present a range of  different strategies for learning latent discrete structures organized into three main groups:
\begin{itemize}
\item In \cref{chapter:relax} we cover techniques based on continuous
relaxations, where the discrete structure $\mbz$, originally belonging to a
discrete and combinatorial set $\cZ$, and \textit{relaxed} to be part of a
larger continuous and polyhedral space, $\conv(\cZ)$. These techniques relax the forward pass mapping by maximizing a regularized objective, so that they can compute its gradient exactly.
Examples of technique falling into this class include attention mechanisms and their structured counterparts, sparse transformations such as sparsemax and $\alpha$-entmax, as well as their structured counterparts, such as Sinkhorn-based methods and SparseMAP, and perturbation-based regularization methods.
\item In \cref{chapter:surrogate} we present several techniques that sidestep the problem of differentiating through discrete structures by building surrogate gradients.
Surrogate gradient methods provide a way to train deterministic models without
compromising on the use of discrete mappings in the forward pass (\eg, argmax or
quantization layers). Unlike the previous class of methods, the forward pass is left unchanged, but an approximation is made in the backward pass. This approximation can often be justified in
principled ways and can work well in practice, and we attempt to shed some light on these principles.
Surrogate gradient methods typically also require continuity of the decoder $g$ \wrt
\(\mbz\): Even though they use only discrete \(\mbz\) in the
forward pass, the backward pass requires the gradient
\(\pd{g}{3}(x,y,\widehat{\z})\).
Examples of methods falling into this class include straight-through gradient estimation, its structured variants, such as SPIGOT, and quantization methods, such as rounding and vector quantization.
\item Finally, in \cref{chapter:expectation} we focus on techniques based on probabilistic estimation. Unlike the previous methods, this class of techniques give the discrete latent a
probabilistic treatment, allowing us to have correct (unbiased) estimation
without compromising discreteness and without any requirements on \(g\).
This is the class of methods with the lowest assumptions in our text. There is a price to
pay in complexity, as we must always consider entire distributions and not just
punctual values. This translates either to computational complexity (for exact
marginalization) or sample complexity (in the case of Monte Carlo methods).
Examples of methods in this class include explicit marginalization and their sparsifications, the score function estimator (often called REINFORCE), the path gradient estimation (based on the reparametrization trick), structured perturbation methods such as Perturb-and-MAP, the Gumbel-Softmax relaxation, mixed random variables such as Gaussian-Sparsemax, and I-MLE---some of these methods involve also gradient surrogates, as the previous class of methods.
\end{itemize}

Throughout the text, we guide the reader through a unifying view, while highlighting connections, strengths, weaknesses,
assumptions, and emerging recommended practices.
A bird's eye view of the discussed classes of methods and relationships between
them is presented in \cref{tab:overview}.

\section{Implementations and Libraries}

The flexible nature of structure representation makes it generally challenging
to fit latent structure learning inside a neat API framework, and
research using methods we cover usually relies on custom implementations
from scratch.
Readers might find useful the code released alongside the work of
\citet{structured_attn,
baziotis-etal-2019-seq,
mihaylova-etal-2020-understanding},
for instance.
Nevertheless, a few general purpose libraries have been developed.
We list some prominent ones.

\begin{itemize}
\item \href{https://github.com/harvardnlp/pytorch-struct}{torch-struct} \citep{alex2020torchstruct}: a PyTorch library for
differentiable structure prediction, supporting differentiable implementations
of marginal inference for a number of supported structure models.
It therefore
provides building blocks used in \cref{chapter:relax,chapter:surrogate,chapter:expectation}.

\item \href{https://github.com/google-deepmind/synjax}{synjax}
\citep{synjax2023}: a jax library with similar scope as torch-struct;
applicable to \cref{chapter:relax,chapter:surrogate,chapter:expectation}.

\item \href{https://github.com/deep-spin/entmax}{entmax} \citep{sparseseq}: a library implementing efficient sparse
argmax operators (\cref{chapter:expectation}) for the categorical (one-of-K)
case, and thus applicable for incremental structures.

\item \href{https://github.com/deep-spin/lp-sparsemap}{lp-sparsemap}
\citep{lp-sparsemap}: a pytorch frontend for differentiable optimization in
factor graphs. Supports a select set of structures (one-of-K, arborescences, matchings)
and arbitrarily complex models formed by interactions between them.
Provides some of the computational primitives discussed in
\cref{chapter:relax,chapter:surrogate}.

\item \href{https://github.com/cvxgrp/cvxpylayers}{cvxpylayers}
\citep{cvxpylayers2019}. A multi-backend library for differentiable convex
optimization written in a flexible API (disciplined parametrized programs);
useful for exploring and prototyping relaxations and projections from
\cref{chapter:relax,chapter:surrogate}.

\item \href{https://jaxopt.github.io/stable/}{jaxopt}
\citep{jaxopt_implicit_diff}: a jax library for differentiable optimization.
Per \cref{chapter:relax}, relaxed structured latents (as well as some types of
surrogate gradients per \cref{chapter:surrogate}) can be
formulated as regularized constrained optimization problems and thus manually
implemented using this library.

\item \href{https://github.com/HEmile/storchastic}{storchastic} \citep{van2021storchastic}: a PyTorch library for
gradient estimation with discrete latent variables, supporting most methods
we discuss in \cref{chapter:expectation} for the discrete case.
\end{itemize}

%\clearpage%
%\thispagestyle{empty}%
\begin{sidewaystable}%\centering
\footnotesize%
\newcommand\smalltabspace{\addlinespace[0.33em]}%
\newcommand\largetabspace{\addlinespace[0.66em]}%
\newcommand\cvgabbr{cvg.}%
\newcommand\approxabbr{approx.}%
\noindent%
\caption{\label{tab:overview}Overview of methods discussed for latent structure
learning. We abbreviate \(\zsurr_0=\zsurr_0(\s)\defeq \argmax_{\z\in\cZ}\DP{\z}{\s}\).
Convergent (\cvgabbr) learning means the approximation error can be controlled.
}%
\begin{tabular}{l l l l l}
\toprule
Method & Description & Learning & Computation oracle &  Notes \\
\midrule
\multicolumn{5}{l}{\textbf{Relaxation methods} \(g(x,y,\bbE[\var{Z}])\)}
\hfill\emph{\cref{chapter:relax}}
 \\
\smalltabspace
Marginals & \(\bbE_{\Pr(\var{Z})\propto \exp f(\cdot)}[\var{Z}]\) & exact & dedicated algorithm & \\
          &          & \approxabbr & local subproblems  & \eg, belief propagation\\
SparseMAP & \(\proj{\conv(Z)}\) & exact & dedicated algorithm &  \\
          &                   & \cvgabbr & argmax & Active Set algorithm \\
          &                   & \approxabbr & local subproblems & dual decomposition \\
Perturbation    & \(\bbE[\zsurr_0(\s+\var{U})]\)& \cvgabbr & argmax & \\
Sinkhorn   & \( \zsurr_S(\s) \) & \cvgabbr & local subproblems & only assignment/transport \\
\largetabspace
\multicolumn{3}{l}{\textbf{Surrogate gradient methods} \(g(x,y,\zsurr_{0}(\s))\)}
& \textcolor{gray}{fwd pass / bwd pass} &
\hfill\emph{\cref{chapter:surrogate}}
\\
\smalltabspace
ST         &\(\pd{\zsurr}{}(\s)\defeq\mathrm{Id}\)& \approxabbr & argmax/--- &  \\
ST-Softmax &\(\pd{\zsurr}{}(\s)\defeq\pd{\softmax}{}\)& \approxabbr & argmax/\(\partial\)marginals & same oracle
as marginals \\
SPIGOT     &
\(\pd{\zsurr}{}(\mbs)(\zvector)
\defeq
\widehat{\z}_0 - \proj{\conv(\cZ)}(\widehat{\z}_0 - \eta\zvector)\)
& \approxabbr & argmax/projection &  same oracle as
SparseMAP \\
LI         &
\(\pd{\zsurr}{}(\mbs)(\zvector)
\defeq
\nicefrac{1}{\eta} \left(
\widehat{\z}_0(\mbs + \eta\zvector) - \widehat{\z}_0(\s)
\right)\)
& \approxabbr & argmax/argmax &  \\
\largetabspace
\multicolumn{5}{l}{\textbf{Probabilistic latent variable methods}
\(\bbE[g(x,y,\var{Z})]\)}
\hfill\emph{\cref{chapter:expectation}}
\\
\smalltabspace
Path gradients & \(\zsurr_{\text{surr}}(\s + \var{U})\)& \approxabbr & same as chosen surrogate &
incl.\ ST-Gumbel, I-MLE \\
SFE & & \cvgabbr & sampling, marginals & A1, A2 not
required \\
Explicit sum
           &\(\Pr(\z)\defeq\) softmax & exact &---& small \(\cZ\). A1,A2 not req. \\
           &\(\Pr(\z)\defeq\) sparsemax & \approxabbr & top-k & A1,A2 not req.\\
           &\(\Pr(\z)\defeq\) SparseMAP & \approxabbr & argmax & A2 not req.\\
\bottomrule
\end{tabular}%
\end{sidewaystable}

\begin{acknowledgements}
This work was partly supported by
the Dutch Research Council
(NWO) via VI.Veni.212.228 and Zwaartekracht Hybrid Intelligence,
the French Agence Nationale de la Recherche (ANR) via grant ANR-23-CE23-0005 (project SEMIAMOR),
the European Union's Horizon Europe research and innovation programme
via UTTER 101070631,
and the
European Research Council (ERC) via StG DeepSPIN 758969 and CoG DECOLLAGE 101088763, as well as
and FCT/MECI through national funds and when applicable co-funded EU funds under UID/50008: Instituto de Telecomunicacões.
All emojis designed by \href{https://openmoji.org/}{OpenMoji} -- the open-source
emoji and icon project, which is licensed \href{https://creativecommons.org/licenses/by-sa/4.0/}{CC BY-SA 4.0}.

\begin{center}{\Huge \euflag}\end{center}

\end{acknowledgements}

\backmatter%

\printbibliography

@inproceedings{komodakisrc,
  title={MRF optimization via dual decomposition: Message-passing revisited.},
  author={Komodakis, Nikos and Paragios, Nikos and Tziritas, Georgios},
  booktitle={ICCV},
  volume={1},
  number={3},
  pages={5},
  year={2007}
}

@inproceedings{rush,
  title={On dual decomposition and linear programming relaxations for natural language processing},
  author={Rush, Alexander M and Sontag, David and Collins, Michael and Jaakkola, Tommi},
  year={2010},
  organization={Association for Computational Linguistics}
}

@article{kim-tutorial,
  title={A tutorial on deep latent variable models of natural language},
  author={Kim, Yoon and Wiseman, Sam and Rush, Alexander M},
  journal={arXiv preprint arXiv:1812.06834},
  year={2018}
}

@inproceedings{roth2005integer,
  title={Integer linear programming inference for conditional random fields},
  author={Roth, Dan and Yih, Wen-tau},
  booktitle={Proceedings of the 22nd international conference on Machine learning},
  pages={736--743},
  year={2005}
}

@inproceedings{punyakanok2004semantic,
  title={Semantic Role Labeling Via Integer Linear Programming Inference},
  author={Punyakanok, Vasin and Roth, Dan and Yih, Wen-tau and Zimak, Dav},
  booktitle={Proceedings of the 20th International Conference on Computational Linguistics},
  pages={1346--1352},
  year={2004}
}

@article{wainwright,
  title={\href{https://people.eecs.berkeley.edu/~wainwrig/Papers/WaiJor08_FTML.pdf}{Graphical models, exponential families, and variational inference.}},
  author={Wainwright, Martin J and Jordan, Michael I},
  journal_full={Foundations and Trends{\textregistered} in Machine Learning},
  journal={Found. Trends Mach. Learn.},
  volume={1},
  number={1--2},
  pages={1--305},
  year={2008},
  publisher={Now Publishers, Inc.}
}

@book{jurafsky-martin,
  title={\href{https://web.stanford.edu/~jurafsky/slp3/}{Speech and Language Processing (3rd ed.)}},
  author={Jurafsky, Dan and Martin, James H},
  year={2018},
  publisher={draft}
}

@article{kg,
  title={\href{https://aclweb.org/anthology/Q16-1023}{Simple and accurate dependency parsing using bidirectional LSTM feature representations}},
  author={Kiperwasser, Eliyahu and Goldberg, Yoav},
  journal={TACL},
  volume={4},
  pages={313--327},
  year={2016}
}

@inproceedings{sparseattn,
  title={\href{https://arxiv.org/abs/1705.07704}{A regularized framework for
    sparse and structured neural attention}},
  author={Niculae, Vlad and Blondel, Mathieu},
  booktitle={Proc. of NIPS},
  year={2017}
}

@inproceedings{sparsemap,
  title={\href{https://arxiv.org/abs/1802.04223}{SparseMAP: Differentiable sparse structured inference}},
  author={Niculae, Vlad and Martins, Andr{\'e} FT and Blondel, Mathieu and Cardie, Claire},
  booktitle={Proc. of ICML},
  year={2018}
}

@InProceedings{lp-sparsemap,
	title = 	 {{LP}-{S}parse{MAP}: Differentiable Relaxed Optimization for Sparse Structured Prediction},
	author =       {Niculae, Vlad and Martins, Andre},
	booktitle = 	 {Proceedings of ICML},
	year = 	 {2020},
	editor = 	 {III, Hal Daumé and Singh, Aarti},
	volume = 	 {119},
	series = 	 {Proceedings of Machine Learning Research},
	publisher =    {PMLR},
	url = 	 {https://proceedings.mlr.press/v119/niculae20a.html},
}

@article{kakade_jmlr,
  title={\href{https://arxiv.org/abs/0910.0610}
         {Regularization techniques for learning with matrices}},
  author={Kakade, Sham M and Shalev-Shwartz, Shai and Tewari, Ambuj},
  journal={Journal of Machine Learning Research},
  volume={13},
  pages={1865--1890},
  year={2012}
}

@book{zalinescu,
  title={\href{http://www.worldscientific.com/worldscibooks/10.1142/5021}
    {Convex Analysis in General Vector Spaces}},
  author={Z{\u{a}}linescu, Constantin},
  year={2002},
  publisher={World Scientific}
}

@inproceedings{smooth_and_strong,
 author = {Meshi, Ofer and Mahdavi, Mehrdad and Schwing, Alexander G},
 title =
 {\href{https://papers.nips.cc/paper/5710-smooth-and-strong-map-inference-with-linear-convergence}
 {Smooth and strong: {MAP} inference with linear convergence}},
 booktitle = {Proc. of NIPS},
 year = {2015},
}

@inproceedings{sparsemax,
  title={\href{https://arxiv.org/abs/1602.02068}
    {From softmax to sparsemax: A sparse model of attention and multi-label
    classification}},
  author={Martins, Andr{\'e} FT and Astudillo, Ram{\'o}n Fernandez},
  booktitle={Proc. of ICML},
  year={2016}
}

@inproceedings{duchi,
  title={\href{https://stanford.edu/~jduchi/projects/DuchiShSiCh08.pdf}
         {Efficient projections onto the $\ell_1$-ball for learning in high
         dimensions}},
  author={Duchi, John C and Shalev-Shwartz, Shai and Singer, Yoram and Chandra, Tushar},
  booktitle={Proc. of ICML},
  year={2008},
}

@article{Condat2016,
  title={\href{https://hal.archives-ouvertes.fr/hal-01056171}{Fast projection onto the simplex and the $\ell_1$ ball}},
  author={Condat, Laurent},
  journal={Mathematical Programming},
  volume={158},
  number={1-2},
  pages={575--585},
  year={2016},
  publisher={Springer}
}

@article{Brucker1984,
  title={\href{https://www.sciencedirect.com/science/article/pii/0167637784900105}{An
         $O(n)$ algorithm for quadratic knapsack problems}},
  author={Brucker, Peter},
  journal={Operations Research Letters},
  volume={3},
  number={3},
  pages={163--166},
  year={1984},
  publisher={Elsevier}
}

@article{Held1974,
  title={\href{https://link.springer.com/article/10.1007/BF01580223}{Validation
         of subgradient optimization}},
  author={Held, Michael and Wolfe, Philip and Crowder, Harlan P},
  journal={Mathematical Programming},
  volume={6},
  number={1},
  pages={62--88},
  year={1974},
  publisher={Springer}
}

@article{hungarian,
  title={\href{http://onlinelibrary.wiley.com/doi/10.1002/nav.3800020109/abstract}{The
         Hungarian method for the assignment problem}},
  author={Kuhn, Harold W},
  journal={Nav. Res. Log.},
  volume={2},
  number={1-2},
  pages={83--97},
  year={1955},
  publisher={Wiley Online Library}
}

@article{lapjv,
  title={\href{https://link.springer.com/article/10.1007/BF02278710}{A shortest
         augmenting path algorithm for dense and sparse linear assignment
         problems}},
  author={Jonker, Roy and Volgenant, Anton},
  journal={Computing},
  volume={38},
  number={4},
  pages={325--340},
  year={1987},
  publisher={Springer}
}

@article{birkhoff,
  title={Tres observaciones sobre el algebra lineal},
  author={Birkhoff, Garrett},
  journal={Univ. Nac. Tucum{\'a}n Rev. Ser. A},
  volume={5},
  pages={147--151},
  year={1946}
}

@inproceedings{gumbel_softmax,
  title={\href{https://arxiv.org/abs/1611.01144}
         {Categorical reparameterization with {G}umbel-{S}oftmax}},
  author={Jang, Eric and Gu, Shixiang and Poole, Ben},
  booktitle={Proc. of ICLR},
  year={2017}
}

@inproceedings{concrete_distribution,
  title={\href{https://arxiv.org/abs/1611.00712}
    {The concrete distribution: A continuous relaxation of discrete random
    variables}},
  author={Maddison, Chris J and Mnih, Andriy and Teh, Yee Whye},
  booktitle={Proc. of ICLR},
  year={2017}
}

@inproceedings{conditional_bengio,
  title={\href{https://arxiv.org/abs/1305.2982}
         {Estimating or propagating gradients through stochastic neurons for
         conditional computation}},
  author={Bengio, Yoshua and L{\'e}onard, Nicholas and Courville, Aaron},
  booktitle={Proc. of NIPS},
  year={2013}
}

@inproceedings{structured_attn,
  title={\href{https://arxiv.org/abs/1702.00887}{Structured attention networks}},
  author={Kim, Yoon and Denton, Carl and Hoang, Loung and Rush, Alexander M},
  booktitle={Proc. of ICLR},
  year={2017}
}

@article{lapata,
  title={\href{https://arxiv.org/abs/1705.09207}{Learning structured text representations}},
  author={Liu, Yang and Lapata, Mirella},
  journal={TACL},
  volume={6},
  pages = {63--75},
  year={2018}
}

@article{ad3,
  title={\href{http://jmlr.org/papers/v16/martins15a.html}{AD3: Alternating directions dual decomposition for MAP inference in graphical models}},
  author={Martins, Andr{\'e} FT and Figueiredo, M{\'a}rio AT and Aguiar, Pedro MQ and Smith, Noah A and Xing, Eric P},
  journal={JMLR},
  volume={16},
  number={1},
  pages={495--545},
  year={2015},
}

@article{Chu1965,
author = {Chu, Yoeng-Jin and Liu, Tseng-Hong},
title = "On the Shortest Arborescence of a Directed Graph",
year = "1965",
volume = "14",
journal = "Science Sinica",
pages = "1396--1400"}

@article{Edmonds1967,
author = {Edmonds, Jack},
year = "1967",
x_title = {\href{https://doi.org/10.6028\%2Fjres.071b.032}{Optimum branchings}},
title = {\href{https://doi.org/10.6028/jres.071b.032}{Optimum branchings}},
journal_full = "Journal of Research of the National Bureau of Standards",
journal={J. Res. Nat. Bur. Stand.},
volume = "71B",
pages = "233--240"}

@article{Kirchhoff1847,
  title={Ueber die Aufl{\"o}sung der Gleichungen, auf welche man bei der Untersuchung der linearen Vertheilung galvanischer Str{\"o}me gef{\"u}hrt wird},
  author={Kirchhoff, Gustav},
  journal={Annalen der Physik},
  volume={148},
  number={12},
  pages={497--508},
  year={1847},
  publisher={Wiley Online Library}
}

@inproceedings{adam,
  title={\href{https://arxiv.org/abs/1412.6980}{Adam: A method for stochastic optimization}},
  author={Kingma, Diederik and Ba, Jimmy},
  booktitle={Proc. of ICLR},
  year={2015}
}

@inproceedings{Li2009,
  title={\href{http://www.mt-archive.info/EMNLP-2009-Li.pdf}{First-and second-order expectation semirings with applications to minimum-risk training on translation forests}},
  author={Li, Zhifei and Eisner, Jason},
  booktitle={Proc. of EMNLP},
  year={2009},
}

@book{bertsekas-nonlin,
  title={\href{http://www.athenasc.com/nonlinbook.html}{Nonlinear Programming}},
  author={Bertsekas, Dimitri P},
  year={1999},
  publisher={Athena Scientific Belmont}
}

@inproceedings{andre-concise,
  title={\href{http://www.aclweb.org/anthology/P09-1039}{Concise integer linear programming formulations for dependency parsing}},
  author={Martins, Andr{\'e} FT and Smith, Noah A and Xing, Eric P},
  booktitle={Proc. of ACL-IJCNLP},
  year={2009}
}

@phdthesis{taskar-thesis,
    title    = {\href{https://homes.cs.washington.edu/~taskar/pubs/thesis.pdf}{Learning structured prediction models: A large margin approach}},
    school   = {Stanford University},
    author   = {Ben Taskar},
    year     = {2004}
}

@inproceedings{cg,
  title={\href{https://arxiv.org/abs/1511.05932}{On the global linear convergence of Frank-Wolfe optimization variants}},
  author={Lacoste-Julien, Simon and Jaggi, Martin},
  booktitle={Proc. of NIPS},
  year={2015}
}

@article{lstm,
  title={\href{http://www.bioinf.jku.at/publications/older/2604.pdf}{Long
         short-term memory}},
  author={Hochreiter, Sepp and Schmidhuber, J{\"u}rgen},
  journal={Neural Computation},
  volume={9},
  number={8},
  pages={1735--1780},
  year={1997}
}

@misc{pytorch,
  author = {PyTorch},
  year = {2017},
  title = {\url{http://pytorch.org}}
}

@book{nocedalwright,
  title={\href{https://doi.org/10.1007/b98874}{Numerical Optimization}},
  author={Nocedal, Jorge and Wright, Stephen},
  year={1999},
  publisher={Springer New York}
}

@article{fw,
  title={\href{https://doi.org/10.1002/nav.3800030109}{An algorithm for quadratic programming}},
  author={Frank, Marguerite and Wolfe, Philip},
  journal={Nav. Res. Log.},
  volume={3},
  number={1-2},
  pages={95--110},
  year={1956},
  publisher={Wiley Online Library}
}

@InProceedings{ves,
  title = {\href{http://proceedings.mlr.press/v15/stoyanov11a.html}{Empirical
    risk minimization of graphical model parameters given approximate inference,
    decoding, and model structure}},
  author = {Veselin Stoyanov and Alexander Ropson and Jason Eisner},
  booktitle = {Proc. of AISTATS},
  year = 	 {2011},
}

@Article{Rabiner1989,
 author =      "Lawrence R. Rabiner",
 title ={\href{https://doi.org/10.1109/5.18626}{A tutorial on Hidden Markov
         Models and selected
                applications in speech recognition}},
 journal =     "P. IEEE",
 year =        "1989",
 volume =      "77",
 number =      "2",
 pages =       "257--286",
}

@article{mnp,
  title={\href{https://link.springer.com/article/10.1007/BF01580381}{Finding the
         nearest point in a polytope}},
  author={Wolfe, Philip},
  journal={Mathematical Programming},
  volume={11},
  number={1},
  pages={128--149},
  year={1976},
  publisher={Springer}
}

@article{dantzig,
  title={\href{https://msp.org/pjm/1955/5-2/pjm-v5-n2-s.pdf}{The generalized simplex method for minimizing a linear form under linear inequality restraints}},
  author={Dantzig, George B and Orden, Alex and Wolfe, Philip},
  journal={Pacific Journal of Mathematics},
  volume={5},
  number={2},
  pages={183--195},
  year={1955}
}

@book{cohen2019bayesian,
	title={\href{https://doi.org/10.1007/978-3-031-02170-1}{Bayesian Analysis in Natural Language Processing}},
	author={Cohen, Shay},
	series={Synth. Lect. Human Lang. Technol.},
	seriesf = {Synthesis Lectures on Human Language Technologies},
	year={2019},
	publisher={Morgan \& Claypool}
}

@article{fylosses,
  title={Learning with {F}enchel-{Y}oung losses},
  author={Blondel, Mathieu and Martins, Andr{\'e} FT and Niculae, Vlad},
  journal={Journal of Machine Learning Research},
  volume={21},
  number={35},
  pages={1--69},
  year={2020}
}

@inproceedings{arthurdp,
  title={\href{https://arxiv.org/abs/1802.03676}{Differentiable dynamic programming for structured prediction
         and attention}},
  author={Mensch, Arthur and Blondel, Mathieu},
  booktitle={Proc. of ICML},
  year={2018}
}

@inproceedings{mena,
  title={\href{https://arxiv.org/abs/1802.08665}{Learning latent permutations with Gumbel-Sinkhorn networks}},
  author={Mena, Gonzalo and Belanger, David and Linderman, Scott and Snoek, Jasper},
  booktitle={Proc. of ICLR},
  year={2018}
}

@InProceedings{vinyes,
  title = {\href{http://proceedings.mlr.press/v54/vinyes17a.html}{
    Fast column generation for atomic norm regularization}},
  author = {Marina Vinyes and Guillaume Obozinski},
  booktitle = {Proc. of AISTATS},
  year = {2017},
}

@article{vernet_2016,
  title={\href{http://jmlr.org/papers/v17/14-294.html}{Composite multiclass losses}},
  author={Williamson, Robert C and Vernet, Elodie and Reid, Mark D},
  journal={Journal of Machine Learning Research},
  year={2016}
}

@inproceedings{drozdov_coadapt,
    author={Drozdov, Andrew and Bowman, Samuel},
    title={The coadaptation problem when learning how and what to compose},
    booktitle={Proc of ReplNLP},
    year={2017}
}

@book{borwein_lewis_convex_analysis,
	title={Convex analysis and nonlinear optimization: theory and examples},
	author={Borwein, Jonathan and Lewis, Adrian S},
	year={2010},
	publisher={Springer Science \& Business Media}
}

@article{bertsekas2000gradient,
  title={Gradient convergence in gradient methods with errors},
  author={Bertsekas, Dimitri P and Tsitsiklis, John N},
  journal={SIAM Journal on Optimization},
  volume={10},
  number={3},
  pages={627--642},
  year={2000},
  publisher={SIAM}
}

@book{barber,
  title={Bayesian reasoning and machine learning},
  author={Barber, David},
  year={2012},
  publisher={Cambridge University Press}
}

@article{read2011classifier,
  title={Classifier chains for multi-label classification},
  author={Read, Jesse and Pfahringer, Bernhard and Holmes, Geoff and Frank, Eibe},
  journal={Machine learning},
  volume={85},
  number={3},
  pages={333--359},
  year={2011},
  publisher={Springer}
}

@article{cohen-spurious,
  title={Elimination of Spurious Ambiguity in Transition-Based Dependency Parsing},
  author={Cohen, Shay B and G{\'o}mez-Rodr{\i}guez, Carlos and Satta, Giorgio},
  journal={preprint arXiv:1206.6735},
  year={2012}
}

@article{kool-jmlr,
  author  = {Wouter Kool and Herke van Hoof and Max Welling},
  title   = {Ancestral Gumbel-Top-k Sampling for Sampling Without Replacement},
  journal = {Journal of Machine Learning Research},
  year    = {2020},
  volume  = {21},
  number  = {47},
  pages   = {1-36},
  url     = {http://jmlr.org/papers/v21/19-985.html}
}

@inproceedings{sparseseq,
    title={\href{https://arxiv.org/abs/1905.05702}{Sparse sequence-to-sequence models}},
    author={Ben Peters and Niculae, Vlad and Martins, Andr{\'e} FT},
    year=2019,
    booktitle={Proc. ACL},
}

@inproceedings{liu-etal-2018-structured,
    title = "Structured Alignment Networks for Matching Sentences",
    author = "Liu, Yang  and
      Gardner, Matt  and
      Lapata, Mirella",
    booktitle = "Proceedings of EMNLP",
    year = "2018",
    publisher = "Association for Computational Linguistics",
    url = "https://aclanthology.org/D18-1184",
    doi = "10.18653/v1/D18-1184"
}

@article{adams2011ranking,
  title={Ranking via sinkhorn propagation},
  author={Adams, Ryan Prescott and Zemel, Richard S},
  journal={preprint arXiv:1106.1925},
  year={2011}
}

@inproceedings{tay2020sparse,
  title={Sparse {S}inkhorn attention},
  author={Tay, Yi and Bahri, Dara and Yang, Liu and Metzler, Donald and Juan, Da-Cheng},
  booktitle={Proceedings of ICML},
  year={2020},
}

@article{sinkhorn1964relationship,
  title={A relationship between arbitrary positive matrices and doubly stochastic matrices},
  author={Sinkhorn, Richard},
  journal={The annals of mathematical statistics},
  volume={35},
  number={2},
  pages={876--879},
  year={1964},
  publisher={JSTOR}
}

@inproceedings{cuturi2013sinkhorn,
  title={Sinkhorn distances: Lightspeed computation of optimal transport},
  author={Cuturi, Marco},
  booktitle={Proceedings of NeurIPS},
  year={2013}
}

@article{compot,
  title={Computational optimal transport},
  author={Peyr{\'e}, Gabriel and Cuturi, Marco},
  journal={Foundations and Trends{\textregistered} in Machine Learning},
  volume={11},
  number={5-6},
  pages={355--607},
  year={2019},
  publisher={Now Publishers, Inc.}
}

@inproceedings{albert,
title={{ALBERT}: A Lite {BERT} for Self-supervised Learning of Language Representations},
author={Zhenzhong Lan and Mingda Chen and Sebastian Goodman and Kevin Gimpel and Piyush Sharma and Radu Soricut},
booktitle={Proceedings of ICLR},
year={2020},
url={https://openreview.net/forum?id=H1eA7AEtvS}
}

@techreport{kasami,
	title={An efficient recognition and syntax-analysis algorithm for context-free languages},
	author={Kasami, Tadao},
	year={1966},
	publisher={University of Illinois}
}

@article{younger,
	title = {Recognition and parsing of context-free languages in time n3},
	journal = {Information and Control},
	volume = {10},
	number = {2},
	pages = {189-208},
	year = {1967},
	issn = {0019-9958},
	doi = {https://doi.org/10.1016/S0019-9958(67)80007-X},
	url = {https://www.sciencedirect.com/science/article/pii/S001999586780007X},
	author = {Daniel H. Younger}
}

@techreport{cocke,
	title={Programming languages and their compilers: preliminary notes},
	year={1970},
	author={Cocke, John and Schwartz, Jacob T.},
	publisher={Courant Institute of Mathematical Sciences, New York University}
}

@article{insideoutside,
	author = {Baker,J. K. },
	title = {Trainable grammars for speech recognition},
	journal = {The Journal of the Acoustical Society of America},
	volume = {65},
	number = {S1},
	pages = {S132-S132},
	year = {1979},
	doi = {10.1121/1.2017061},
	
	URL = { 
	https://doi.org/10.1121/1.2017061
	
	},
	eprint = { 
	https://doi.org/10.1121/1.2017061
	
	}
	
}

@ARTICLE{dtw, 
	author={Sakoe, H. and Chiba, S.}, 
	journal={IEEE Transactions on Acoustics, Speech, and Signal Processing}, 
	title={Dynamic programming algorithm optimization for spoken word recognition}, 
	year={1978},
	volume={26},
	number={1},
	pages={43-49},
	doi={10.1109/TASSP.1978.1163055}
}

@InProceedings{softdtw,
	title = 	 {Soft-{DTW}: a Differentiable Loss Function for Time-Series},
	author =       {Marco Cuturi and Mathieu Blondel},
	booktitle = 	 {Proceedings of ICML},
	year = 	 {2017},
	editor = 	 {Precup, Doina and Teh, Yee Whye},
	volume = 	 {70},
	series = 	 {Proceedings of Machine Learning Research},
	publisher =    {PMLR},
	url = 	 {https://proceedings.mlr.press/v70/cuturi17a.html},
}

@article{valiant,
	title = {The complexity of computing the permanent},
	journal = {Theoretical Computer Science},
	volume = {8},
	number = {2},
	pages = {189--201},
	year = {1979},
	issn = {0304-3975},
	doi = {https://doi.org/10.1016/0304-3975(79)90044-6},
	url = {https://www.sciencedirect.com/science/article/pii/0304397579900446},
	author = {L.G. Valiant}
}

@book{garey1979computers,
	title={Computers and intractability},
	author={Garey, Michael R and Johnson, David S},
	volume={174},
	year={1979},
	publisher={freeman San Francisco}
}

@inproceedings{transformers,
	author = {Vaswani, Ashish and Shazeer, Noam and Parmar, Niki and Uszkoreit, Jakob and Jones, Llion and Gomez, Aidan N and Kaiser, \L ukasz and Polosukhin, Illia},
	booktitle = {Advances in Neural Information Processing Systems},
	editor = {I. Guyon and U. V. Luxburg and S. Bengio and H. Wallach and R. Fergus and S. Vishwanathan and R. Garnett},
	pages = {},
	publisher = {Curran Associates, Inc.},
	title = {Attention is All you Need},
	url = {https://proceedings.neurips.cc/paper/2017/file/3f5ee243547dee91fbd053c1c4a845aa-Paper.pdf},
	volume = {30},
	year = {2017}
}

@inproceedings{seq2seq_sutskever,
	author = {Sutskever, Ilya and Vinyals, Oriol and Le, Quoc V},
	booktitle = {Advances in Neural Information Processing Systems},
	editor = {Z. Ghahramani and M. Welling and C. Cortes and N. Lawrence and K. Q. Weinberger},
	pages = {},
	publisher = {Curran Associates, Inc.},
	title = {Sequence to Sequence Learning with Neural Networks},
	url = {https://proceedings.neurips.cc/paper/2014/file/a14ac55a4f27472c5d894ec1c3c743d2-Paper.pdf},
	volume = {27},
	year = {2014}
}

@inproceedings{seq2seq_cho,
	title = "Learning Phrase Representations using {RNN} Encoder{--}Decoder for Statistical Machine Translation",
	author = {Cho, Kyunghyun  and
	van Merri{\"e}nboer, Bart  and
	Gulcehre, Caglar  and
	Bahdanau, Dzmitry  and
	Bougares, Fethi  and
	Schwenk, Holger  and
	Bengio, Yoshua},
	booktitle = "Proceedings of the 2014 Conference on Empirical Methods in Natural Language Processing ({EMNLP})",
	month = oct,
	year = "2014",
	address = "Doha, Qatar",
	publisher = "Association for Computational Linguistics",
	url = "https://aclanthology.org/D14-1179",
	doi = "10.3115/v1/D14-1179",
	pages = "1724--1734",
}

@inproceedings{wong2021leveraging,
  title={Leveraging language to learn program abstractions and search heuristics},
  author={Wong, Catherine and Ellis, Kevin M and Tenenbaum, Joshua and Andreas, Jacob},
  booktitle={Proceedings of ICML},
  year={2021},
  organization={PMLR}
}

@inproceedings{correia2019adaptively,
  title={Adaptively Sparse Transformers},
  author={Correia, Gon{\c{c}}alo M and Niculae, Vlad and Martins, Andr{\'e} FT},
  booktitle={Proceedings of EMNLP-IJCNLP},
  year={2019}
}

@article{domke,
  title={\href{https://arxiv.org/abs/1301.3193}{Learning graphical model parameters with approximate marginal inference}},
  author={Domke, Justin},
journal = {IEEE Transactions on Pattern Analysis and Machine Intelligence},
  volume={35},
  number={10},
  pages={2454--2467},
  year={2013},
  publisher={IEEE}
}

@inproceedings{liang2010learning,
  title={Learning programs: A hierarchical {B}ayesian approach},
  author={Liang, Percy and Jordan, Michael I and Klein, Dan},
  booktitle={Proceedings of ICML},
  year={2010}
}

@inproceedings{zmigrod-etal-2021-efficient,
    title = "Efficient Sampling of Dependency Structure",
    author = "Zmigrod, Ran  and
      Vieira, Tim  and
      Cotterell, Ryan",
    booktitle = "Proceedings of EMNLP",
    month = nov,
    year = "2021",
    address = "Online and Punta Cana, Dominican Republic",
    publisher = "Association for Computational Linguistics",
    url = "https://aclanthology.org/2021.emnlp-main.824",
    doi = "10.18653/v1/2021.emnlp-main.824",
}

@inproceedings{caio-iclr,
  title={\href{https://arxiv.org/abs/1807.09875}{Differentiable Perturb-and-Parse: Semi-Supervised Parsing with a Structured Variational Autoencoder}},
  author={Corro, Caio and Titov, Ivan},
  booktitle={Proc. of ICLR},
  year={2019}
}

@inproceedings{caio-acl,
author = {Corro, Caio and Titov, Ivan},
booktitle = {Proc. of ACL},
title = {\href{https://www.aclweb.org/anthology/P19-1551/}{Learning latent trees with stochastic perturbations and differentiable dynamic programming}},
year = {2019},
}

@inproceedings{papandreou2011perturb,
  title={Perturb-and-map random fields: Using discrete optimization to learn and sample from energy models},
  author={Papandreou, George and Yuille, Alan L},
  booktitle={2011 International Conference on Computer Vision},
  pages={193--200},
  year={2011},
  organization={IEEE}
}

@inproceedings{hazan2012partition,
  title={On the partition function and random maximum a-posteriori perturbations},
  author={Hazan, Tamir and Jaakkola, Tommi},
  booktitle={Proceedings of ICML},
  year={2012}
}

@inproceedings{hazan2013gibbs,
 author = {Hazan, Tamir and Maji, Subhransu and Jaakkola, Tommi},
 booktitle = {Advances in Neural Information Processing Systems},
 editor = {C.J. Burges and L. Bottou and M. Welling and Z. Ghahramani and K.Q. Weinberger},
 pages = {},
 publisher = {Curran Associates, Inc.},
 title = {On Sampling from the Gibbs Distribution with Random Maximum A-Posteriori Perturbations},
 url = {https://proceedings.neurips.cc/paper/2013/file/443cb001c138b2561a0d90720d6ce111-Paper.pdf},
 volume = {26},
 year = {2013}
}

@inproceedings{keith-etal-2018-monte,
    title = "{M}onte {C}arlo Syntax Marginals for Exploring and Using Dependency Parses",
    author = "Keith, Katherine  and
      Blodgett, Su Lin  and
      O{'}Connor, Brendan",
    booktitle = "Proceedings of the 2018 Conference of the North {A}merican Chapter of the Association for Computational Linguistics: Human Language Technologies, Volume 1 (Long Papers)",
    month = jun,
    year = "2018",
    address = "New Orleans, Louisiana",
    publisher = "Association for Computational Linguistics",
    url = "https://aclanthology.org/N18-1084",
    doi = "10.18653/v1/N18-1084",
}

@inproceedings{kingma2013auto,
  title={Auto-encoding {V}ariational {B}ayes},
  author={Kingma, Diederik P and Welling, Max},
  booktitle={Proceedings of ICLR},
  year={2014}
}

@inproceedings{kingma2014semisup,
 author = {Kingma, Durk P and Mohamed, Shakir and Jimenez Rezende, Danilo and Welling, Max},
 booktitle = {Advances in Neural Information Processing Systems},
 editor = {Z. Ghahramani and M. Welling and C. Cortes and N. Lawrence and K.Q. Weinberger},
 pages = {},
 publisher = {Curran Associates, Inc.},
 title = {Semi-supervised Learning with Deep Generative Models},
 url = {https://proceedings.neurips.cc/paper/2014/file/d523773c6b194f37b938d340d5d02232-Paper.pdf},
 volume = {27},
 year = {2014}
}

@InProceedings{rezende2014dgm,
	title = 	 {Stochastic Backpropagation and Approximate Inference in Deep Generative Models},
	author = 	 {Rezende, Danilo Jimenez and Mohamed, Shakir and Wierstra, Daan},
	booktitle = 	 {Proceedings of the 31st International Conference on Machine Learning},
	pages = 	 {1278--1286},
	year = 	 {2014},
	editor = 	 {Xing, Eric P. and Jebara, Tony},
	volume = 	 {32},
	number =       {2},
	series = 	 {Proceedings of Machine Learning Research},
	address = 	 {Bejing, China},
	month = 	 {22--24 Jun},
	publisher =    {PMLR},
	url = 	 {https://proceedings.mlr.press/v32/rezende14.html}
}

@article{bertsekas1988auction,
  title={The auction algorithm: A distributed relaxation method for the assignment problem},
  author={Bertsekas, Dimitri P},
  journal={Annals of Operations Research},
  volume={14},
  number={1},
  pages={105--123},
  year={1988},
  publisher={Springer}
}

@inproceedings{sagae2005shiftreduce,
	title = "\href{https://aclanthology.org/W05-1513}{A Classifier-Based Parser with Linear Run-Time Complexity}",
	author = "Sagae, Kenji  and
	Lavie, Alon",
	booktitle = "Proceedings of the Ninth International Workshop on Parsing Technology",
	month = oct,
	year = "2005",
	address = "Vancouver, British Columbia",
	publisher = "Association for Computational Linguistics",
	pages = "125--132",
}

@inproceedings{stanojevic2022sampling,
    title = "Unbiased and Efficient Sampling of Dependency Trees",
    author = "Stanojevi{\'c}, Milo{\v{s}}",
    editor = "Goldberg, Yoav  and
      Kozareva, Zornitsa  and
      Zhang, Yue",
    booktitle = "Proceedings of the 2022 Conference on Empirical Methods in Natural Language Processing",
    month = dec,
    year = "2022",
    address = "Abu Dhabi, United Arab Emirates",
    publisher = "Association for Computational Linguistics",
    url = "https://aclanthology.org/2022.emnlp-main.110",
    doi = "10.18653/v1/2022.emnlp-main.110",
    pages = "1691--1706",
}

@article{monte_carlo_grad,
  title={Monte Carlo Gradient Estimation in Machine Learning.},
  author={Mohamed, Shakir and Rosca, Mihaela and Figurnov, Michael and Mnih, Andriy},
  journal={J. Mach. Learn. Res.},
  volume={21},
  number={132},
  pages={1--62},
  year={2020}
}

@article{rubinstein1976monte,
  title={A {Monte Carlo} method for estimating the gradient in a stochastic network},
  author={Rubinstein, RY},
  journal={Unpublished manuscript, Technion, Haifa, Israel},
  year={1976}
}

@inproceedings{paisley2012variational, 
author = {Paisley, John and Blei, David M. and Jordan, Michael I.}, 
title = {\href{https://arxiv.org/abs/1206.6430}{Variational bayesian inference with stochastic search}}, 
year = {2012}, 
booktitle = {Proc. ICML}
}

@article{Williams1992,
author = {Williams, Ronald J.},
journal = {Machine Learning},
number = {3-4},
pages = {229--256},
publisher = {Kluwer Academic Publishers},
title = {\href{http://link.springer.com/10.1007/BF00992696}{Simple statistical gradient-following algorithms for connectionist reinforcement learning}},
volume = {8},
year = {1992}
}

@inproceedings{correia2020EfficientMarginalizationDiscrete, title = {Efficient {{Marginalization}} of {{Discrete}} and {{Structured Latent Variables}} via {{Sparsity}}}, booktitle = {Proceedings of {{NeurIPS}}}, author = {Correia, Gonçalo M. and Niculae, Vlad and Aziz, Wilker and Martins, André F. T.}, year = {2020}, url = {http://arxiv.org/abs/2007.01919}}

@inproceedings{zantedeschi2022dag,
  title={DAG learning on the Permutahedron},
  author={Zantedeschi, Valentina and Kaddour, Jean and Franceschi, Luca and Kusner, Matt and Niculae, Vlad},
  booktitle={ICLR2022 Workshop on the Elements of Reasoning: Objects, Structure and Causality},
  year={2022}
}

@article{deng2018latent,
  title={Latent alignment and variational attention},
  author={Deng, Yuntian and Kim, Yoon and Chiu, Justin and Guo, Demi and Rush, Alexander},
  journal={Advances in Neural Information Processing Systems},
  volume={31},
  year={2018}
}

@inproceedings{fu2020gumbelcrfs,
 author = {Fu, Yao and Tan, Chuanqi and Bi, Bin and Chen, Mosha and Feng, Yansong and Rush, Alexander},
 booktitle = {Advances in Neural Information Processing Systems},
 editor = {H. Larochelle and M. Ranzato and R. Hadsell and M.F. Balcan and H. Lin},
 pages = {20259--20271},
 publisher = {Curran Associates, Inc.},
 title = {Latent Template Induction with Gumbel-CRFs},
 url = {https://proceedings.neurips.cc/paper/2020/file/ea119a40c1592979f51819b0bd38d39d-Paper.pdf},
 volume = {33},
 year = {2020}
}

@article{paulus2020gradient,
  title={Gradient estimation with stochastic softmax tricks},
  author={Paulus, Max and Choi, Dami and Tarlow, Daniel and Krause, Andreas and Maddison, Chris J},
  journal={Advances in Neural Information Processing Systems},
  volume={33},
  pages={5691--5704},
  year={2020}
}

@article{huijben2022review,
  title={A Review of the Gumbel-max Trick and its Extensions for Discrete Stochasticity in Machine Learning},
  author={Huijben, Iris AM and Kool, Wouter and Paulus, Max Benedikt and Van Sloun, Ruud JG},
  journal={IEEE Transactions on Pattern Analysis and Machine Intelligence},
  year={2022},
  publisher={IEEE}
}

@article{kool2020topk,
  author  = {Wouter Kool and Herke van Hoof and Max Welling},
  title   = {Ancestral Gumbel-Top-k Sampling for Sampling Without Replacement},
  journal = {Journal of Machine Learning Research},
  year    = {2020},
  volume  = {21},
  number  = {47},
  pages   = {1--36},
  url     = {http://jmlr.org/papers/v21/19-985.html}
}

@inproceedings{rennie2017self,
  title={Self-critical sequence training for image captioning},
  author={Rennie, Steven J and Marcheret, Etienne and Mroueh, Youssef and Ross, Jerret and Goel, Vaibhava},
  booktitle={Proceedings of the IEEE conference on computer vision and pattern recognition},
  pages={7008--7024},
  year={2017}
}

@inproceedings{Mnih2014,
author = {Mnih, Andriy and Gregor, Karol},
booktitle = {Proceedings of ICML},
title = {\href{http://arxiv.org/abs/1402.0030}{Neural Variational Inference and Learning in Belief Networks}},
year = {2014}
}

@inproceedings{Tucker2017,
author = {Tucker, George and Mnih, Andriy and Maddison, Chris J. and Lawson, Dieterich and Sohl-Dickstein, Jascha},
booktitle = {Proceedings of NeurIPS},
title = {\href{http://arxiv.org/abs/1703.07370}{REBAR: Low-variance, unbiased gradient estimates for discrete latent variable models}},
year = {2017}
}

@inproceedings{MuProp,
  title={\href{https://arxiv.org/pdf/1511.05176.pdf}{MuProp: unbiased backpropagation for stochastic neural networks}},
  author={Gu, Shixiang and Levine, Sergey and Sutskever, Ilya and Mnih, Andriy},
  booktitle={Proc. ICLR},
  year={2016}
}

@inproceedings{RELAX,
  title={\href{https://arxiv.org/pdf/1711.00123.pdf}{Backpropagation through the void: optimizing control variates for black-box gradient estimation}},
  author={Grathwohl, Will and Choi, Dami and Wu, Yuhuai and Roeder, Geoffrey and Duvenaud, David},
  booktitle={Proc. ICLR},
  year={2018}
}

@InProceedings{RB19,
  title = 	 {\href{http://proceedings.mlr.press/v97/liu19c/liu19c.pdf}{Rao-Blackwellized stochastic gradients for discrete distributions}},
  author = 	 {Liu, Runjing and Regier, Jeffrey and Tripuraneni, Nilesh and Jordan, Michael and Mcauliffe, Jon},
  booktitle = 	 {Proc. ICML},
  year = 	 {2019}
}

@inproceedings{
Kool2020Estimating,
title={\href{https://openreview.net/forum?id=rklEj2EFvB}{Estimating Gradients for Discrete Random Variables by Sampling without Replacement}},
author={Wouter Kool and Herke van Hoof and Max Welling},
booktitle={Proc. ICLR},
year={2020}
}

@inproceedings{farinhas2022sparse,
  title={Sparse Communication via Mixed Distributions},
  author={Farinhas, Ant{\'o}nio and Aziz, Wilker and Niculae, Vlad and Martins, Andr{\'e} FT},
  booktitle={International Conference on Learning Representations},
  year={2022}
}

@inproceedings{louizos2018learning,
title={Learning Sparse Neural Networks through $L_0$ Regularization},
author={Christos Louizos and Max Welling and Diederik P. Kingma},
booktitle={International Conference on Learning Representations},
year={2018},
url={https://openreview.net/forum?id=H1Y8hhg0b},
}

@article{palmer2017methods,
  title={Methods for Stochastic Collection and Replenishment (SCAR) optimisation for persistent autonomy},
  author={Palmer, Andrew W and Hill, Andrew J and Scheding, Steven J},
  journal={Robotics and Autonomous Systems},
  volume={87},
  pages={51--65},
  year={2017},
  publisher={Elsevier}
}

@inproceedings{bastings2019interpretable,
    title = "Interpretable Neural Predictions with Differentiable Binary Variables",
    author = "Bastings, Jasmijn  and
      Aziz, Wilker  and
      Titov, Ivan",
    booktitle = "Proceedings of the 57th Annual Meeting of the Association for Computational Linguistics",
    month = jul,
    year = "2019",
    address = "Florence, Italy",
    publisher = "Association for Computational Linguistics",
    url = "https://aclanthology.org/P19-1284",
    doi = "10.18653/v1/P19-1284",
    pages = "2963--2977",
}

@article{kumaraswamy1980generalized,
  title={A generalized probability density function for double-bounded random processes},
  author={Kumaraswamy, Ponnambalam},
  journal={Journal of hydrology},
  volume={46},
  number={1-2},
  pages={79--88},
  year={1980},
  publisher={Elsevier}
}

@article{mitchell1988bayesian,
  title={Bayesian variable selection in linear regression},
  author={Mitchell, Toby J and Beauchamp, John J},
  journal={Journal of the American Statistical Association},
  volume={83},
  number={404},
  pages={1023--1032},
  year={1988},
  publisher={Taylor \& Francis Group}
}

@article{ishwaran2005spike,
  title={Spike and slab variable selection: frequentist and Bayesian strategies},
  author={Ishwaran, Hemant and Rao, J Sunil and others},
  journal={Annals of Statistics},
  volume={33},
  number={2},
  pages={730--773},
  year={2005},
  publisher={Institute of Mathematical Statistics}
}

@inproceedings{rolfe2016discrete,
  author    = {Jason Tyler Rolfe},
  title     = {Discrete Variational Autoencoders},
  booktitle = {5th International Conference on Learning Representations, {ICLR} 2017,
               Toulon, France, April 24-26, 2017, Conference Track Proceedings},
  publisher = {OpenReview.net},
  year      = {2017},
  url       = {https://openreview.net/forum?id=ryMxXPFex},
  bibsource = {dblp computer science bibliography, https://dblp.org}
}

@InProceedings{vahdat2018dvae++,
  title = 	 {{DVAE}++: Discrete Variational Autoencoders with Overlapping Transformations},
  author =       {Vahdat, Arash and Macready, William and Bian, Zhengbing and Khoshaman, Amir and Andriyash, Evgeny},
  booktitle = 	 {Proceedings of the 35th International Conference on Machine Learning},
  pages = 	 {5035--5044},
  year = 	 {2018},
  editor = 	 {Dy, Jennifer and Krause, Andreas},
  volume = 	 {80},
  series = 	 {Proceedings of Machine Learning Research},
  month = 	 {10--15 Jul},
  publisher =    {PMLR},
  url = 	 {https://proceedings.mlr.press/v80/vahdat18a.html}
}

@article{niepert2021implicit,
  title={Implicit MLE: backpropagating through discrete exponential family distributions},
  author={Niepert, Mathias and Minervini, Pasquale and Franceschi, Luca},
  journal={Advances in Neural Information Processing Systems},
  volume={34},
  pages={14567--14579},
  year={2021}
}

@article{hinton1997generative,
  title={Generative models for discovering sparse distributed representations},
  author={Hinton, Geoffrey E and Ghahramani, Zoubin},
  journal={Philosophical Transactions of the Royal Society of London. Series B: Biological Sciences},
  volume={352},
  number={1358},
  pages={1177--1190},
  year={1997},
  publisher={The Royal Society}
}

@inproceedings{spigot,
    title = "Backpropagating through Structured Argmax using a {SPIGOT}",
    author = "Peng, Hao  and Thomson, Sam  and Smith, Noah A.",
    booktitle = "Proceedings of the 56th Annual Meeting of the Association for Computational Linguistics (Volume 1: Long Papers)",
    month = jul,
    year = "2018",
    address = "Melbourne, Australia",
    publisher = "Association for Computational Linguistics",
    url = "https://aclanthology.org/P18-1173",
    doi = "10.18653/v1/P18-1173",
    pages = "1863--1873"
}

@inproceedings{mihaylova-etal-2020-understanding,
    title = "Understanding the Mechanics of {SPIGOT}: Surrogate Gradients for Latent Structure Learning",
    author = "Mihaylova, Tsvetomila  and
      Niculae, Vlad  and
      Martins, Andr{\'e} F. T.",
    booktitle = "Proceedings of the 2020 Conference on Empirical Methods in Natural Language Processing (EMNLP)",
    year = "2020",
    address = "Online",
    publisher = "Association for Computational Linguistics",
    url = "https://aclanthology.org/2020.emnlp-main.171",
    doi = "10.18653/v1/2020.emnlp-main.171",
}

@incollection{quantize,
    title={A Survey of Quantization Methods for Efficient Neural Network
        Inference},
    author={Gholami, Amir and Kim, Sehoon and Dong, Zhen and Yao, Zhewei and Mahoney, Michael W and Keutzer, Kurt},
    booktitle={Low-Power Computer Vision},
    editor={Thiruvathukal, George K and Lu, Yung-Hsiang and Kim, Jaeyoun and
        Chen, Yiran and Chen, Bo},
    year={2022},
    publisher={Chapman and Hall/CRC}
}

@inproceedings{idf,
 author = {Hoogeboom, Emiel and Peters, Jorn and van den Berg, Rianne and Welling, Max},
 booktitle = {Advances in Neural Information Processing Systems},
 title = {\href{https://papers.neurips.cc/paper/2019/hash/9e9a30b74c49d07d8150c8c83b1ccf07-Abstract.html}{Integer discrete flows and lossless compression}},
 year = {2019}
}

@article{caruana1997multitask,
  title={Multitask learning},
  author={Caruana, Rich},
  journal={Machine learning},
  volume={28},
  number={1},
  pages={41--75},
  year={1997},
  publisher={Springer}
}

@article{goldstein1964convex,
  title={Convex programming in Hilbert space},
  author={Goldstein, Alan A},
  journal={Bulletin of the American Mathematical Society},
  volume={70},
  number={5},
  pages={709--710},
  year={1964}
}

@article{levitin1966constrained,
  title={Constrained minimization methods},
  author={Levitin, Evgeny S and Polyak, Boris T},
  journal={USSR Computational mathematics and mathematical physics},
  volume={6},
  number={5},
  pages={1--50},
  year={1966}
}

@inproceedings{diffbb,
title={Differentiation of Blackbox Combinatorial Solvers},
author={Marin Vlastelica Pogančić and Anselm Paulus and Vit Musil and Georg Martius and Michal Rolinek},
booktitle={International Conference on Learning Representations},
year={2020},
url={https://openreview.net/forum?id=BkevoJSYPB}
}

@InProceedings{xu-hardattn,
  title = 	 {Show, Attend and Tell: Neural Image Caption Generation with Visual Attention},
  author = 	 {Xu, Kelvin and Ba, Jimmy and Kiros, Ryan and Cho, Kyunghyun and Courville, Aaron and Salakhudinov, Ruslan and Zemel, Rich and Bengio, Yoshua},
  booktitle = 	 {Proceedings of the 32nd International Conference on Machine Learning},
  pages = 	 {2048--2057},
  year = 	 {2015},
  editor = 	 {Bach, Francis and Blei, David},
  volume = 	 {37},
  series = 	 {Proceedings of Machine Learning Research},
  address = 	 {Lille, France},
  month = 	 {07--09 Jul},
  publisher =    {PMLR},
  url = 	 {https://proceedings.mlr.press/v37/xuc15.html},
}

@book{mml,
  title={\href{https://mml-book.github.io/}{Mathematics for machine learning}},
  author={Deisenroth, Marc Peter and Faisal, A Aldo and Ong, Cheng Soon},
  year={2020},
  publisher={Cambridge University Press}
}

@article{lecuyer,
  title={Note: On the interchange of derivative and expectation for likelihood ratio derivative estimators},
  author={L'Ecuyer, Pierre},
  journal={Management Science},
  volume={41},
  number={4},
  pages={738--747},
  year={1995},
  publisher={INFORMS}
}

@inproceedings{Lazaridou2017,
author = {Lazaridou, Angeliki and Peysakhovich, Alexander and Baroni, Marco},
booktitle = {Proc. ICLR},
title = {\href{http://arxiv.org/abs/1612.07182}{Multi-agent cooperation and the emergence of (natural) language}},
year = {2017}
}

@inproceedings{Havrylov2017,
author = {Havrylov, Serhii and Titov, Ivan},
booktitle = {Proc. NeurIPS},
title = {\href{http://arxiv.org/abs/1705.11192}{Emergence of language with multi-agent games: Learning to communicate with sequences of symbols}},
year = {2017}
}

@inproceedings{kyrillidis2013sparse,
  title={\href{http://proceedings.mlr.press/v28/kyrillidis13.pdf}{Sparse projections onto the simplex}},
  author={Kyrillidis, Anastasios and Becker, Stephen and Cevher, Volkan and Koch, Christoph},
  booktitle={Proc. ICML},
  year={2013}
}

@article{berthet2020learning,
  title={Learning with differentiable pertubed optimizers},
  author={Berthet, Quentin and Blondel, Mathieu and Teboul, Olivier and Cuturi, Marco and Vert, Jean-Philippe and Bach, Francis},
  journal={Advances in neural information processing systems},
  volume={33},
  pages={9508--9519},
  year={2020}
}

@book{gumbel1954statistical,
  title={Statistical theory of extreme values and some practical applications: a series of lectures},
  author={Gumbel, Emil Julius},
  volume={33},
  year={1954},
  publisher={US Government Printing Office}
}

@article{yellott1977relationship,
  title={The relationship between Luce's choice axiom, Thurstone's theory of comparative judgment, and the double exponential distribution},
  author={Yellott, Jr., John I},
  journal={Journal of Mathematical Psychology},
  volume={15},
  number={2},
  pages={109--144},
  year={1977},
  publisher={Elsevier}
}

@inproceedings{vqvae,
 author = {van den Oord, Aaron and Vinyals, Oriol and kavukcuoglu, koray},
 booktitle = {Advances in Neural Information Processing Systems},
 editor = {I. Guyon and U. Von Luxburg and S. Bengio and H. Wallach and R. Fergus and S. Vishwanathan and R. Garnett},
 pages = {},
 publisher = {Curran Associates, Inc.},
 title = {Neural Discrete Representation Learning},
 url = {https://proceedings.neurips.cc/paper/2017/file/7a98af17e63a0ac09ce2e96d03992fbc-Paper.pdf},
 volume = {30},
 year = {2017}
}

@article{vintsyuk1968speech,
  title={Speech discrimination by dynamic programming},
  author={Vintsyuk, Taras K},
  journal={Cybernetics},
  volume={4},
  number={1},
  pages={52--57},
  year={1968},
  publisher={Springer}
}

@article{needleman,
title = {A general method applicable to the search for similarities in the amino acid sequence of two proteins},
journal = {Journal of Molecular Biology},
volume = {48},
number = {3},
pages = {443-453},
year = {1970},
issn = {0022-2836},
doi = {https://doi.org/10.1016/0022-2836(70)90057-4},
url = {https://www.sciencedirect.com/science/article/pii/0022283670900574},
author = {Saul B. Needleman and Christian D. Wunsch},
}

@article{wagner,
author = {Wagner, Robert A. and Fischer, Michael J.},
title = {The String-to-String Correction Problem},
year = {1974},
issue_date = {Jan. 1974},
publisher = {Association for Computing Machinery},
address = {New York, NY, USA},
volume = {21},
number = {1},
issn = {0004-5411},
url = {https://doi.org/10.1145/321796.321811},
doi = {10.1145/321796.321811},
journal = {J. ACM},
month = {jan},
pages = {168–173},
numpages = {6}
}

@inproceedings{bridle1989softmax,
	author = {Bridle, John},
	booktitle = {Advances in Neural Information Processing Systems},
	editor = {D. Touretzky},
	pages = {},
	publisher = {Morgan-Kaufmann},
	title = {Training Stochastic Model Recognition Algorithms as Networks can Lead to Maximum Mutual Information Estimation of Parameters},
	url = {https://proceedings.neurips.cc/paper_files/paper/1989/file/0336dcbab05b9d5ad24f4333c7658a0e-Paper.pdf},
	volume = {2},
	year = {1989}
}

@inproceedings{vapnik1991erm,
	author = {Vapnik, V.},
	booktitle = {Advances in Neural Information Processing Systems},
	editor = {J. Moody and S. Hanson and R.P. Lippmann},
	pages = {},
	publisher = {Morgan-Kaufmann},
	title = {Principles of Risk Minimization for Learning Theory},
	url = {https://proceedings.neurips.cc/paper_files/paper/1991/file/ff4d5fbbafdf976cfdc032e3bde78de5-Paper.pdf},
	volume = {4},
	year = {1991}
}

@article{bartlett2006convexity,
	author = {Peter L Bartlett, Michael I Jordan and Jon D McAuliffe},
	title = {Convexity, Classification, and Risk Bounds},
	journal = {Journal of the American Statistical Association},
	volume = {101},
	number = {473},
	pages = {138--156},
	year = {2006},
	publisher = {Taylor \& Francis},
	doi = {10.1198/016214505000000907},
	URL = { https://doi.org/10.1198/016214505000000907},
	eprint = { https://doi.org/10.1198/016214505000000907}
}

@article{reid2010binary,
	author  = {Mark D. Reid and Robert C. Williamson},
	title   = {Composite Binary Losses},
	journal = {Journal of Machine Learning Research},
	year    = {2010},
	volume  = {11},
	number  = {83},
	pages   = {2387--2422},
	url     = {http://jmlr.org/papers/v11/reid10a.html}
}

@phdthesis{linnainmaa1970representation,
	title={The representation of the cumulative rounding error of an algorithm as a Taylor expansion of the local rounding errors},
	author={Linnainmaa, Seppo},
	year={1970},
	school={Master’s Thesis (in Finnish), Univ. Helsinki}
}

@inproceedings{zanon2022segmentation,
	title = "Unsupervised Word Segmentation from Discrete Speech Units in Low-Resource Settings",
	author = "Zanon Boito, Marcely  and
	Yusuf, Bolaji  and
	Ondel, Lucas  and
	Villavicencio, Aline  and
	Besacier, Laurent",
	editor = "Melero, Maite  and
	Sakti, Sakriani  and
	Soria, Claudia",
	booktitle = "Proceedings of the 1st Annual Meeting of the ELRA/ISCA Special Interest Group on Under-Resourced Languages",
	month = jun,
	year = "2022",
	address = "Marseille, France",
	publisher = "European Language Resources Association",
	url = "https://aclanthology.org/2022.sigul-1.1",
	pages = "1--9",
}

@inproceedings{goldwater2007bayesiantagger,
	title = "A fully {B}ayesian approach to unsupervised part-of-speech tagging",
	author = "Goldwater, Sharon  and
	Griffiths, Tom",
	editor = "Zaenen, Annie  and
	van den Bosch, Antal",
	booktitle = "Proceedings of the 45th Annual Meeting of the Association of Computational Linguistics",
	month = jun,
	year = "2007",
	address = "Prague, Czech Republic",
	publisher = "Association for Computational Linguistics",
	url = "https://aclanthology.org/P07-1094",
	pages = "744--751",
}

@inproceedings{goldwater2006contextual,
	title = "Contextual Dependencies in Unsupervised Word Segmentation",
	author = "Goldwater, Sharon  and
	Griffiths, Thomas L.  and
	Johnson, Mark",
	editor = "Calzolari, Nicoletta  and
	Cardie, Claire  and
	Isabelle, Pierre",
	booktitle = "Proceedings of the 21st International Conference on Computational Linguistics and 44th Annual Meeting of the Association for Computational Linguistics",
	month = jul,
	year = "2006",
	address = "Sydney, Australia",
	publisher = "Association for Computational Linguistics",
	url = "https://aclanthology.org/P06-1085",
	doi = "10.3115/1220175.1220260",
	pages = "673--680",
}

@article{venkataraman2001segmentation,
	title = "A Statistical Model for Word Discovery in Transcribed Speech",
	author = "Venkataraman, Anand",
	editor = "Hirschberg, Julia",
	journal = "Computational Linguistics",
	volume = "27",
	number = "3",
	year = "2001",
	address = "Cambridge, MA",
	publisher = "MIT Press",
	url = "https://aclanthology.org/J01-3002",
	doi = "10.1162/089120101317066113",
	pages = "351--372",
}

@article{brent1999segmentation,
	title={An efficient, probabilistically sound algorithm for segmentation and word discovery},
	author={Brent, Michael R},
	journal={Machine Learning},
	volume={34},
	pages={71--105},
	year={1999},
	publisher={Springer}
}

@inproceedings{beal2001infinitehmm,
	author = {Beal, Matthew and Ghahramani, Zoubin and Rasmussen, Carl},
	booktitle = {Advances in Neural Information Processing Systems},
	editor = {T. Dietterich and S. Becker and Z. Ghahramani},
	pages = {},
	publisher = {MIT Press},
	title = {The Infinite Hidden Markov Model},
	url = {https://proceedings.neurips.cc/paper_files/paper/2001/file/e3408432c1a48a52fb6c74d926b38886-Paper.pdf},
	volume = {14},
	year = {2001}
}

@inproceedings{klein2004dmv,
	title = "Corpus-Based Induction of Syntactic Structure: Models of Dependency and Constituency",
	author = "Klein, Dan  and
	Manning, Christopher",
	booktitle = "Proceedings of the 42nd Annual Meeting of the Association for Computational Linguistics ({ACL}-04)",
	month = jul,
	year = "2004",
	address = "Barcelona, Spain",
	url = "https://aclanthology.org/P04-1061",
	doi = "10.3115/1218955.1219016",
	pages = "478--485",
}

@inproceedings{clark2001unsupervisedpcfg,
	title = "Unsupervised induction of stochastic context-free grammars using distributional clustering",
	author = "Clark, Alexander",
	booktitle = "Proceedings of the {ACL} 2001 Workshop on Computational Natural Language Learning ({C}on{LL})",
	year = "2001",
	url = "https://aclanthology.org/W01-0713",
}

@inproceedings{johnson2007bayesianpcfg,
	title = "{B}ayesian Inference for {PCFG}s via {M}arkov Chain {M}onte {C}arlo",
	author = "Johnson, Mark  and
	Griffiths, Thomas  and
	Goldwater, Sharon",
	editor = "Sidner, Candace  and
	Schultz, Tanja  and
	Stone, Matthew  and
	Zhai, ChengXiang",
	booktitle = "Human Language Technologies 2007: The Conference of the North {A}merican Chapter of the Association for Computational Linguistics; Proceedings of the Main Conference",
	month = apr,
	year = "2007",
	address = "Rochester, New York",
	publisher = "Association for Computational Linguistics",
	url = "https://aclanthology.org/N07-1018",
	pages = "139--146",
}

@article{lari1990insideoutside,
	title = {The estimation of stochastic context-free grammars using the Inside-Outside algorithm},
	journal = {Computer Speech \& Language},
	volume = {4},
	number = {1},
	pages = {35-56},
	year = {1990},
	issn = {0885-2308},
	doi = {https://doi.org/10.1016/0885-2308(90)90022-X},
	url = {https://www.sciencedirect.com/science/article/pii/088523089090022X},
	author = {K. Lari and S.J. Young}
}

@article{chen2009latentperm,
	title={Content modeling using latent permutations},
	author={Chen, Harr and Branavan, SRK and Barzilay, Regina and Karger, David R},
	journal={Journal of Artificial Intelligence Research},
	volume={36},
	pages={129--163},
	year={2009}
}

@inproceedings{eisenstein2008topicsegmentation,
	title = "{B}ayesian Unsupervised Topic Segmentation",
	author = "Eisenstein, Jacob  and
	Barzilay, Regina",
	editor = "Lapata, Mirella  and
	Ng, Hwee Tou",
	booktitle = "Proceedings of the 2008 Conference on Empirical Methods in Natural Language Processing",
	month = oct,
	year = "2008",
	address = "Honolulu, Hawaii",
	publisher = "Association for Computational Linguistics",
	url = "https://aclanthology.org/D08-1035",
	pages = "334--343",
}

@inproceedings{du2013topicsegmentation,
	title = "Topic Segmentation with a Structured Topic Model",
	author = "Du, Lan  and
	Buntine, Wray  and
	Johnson, Mark",
	editor = "Vanderwende, Lucy  and
	Daum{\'e} III, Hal  and
	Kirchhoff, Katrin",
	booktitle = "Proceedings of the 2013 Conference of the North {A}merican Chapter of the Association for Computational Linguistics: Human Language Technologies",
	month = jun,
	year = "2013",
	address = "Atlanta, Georgia",
	publisher = "Association for Computational Linguistics",
	url = "https://aclanthology.org/N13-1019",
	pages = "190--200",
}

@article{saul1996mft,
	title={Mean field theory for sigmoid belief networks},
	author={Saul, Lawrence K and Jaakkola, Tommi and Jordan, Michael I},
	journal={Journal of artificial intelligence research},
	volume={4},
	pages={61--76},
	year={1996}
}

@article{hinton2002poe,
	author = {Hinton, Geoffrey E.},
	title = "{Training Products of Experts by Minimizing Contrastive Divergence}",
	journal = {Neural Computation},
	volume = {14},
	number = {8},
	pages = {1771-1800},
	year = {2002},
	month = {08},
	issn = {0899-7667},
	doi = {10.1162/089976602760128018},
	url = {https://doi.org/10.1162/089976602760128018},
	eprint = {https://direct.mit.edu/neco/article-pdf/14/8/1771/815447/089976602760128018.pdf},
}

@article{hinton2006dbn,
	author = {Hinton, Geoffrey E. and Osindero, Simon and Teh, Yee-Whye},
	title = "{A Fast Learning Algorithm for Deep Belief Nets}",
	journal = {Neural Computation},
	volume = {18},
	number = {7},
	pages = {1527-1554},
	year = {2006},
	month = {07},
	issn = {0899-7667},
	doi = {10.1162/neco.2006.18.7.1527},
	url = {https://doi.org/10.1162/neco.2006.18.7.1527},
	eprint = {https://direct.mit.edu/neco/article-pdf/18/7/1527/816558/neco.2006.18.7.1527.pdf},
}

@article{peterson1987mft,
	title={A mean field theory learning algorithm for neural network},
	author={Peterson, Carsten},
	journal={Complex systems},
	volume={1},
	pages={995--1019},
	year={1987}
}

@article{neal1992gibbs,
	title = {Connectionist learning of belief networks},
	journal = {Artificial Intelligence},
	volume = {56},
	number = {1},
	pages = {71-113},
	year = {1992},
	issn = {0004-3702},
	doi = {https://doi.org/10.1016/0004-3702(92)90065-6},
	url = {https://www.sciencedirect.com/science/article/pii/0004370292900656},
	author = {Radford M. Neal}
}

@article{ackley1985boltzmann,
	title={A learning algorithm for Boltzmann machines},
	author={Ackley, David H and Hinton, Geoffrey E and Sejnowski, Terrence J},
	journal={Cognitive science},
	volume={9},
	number={1},
	pages={147--169},
	year={1985},
	publisher={Elsevier}
}

@InProceedings{salakhutdinov2009deepbm,
	title = 	 {Deep Boltzmann Machines},
	author = 	 {Salakhutdinov, Ruslan and Hinton, Geoffrey},
	booktitle = 	 {Proceedings of the Twelth International Conference on Artificial Intelligence and Statistics},
	pages = 	 {448--455},
	year = 	 {2009},
	editor = 	 {van Dyk, David and Welling, Max},
	volume = 	 {5},
	series = 	 {Proceedings of Machine Learning Research},
	address = 	 {Hilton Clearwater Beach Resort, Clearwater Beach, Florida USA},
	month = 	 {16--18 Apr},
	publisher =    {PMLR},
	url = 	 {https://proceedings.mlr.press/v5/salakhutdinov09a.html}
}

@article{parisi1988statistical,
	title={Statistical field theory},
	author={Parisi, Giorgio and Shankar, Ramamurti},
	year={1988},
	publisher={Westview Press}
}

@article{dempster1977em,
	author = {Dempster, A. P. and Laird, N. M. and Rubin, D. B.},
	title = {Maximum Likelihood from Incomplete Data Via the EM Algorithm},
	journal = {Journal of the Royal Statistical Society: Series B (Methodological)},
	volume = {39},
	number = {1},
	pages = {1-22},
	doi = {https://doi.org/10.1111/j.2517-6161.1977.tb01600.x},
	url = {https://rss.onlinelibrary.wiley.com/doi/abs/10.1111/j.2517-6161.1977.tb01600.x},
	eprint = {https://rss.onlinelibrary.wiley.com/doi/pdf/10.1111/j.2517-6161.1977.tb01600.x},
	year = {1977}
}

@book{evaluatingderivatives,
author = {Griewank, Andreas and Walther, Andrea},
title = {Evaluating Derivatives},
publisher = {Society for Industrial and Applied Mathematics},
year = {2008},
doi = {10.1137/1.9780898717761},
address = {},
edition   = {Second},
URL = {https://epubs.siam.org/doi/abs/10.1137/1.9780898717761},
eprint = {https://epubs.siam.org/doi/pdf/10.1137/1.9780898717761}
}

@Inbook{Bishop1998,
author="Bishop, Christopher M.",
editor="Jordan, Michael I.",
title="Latent Variable Models",
bookTitle="Learning in Graphical Models",
year="1998",
publisher="Springer Netherlands",
address="Dordrecht",
pages="371--403",
isbn="978-94-011-5014-9",
doi="10.1007/978-94-011-5014-9_13",
url="https://doi.org/10.1007/978-94-011-5014-9_13"
}

@ARTICLE{replearning,
  author={Bengio, Yoshua and Courville, Aaron and Vincent, Pascal},
  journal={IEEE Transactions on Pattern Analysis and Machine Intelligence}, 
  title={Representation Learning: A Review and New Perspectives}, 
  year={2013},
  volume={35},
  number={8},
  pages={1798-1828},
doi={10.1109/TPAMI.2013.50}}

@article{hitchcock1941distribution,
  title={The distribution of a product from several sources to numerous localities},
  author={Hitchcock, Frank L},
  journal={Journal of mathematics and physics},
  volume={20},
  number={1--4},
  pages={224--230},
  year={1941},
  publisher={Wiley Online Library}
}

@article{kanto,
    title={On the translocation of masses},
    author={Kantorovich, LV},
    journal={Dokl Asad Nauk SSSR},
    volume={37},
    number={7--8},
    pages={227--229},
    year={1942}
}

@ARTICLE{mlclf,
title = {Multi-Label Classification: An Overview},
author = {Tsoumakas, Grigorios and Katakis, Ioannis},
year = {2007},
journal = {International Journal of Data Warehousing and Mining (IJDWM)},
volume = {3},
number = {3},
pages = {1-13},
url = {https://EconPapers.repec.org/RePEc:igg:jdwm00:v:3:y:2007:i:3:p:1-13}
}

@book{pml1book,
 author = "Kevin P. Murphy",
 title = "Probabilistic Machine Learning: An introduction",
 publisher = "MIT Press",
 year = 2022,
 url = "probml.ai"
}

@book{tutte2001graph,
  title={Graph theory},
  author={Tutte, William Thomas},
  volume={21},
  year={2001},
  publisher={Cambridge university press}
}

@inproceedings{martins-etal-2009-concise,
    title = "Concise Integer Linear Programming Formulations for Dependency Parsing",
    author = "Martins, Andr{\'e}  and
      Smith, Noah  and
      Xing, Eric",
    editor = "Su, Keh-Yih  and
      Su, Jian  and
      Wiebe, Janyce  and
      Li, Haizhou",
    booktitle = "Proceedings of the Joint Conference of the 47th Annual Meeting of the {ACL} and the 4th International Joint Conference on Natural Language Processing of the {AFNLP}",
    month = aug,
    year = "2009",
    address = "Suntec, Singapore",
    publisher = "Association for Computational Linguistics",
    url = "https://aclanthology.org/P09-1039",
    pages = "342--350",
}

@article{montecarlo,
author = {Nicholas Metropolis and S. Ulam},
title = {The Monte Carlo Method},
journal = {Journal of the American Statistical Association},
volume = {44},
number = {247},
pages = {335--341},
year = {1949},
publisher = {ASA Website},
doi = {10.1080/01621459.1949.10483310},

    note ={PMID: 18139350},


URL = { 
    
    
        https://www.tandfonline.com/doi/abs/10.1080/01621459.1949.10483310
    

},
eprint = { 
    
    
        https://www.tandfonline.com/doi/pdf/10.1080/01621459.1949.10483310
    

}

}

@book{pflug2012optimization,
  title={Optimization of stochastic models: the interface between simulation and optimization},
  author={Pflug, Georg Ch},
  volume={373},
  year={2012},
  publisher={Springer Science \& Business Media}
}

@book{glasserman1990gradient,
  title={Gradient estimation via perturbation analysis},
  author={Glasserman, Paul},
  volume={116},
  year={1990},
  publisher={Springer Science \& Business Media}
}

@article{ho1983optimization,
  title={Optimization and perturbation analysis of queueing networks},
  author={Ho, YC and Cao, XR},
  journal={Journal of Optimization Theory and Applications},
  volume={40},
  number={4},
  pages={559--582},
  year={1983}
}

@article{rubinstein1992sensitivity,
  title={Sensitivity analysis of discrete event systems by the “push out” method},
  author={Rubinstein, Reuven Y},
  journal={Annals of Operations Research},
  volume={39},
  number={1},
  pages={229--250},
  year={1992},
  publisher={Springer}
}

@misc{hinton2012coursera,
	author = {Hinton, Geoffrey E},
	title = {Neural networks for machine learning},
	year = {2012},
	howpublished = {Coursera, video lectures}
}

@article{casella1996raoblackwellisation,
	ISSN = {00063444, 14643510},
	URL = {http://www.jstor.org/stable/2337434},
	author = {George Casella and Christian P. Robert},
	journal = {Biometrika},
	number = {1},
	pages = {81--94},
	publisher = {[Oxford University Press, Biometrika Trust]},
	title = {Rao-Blackwellisation of Sampling Schemes},
	urldate = {2024-11-06},
	volume = {83},
	year = {1996}
}

@InProceedings{titsias2014varbayes,
	title = 	 {Doubly Stochastic Variational Bayes for non-Conjugate Inference},
	author = 	 {Titsias, Michalis and Lázaro-Gredilla, Miguel},
	booktitle = 	 {Proceedings of the 31st International Conference on Machine Learning},
	pages = 	 {1971--1979},
	year = 	 {2014},
	editor = 	 {Xing, Eric P. and Jebara, Tony},
	volume = 	 {32},
	number =       {2},
	series = 	 {Proceedings of Machine Learning Research},
	address = 	 {Bejing, China},
	month = 	 {22--24 Jun},
	publisher =    {PMLR},
	url = 	 {https://proceedings.mlr.press/v32/titsias14.html}
}

@inproceedings{titsias2015blackboxvi,
	author = {Titsias, Michalis K and L\'{a}zaro-Gredilla, Miguel},
	booktitle = {Advances in Neural Information Processing Systems},
	editor = {C. Cortes and N. Lawrence and D. Lee and M. Sugiyama and R. Garnett},
	pages = {},
	publisher = {Curran Associates, Inc.},
	title = {Local Expectation Gradients for Black Box Variational Inference},
	url = {https://proceedings.neurips.cc/paper_files/paper/2015/file/1373b284bc381890049e92d324f56de0-Paper.pdf},
	volume = {28},
	year = {2015}
}

@InProceedings{ranganath2014blackboxvi,
	title = 	 {{Black Box Variational Inference}},
	author = 	 {Ranganath, Rajesh and Gerrish, Sean and Blei, David},
	booktitle = 	 {Proceedings of the Seventeenth International Conference on Artificial Intelligence and Statistics},
	pages = 	 {814--822},
	year = 	 {2014},
	editor = 	 {Kaski, Samuel and Corander, Jukka},
	volume = 	 {33},
	series = 	 {Proceedings of Machine Learning Research},
	address = 	 {Reykjavik, Iceland},
	month = 	 {22--25 Apr},
	publisher =    {PMLR},
	url = 	 {https://proceedings.mlr.press/v33/ranganath14.html}
}

@inproceedings{johnson2016svae,
	author = {Johnson, Matthew J and Duvenaud, David K and Wiltschko, Alex and Adams, Ryan P and Datta, Sandeep R},
	booktitle = {Advances in Neural Information Processing Systems},
	editor = {D. Lee and M. Sugiyama and U. Luxburg and I. Guyon and R. Garnett},
	pages = {},
	publisher = {Curran Associates, Inc.},
	title = {Composing graphical models with neural networks for structured representations and fast inference},
	url = {https://proceedings.neurips.cc/paper_files/paper/2016/file/7d6044e95a16761171b130dcb476a43e-Paper.pdf},
	volume = {29},
	year = {2016}
}

@inproceedings{
	lin2018variational,
	title={Variational Message Passing with Structured Inference Networks},
	author={Wu Lin and Mohammad Emtiyaz Khan and Nicolas Hubacher},
	booktitle={International Conference on Learning Representations},
	year={2018},
	url={https://openreview.net/forum?id=HyH9lbZAW},
}

@InProceedings{zhao2023rsvae,
	title = 	 {Revisiting Structured Variational Autoencoders},
	author =       {Zhao, Yixiu and Linderman, Scott},
	booktitle = 	 {Proceedings of the 40th International Conference on Machine Learning},
	pages = 	 {42046--42057},
	year = 	 {2023},
	editor = 	 {Krause, Andreas and Brunskill, Emma and Cho, Kyunghyun and Engelhardt, Barbara and Sabato, Sivan and Scarlett, Jonathan},
	volume = 	 {202},
	series = 	 {Proceedings of Machine Learning Research},
	month = 	 {23--29 Jul},
	publisher =    {PMLR},
	url = 	 {https://proceedings.mlr.press/v202/zhao23c.html}
}

@InProceedings{pearce2020gpvae,
	title = 	 { The Gaussian Process Prior VAE for
	Interpretable Latent Dynamics from Pixels},
	author =       {Pearce, Michael},
	booktitle = 	 {Proceedings of The 2nd Symposium on
	Advances in Approximate Bayesian Inference},
	pages = 	 {1--12},
	year = 	 {2020},
	editor = 	 {Zhang, Cheng and Ruiz, Francisco and Bui, Thang and Dieng, Adji Bousso and Liang, Dawen},
	volume = 	 {118},
	series = 	 {Proceedings of Machine Learning Research},
	month = 	 {08 Dec},
	publisher =    {PMLR},
	url = 	 {https://proceedings.mlr.press/v118/pearce20a.html}
}

@inproceedings{yao2022structuralgen,
	title = "Structural generalization is hard for sequence-to-sequence models",
	author = "Yao, Yuekun  and
	Koller, Alexander",
	editor = "Goldberg, Yoav  and
	Kozareva, Zornitsa  and
	Zhang, Yue",
	booktitle = "Proceedings of the 2022 Conference on Empirical Methods in Natural Language Processing",
	month = dec,
	year = "2022",
	address = "Abu Dhabi, United Arab Emirates",
	publisher = "Association for Computational Linguistics",
	url = "https://aclanthology.org/2022.emnlp-main.337",
	doi = "10.18653/v1/2022.emnlp-main.337",
	pages = "5048--5062",
}

@article{bogin2021latemtcomp,
	title = "Latent Compositional Representations Improve Systematic Generalization in Grounded Question Answering",
	author = "Bogin, Ben  and
	Subramanian, Sanjay  and
	Gardner, Matt  and
	Berant, Jonathan",
	editor = "Roark, Brian  and
	Nenkova, Ani",
	journal = "Transactions of the Association for Computational Linguistics",
	volume = "9",
	year = "2021",
	address = "Cambridge, MA",
	publisher = "MIT Press",
	url = "https://aclanthology.org/2021.tacl-1.12",
	doi = "10.1162/tacl_a_00361",
	pages = "195--210",
}

@inproceedings{liu2021compgen,
	title = "Learning Algebraic Recombination for Compositional Generalization",
	author = "Liu, Chenyao  and
	An, Shengnan  and
	Lin, Zeqi  and
	Liu, Qian  and
	Chen, Bei  and
	Lou, Jian-Guang  and
	Wen, Lijie  and
	Zheng, Nanning  and
	Zhang, Dongmei",
	editor = "Zong, Chengqing  and
	Xia, Fei  and
	Li, Wenjie  and
	Navigli, Roberto",
	booktitle = "Findings of the Association for Computational Linguistics: ACL-IJCNLP 2021",
	month = aug,
	year = "2021",
	address = "Online",
	publisher = "Association for Computational Linguistics",
	url = "https://aclanthology.org/2021.findings-acl.97",
	doi = "10.18653/v1/2021.findings-acl.97",
	pages = "1129--1144",
}

@inproceedings{
	zhou2024length,
	title={What Algorithms can Transformers Learn? A Study in Length Generalization},
	author={Hattie Zhou and Arwen Bradley and Etai Littwin and Noam Razin and Omid Saremi and Joshua M. Susskind and Samy Bengio and Preetum Nakkiran},
	booktitle={The Twelfth International Conference on Learning Representations},
	year={2024},
	url={https://openreview.net/forum?id=AssIuHnmHX}
}

@inproceedings{anil2022length,
	author = {Anil, Cem and Wu, Yuhuai and Andreassen, Anders and Lewkowycz, Aitor and Misra, Vedant and Ramasesh, Vinay and Slone, Ambrose and Gur-Ari, Guy and Dyer, Ethan and Neyshabur, Behnam},
	booktitle = {Advances in Neural Information Processing Systems},
	editor = {S. Koyejo and S. Mohamed and A. Agarwal and D. Belgrave and K. Cho and A. Oh},
	pages = {38546--38556},
	publisher = {Curran Associates, Inc.},
	title = {Exploring Length Generalization in Large Language Models},
	url = {https://proceedings.neurips.cc/paper_files/paper/2022/file/fb7451e43f9c1c35b774bcfad7a5714b-Paper-Conference.pdf},
	volume = {35},
	year = {2022}
}

@article{narayan2023conditionalgen,
	title = "Conditional Generation with a Question-Answering Blueprint",
	author = "Narayan, Shashi  and
	Maynez, Joshua  and
	Amplayo, Reinald Kim  and
	Ganchev, Kuzman  and
	Louis, Annie  and
	Huot, Fantine  and
	Sandholm, Anders  and
	Das, Dipanjan  and
	Lapata, Mirella",
	journal = "Transactions of the Association for Computational Linguistics",
	volume = "11",
	year = "2023",
	address = "Cambridge, MA",
	publisher = "MIT Press",
	url = "https://aclanthology.org/2023.tacl-1.55",
	doi = "10.1162/tacl_a_00583",
	pages = "974--996",
}

@article{xu2022latentqueries,
	title = "Document Summarization with Latent Queries",
	author = "Xu, Yumo  and
	Lapata, Mirella",
	editor = "Roark, Brian  and
	Nenkova, Ani",
	journal = "Transactions of the Association for Computational Linguistics",
	volume = "10",
	year = "2022",
	address = "Cambridge, MA",
	publisher = "MIT Press",
	url = "https://aclanthology.org/2022.tacl-1.36",
	doi = "10.1162/tacl_a_00480",
	pages = "623--638",
}

@inproceedings{liu2023visualstorytelling,
	title = "Visual Storytelling with Question-Answer Plans",
	author = "Liu, Danyang  and
	Lapata, Mirella  and
	Keller, Frank",
	editor = "Bouamor, Houda  and
	Pino, Juan  and
	Bali, Kalika",
	booktitle = "Findings of the Association for Computational Linguistics: EMNLP 2023",
	month = dec,
	year = "2023",
	address = "Singapore",
	publisher = "Association for Computational Linguistics",
	url = "https://aclanthology.org/2023.findings-emnlp.386",
	doi = "10.18653/v1/2023.findings-emnlp.386",
	pages = "5800--5813",
}

@inproceedings{
	van2018relational,
	title={Relational Neural Expectation Maximization: Unsupervised Discovery of Objects and their Interactions},
	author={Sjoerd van Steenkiste and Michael Chang and Klaus Greff and Jürgen Schmidhuber},
	booktitle={International Conference on Learning Representations},
	year={2018},
	url={https://openreview.net/forum?id=ryH20GbRW},
}

@inproceedings{locatello2020slotatt,
	author = {Locatello, Francesco and Weissenborn, Dirk and Unterthiner, Thomas and Mahendran, Aravindh and Heigold, Georg and Uszkoreit, Jakob and Dosovitskiy, Alexey and Kipf, Thomas},
	booktitle = {Advances in Neural Information Processing Systems},
	editor = {H. Larochelle and M. Ranzato and R. Hadsell and M.F. Balcan and H. Lin},
	pages = {11525--11538},
	publisher = {Curran Associates, Inc.},
	title = {Object-Centric Learning with Slot Attention},
	url = {https://proceedings.neurips.cc/paper_files/paper/2020/file/8511df98c02ab60aea1b2356c013bc0f-Paper.pdf},
	volume = {33},
	year = {2020}
}

@InProceedings{greff2019multiobject,
	title = 	 {Multi-Object Representation Learning with Iterative Variational Inference},
	author =       {Greff, Klaus and Kaufman, Rapha{\"e}l Lopez and Kabra, Rishabh and Watters, Nick and Burgess, Christopher and Zoran, Daniel and Matthey, Loic and Botvinick, Matthew and Lerchner, Alexander},
	booktitle = 	 {Proceedings of the 36th International Conference on Machine Learning},
	pages = 	 {2424--2433},
	year = 	 {2019},
	editor = 	 {Chaudhuri, Kamalika and Salakhutdinov, Ruslan},
	volume = 	 {97},
	series = 	 {Proceedings of Machine Learning Research},
	month = 	 {09--15 Jun},
	publisher =    {PMLR},
	url = 	 {https://proceedings.mlr.press/v97/greff19a.html}
}

@inproceedings{elsayed2022savi,
	author = {Elsayed, Gamaleldin and Mahendran, Aravindh and van Steenkiste, Sjoerd and Greff, Klaus and Mozer, Michael C and Kipf, Thomas},
	booktitle = {Advances in Neural Information Processing Systems},
	editor = {S. Koyejo and S. Mohamed and A. Agarwal and D. Belgrave and K. Cho and A. Oh},
	pages = {28940--28954},
	publisher = {Curran Associates, Inc.},
	title = {SAVi++: Towards End-to-End Object-Centric Learning from Real-World Videos},
	url = {https://proceedings.neurips.cc/paper_files/paper/2022/file/ba1a6ba05319e410f0673f8477a871e3-Paper-Conference.pdf},
	volume = {35},
	year = {2022}
}

@article{glynn1990likelihood,
  title={Likelihood ratio gradient estimation for stochastic systems},
  author={Glynn, Peter W},
  journal={Communications of the ACM},
  volume={33},
  number={10},
  pages={75--84},
  year={1990},
  publisher={ACM New York, NY, USA}
}

@thesis{harpyspeech,
	author = {Lowerre, Bruce T.},
	title = {The {HARPY} speech recognition system},
	type = {PhD},
	institution = {Carnegie-Mellon University},
	date = {1976}
}

@article{meister-etal-2020-best,
title = "Best-First Beam Search",
author = "Meister, Clara  and
Vieira, Tim  and
Cotterell, Ryan",
editor = "Johnson, Mark  and
Roark, Brian  and
Nenkova, Ani",
journal = "Transactions of the Association for Computational Linguistics",
volume = "8",
year = "2020",
address = "Cambridge, MA",
publisher = "MIT Press",
url = "https://aclanthology.org/2020.tacl-1.51",
doi = "10.1162/tacl_a_00346",
pages = "795--809",
}

@InProceedings{pmlr-v97-kool19a,
title = 	 {Stochastic Beams and Where To Find Them: The {G}umbel-Top-k Trick for Sampling Sequences Without Replacement},
author =       {Kool, Wouter and Van Hoof, Herke and Welling, Max},
booktitle = 	 {Proceedings of the 36th International Conference on Machine Learning},
pages = 	 {3499--3508},
year = 	 {2019},
editor = 	 {Chaudhuri, Kamalika and Salakhutdinov, Ruslan},
volume = 	 {97},
series = 	 {Proceedings of Machine Learning Research},
month = 	 {09--15 Jun},
publisher =    {PMLR},
url = 	 {https://proceedings.mlr.press/v97/kool19a.html}
}

@inproceedings{meister-etal-2021-conditional,
title = "Conditional {P}oisson Stochastic Beams",
author = "Meister, Clara  and
Amini, Afra  and
Vieira, Tim  and
Cotterell, Ryan",
editor = "Moens, Marie-Francine  and
Huang, Xuanjing  and
Specia, Lucia  and
Yih, Scott Wen-tau",
booktitle = "Proceedings of the 2021 Conference on Empirical Methods in Natural Language Processing",
month = nov,
year = "2021",
address = "Online and Punta Cana, Dominican Republic",
publisher = "Association for Computational Linguistics",
url = "https://aclanthology.org/2021.emnlp-main.52",
doi = "10.18653/v1/2021.emnlp-main.52",
pages = "664--681",
}

@inproceedings{meister-etal-2020-beam,
title = "If beam search is the answer, what was the question?",
author = "Meister, Clara  and
Cotterell, Ryan  and
Vieira, Tim",
editor = "Webber, Bonnie  and
Cohn, Trevor  and
He, Yulan  and
Liu, Yang",
booktitle = "Proceedings of the 2020 Conference on Empirical Methods in Natural Language Processing (EMNLP)",
month = nov,
year = "2020",
address = "Online",
publisher = "Association for Computational Linguistics",
url = "https://aclanthology.org/2020.emnlp-main.170",
doi = "10.18653/v1/2020.emnlp-main.170",
pages = "2173--2185",
}

@inproceedings{kasai-etal-2024-call,
title = "A Call for Clarity in Beam Search: How It Works and When It Stops",
author = "Kasai, Jungo  and
Sakaguchi, Keisuke  and
Le Bras, Ronan  and
Radev, Dragomir  and
Choi, Yejin  and
Smith, Noah A.",
editor = "Calzolari, Nicoletta  and
Kan, Min-Yen  and
Hoste, Veronique  and
Lenci, Alessandro  and
Sakti, Sakriani  and
Xue, Nianwen",
booktitle = "Proceedings of the 2024 Joint International Conference on Computational Linguistics, Language Resources and Evaluation (LREC-COLING 2024)",
month = may,
year = "2024",
address = "Torino, Italia",
publisher = "ELRA and ICCL",
url = "https://aclanthology.org/2024.lrec-main.7",
pages = "77--90",
}

@inproceedings{eisner2016tutorial,
	title = "Inside-Outside and Forward-Backward Algorithms Are Just Backprop (tutorial paper)",
	author = "Eisner, Jason",
	editor = "Chang, Kai-Wei  and
	Chang, Ming-Wei  and
	Rush, Alexander  and
	Srikumar, Vivek",
	booktitle = "Proceedings of the Workshop on Structured Prediction for {NLP}",
	month = nov,
	year = "2016",
	address = "Austin, TX",
	publisher = "Association for Computational Linguistics",
	url = "https://aclanthology.org/W16-5901",
	doi = "10.18653/v1/W16-5901",
	pages = "1--17",
}

@ARTICLE{forney1973viterbi,
	author={Forney, G.D.},
	journal={Proceedings of the IEEE}, 
	title={The {V}iterbi algorithm}, 
	year={1973},
	volume={61},
	number={3},
	pages={268-278},
	doi={10.1109/PROC.1973.9030}}

@inproceedings{huang-2008-advanced,
    title = "Advanced Dynamic Programming in Semiring and Hypergraph Frameworks",
    author = "Huang, Liang",
    editor = "Huang, Liang",
    booktitle = "Coling 2008: Advanced Dynamic Programming in Computational Linguistics: Theory, Algorithms and Applications - Tutorial notes",
    month = aug,
    year = "2008",
    address = "Manchester, UK",
    publisher = "Coling 2008 Organizing Committee",
    url = "https://aclanthology.org/C08-5001",
    pages = "1--18",
}

@article{bellman,
  title={The theory of dynamic programming},
  author={Bellman, Richard},
  journal={Bulletin of the American Mathematical Society},
  volume={60},
  number={6},
  pages={503--515},
  year={1954}
}

@article{viterbi,
  author={Viterbi, Andrew},
  journal={IEEE Transactions on Information Theory}, 
  title={Error bounds for convolutional codes and an asymptotically optimum decoding algorithm}, 
  year={1967},
  volume={13},
  number={2},
  pages={260-269},
  doi={10.1109/TIT.1967.1054010}}

@article{mohri,
author = {Mohri, Mehryar},
title = {Semiring frameworks and algorithms for shortest-distance problems},
year = {2002},
issue_date = {January 2002},
publisher = {Otto-von-Guericke-Universitat},
address = {DEU},
volume = {7},
number = {3},
issn = {1430-189X},
journal = {J. Autom. Lang. Comb.},
pages = {321–350},
numpages = {30}
}

@article{ffbs,
author = {Frühwirth-Schnatter, Sylvia},
title = {Data augmentation and dynamic linear models},
journal = {Journal of Time Series Analysis},
volume = {15},
number = {2},
pages = {183-202},
doi = {https://doi.org/10.1111/j.1467-9892.1994.tb00184.x},
url = {https://onlinelibrary.wiley.com/doi/abs/10.1111/j.1467-9892.1994.tb00184.x},
eprint = {https://onlinelibrary.wiley.com/doi/pdf/10.1111/j.1467-9892.1994.tb00184.x},
year = {1994}
}

@inproceedings{baum,
  title={An inequality and associated maximization technique in statistical estimation of probabilistic functions of a Markov process},
  author={Leonard E. Baum},
  year={1972},
  url={https://api.semanticscholar.org/CorpusID:60804212}
}

@inproceedings{lafferty2001conditional,
  title={Conditional random fields: Probabilistic models for segmenting and labeling sequence data},
  author={Lafferty, John and McCallum, Andrew and Pereira, Fernando and others},
  booktitle={Icml},
  volume={1},
  number={2},
  pages={3},
  year={2001},
  organization={Williamstown, MA}
}

@inproceedings{baziotis-etal-2019-seq,
    title = "SEQ$^3$: Differentiable Sequence-to-Sequence-to-Sequence Autoencoder for Unsupervised Abstractive Sentence Compression",
    author = "Baziotis, Christos  and
      Androutsopoulos, Ion  and
      Konstas, Ioannis  and
      Potamianos, Alexandros",
    editor = "Burstein, Jill  and
      Doran, Christy  and
      Solorio, Thamar",
    booktitle = "Proceedings of the 2019 Conference of the North {A}merican Chapter of the Association for Computational Linguistics: Human Language Technologies, Volume 1 (Long and Short Papers)",
    month = jun,
    year = "2019",
    address = "Minneapolis, Minnesota",
    publisher = "Association for Computational Linguistics",
    url = "https://aclanthology.org/N19-1071",
    doi = "10.18653/v1/N19-1071",
    pages = "673--681",
}

@article{van2021storchastic,
  title={Storchastic: A Framework for General Stochastic Automatic Differentiation},
  author={van Krieken, Emile and Tomczak, Jakub M and Teije, Annette ten},
  booktitle = {Advances in Neural Information Processing Systems},
  editor = {M. Ranzato and A. Beygelzimer and Y. Dauphin and P.S. Liang and J. Wortman Vaughan},
  pages = {7574--7587},
  url = {https://proceedings.neurips.cc/paper_files/paper/2021/file/3dfe2f633108d604df160cd1b01710db-Paper.pdf},
  volume = {34},
  year={2021}
}

@misc{alex2020torchstruct,
    title={Torch-Struct: Deep Structured Prediction Library},
    author={Alexander M. Rush},
    year={2020},
    eprint={2002.00876},
    archivePrefix={arXiv},
    primaryClass={cs.CL}
}

@article{jaxopt_implicit_diff,
  title={Efficient and Modular Implicit Differentiation},
  author={Blondel, Mathieu and Berthet, Quentin and Cuturi, Marco and Frostig, Roy
   and Hoyer, Stephan and Llinares-L{\'o}pez, Felipe and Pedregosa, Fabian
   and Vert, Jean-Philippe},
  journal={arXiv preprint arXiv:2105.15183},
  year={2021}
}

@article{synjax2023,
      title="{SynJax: Structured Probability Distributions for JAX}",
      author={Milo\v{s} Stanojevi\'{c} and Laurent Sartran},
      year={2023},
      journal={arXiv preprint arXiv:2308.03291},
      url={https://arxiv.org/abs/2308.03291},
}

@inproceedings{cvxpylayers2019,
  author={Agrawal, A. and Amos, B. and Barratt, S. and Boyd, S. and Diamond, S. and Kolter, Z.},
  title={Differentiable Convex Optimization Layers},
  booktitle={Advances in Neural Information Processing Systems},
  year={2019},
}

@inproceedings{
loshchilov2018decoupled,
title={Decoupled Weight Decay Regularization},
author={Ilya Loshchilov and Frank Hutter},
booktitle={International Conference on Learning Representations},
year={2019},
url={https://openreview.net/forum?id=Bkg6RiCqY7},
}

@article{robbinsmonro,
author = {Herbert Robbins and Sutton Monro},
title = {{A Stochastic Approximation Method}},
volume = {22},
journal = {The Annals of Mathematical Statistics},
number = {3},
publisher = {Institute of Mathematical Statistics},
pages = {400--407},
year = {1951},
doi = {10.1214/aoms/1177729586},
URL = {https://doi.org/10.1214/aoms/1177729586}
}

@inproceedings{nesterov1983method,
  title={A method for solving the convex programming problem with convergence
      rate $O(1/k^2)$},
  author={Nesterov, Yurii},
  booktitle={Dokl akad nauk Sssr},
  volume={269},
  pages={543},
  year={1983}
}

@article{acceleration,
url = {http://dx.doi.org/10.1561/2400000036},
year = {2021},
volume = {5},
journal = {Foundations and Trends® in Optimization},
title = {Acceleration Methods},
doi = {10.1561/2400000036},
issn = {2167-3888},
number = {1-2},
pages = {1-245},
author = {Alexandre d’Aspremont and Damien Scieur and Adrien Taylor}
}

\end{document}